\documentclass{article}

\usepackage{amsthm}

\usepackage[accent=0072B2,font=palatino,headings=sans]{simpleicml}
\usepackage{sectionnav}
\makeatother
\usepackage{capt-of} 

\usepackage{graphicx} 
\usepackage{wrapfig}
\usepackage{booktabs}
\usepackage{multirow}
\usepackage{makecell}
\usepackage{pifont}
\usepackage{colortbl}
\usepackage{xcolor}
\usepackage{array}
\usepackage{caption}
\usepackage{enumitem}
\usepackage{tabularx}
\usepackage{booktabs}

\newcommand{\lbl}[1]{\textbf{\textcolor{teal!60!black}{\textsf{#1}}}}
\newcommand{\modulename}[2]{\colorbox{#1!10}{\makebox[\dimexpr\linewidth-2\fboxsep\relax][l]{\textbf{\textsf{#2}}}}}

\newcommand{\var}[1]{\texttt{#1}}
\newcommand{\inlbl}[1]{\textbf{#1}}
\newcommand{\jsonbox}[1]{\colorbox{gray!10}{\texttt{#1}}}
\newcommand{\promptsection}[2]{%
    \medskip\noindent{\small\color{#1}\textbf{#2}}\par\smallskip%
}
\newcommand{\schema}[1]{\textbf{\textcolor{teal!40!black}{\textsc{#1}}}}
\newcommand{\jsontext}[1]{{\small\fontfamily{cmss}\selectfont #1}}

\newcommand{\placeholder}[1]{\textit{\textlangle#1\textrangle}}
\newcommand{\jsonsym}[1]{\textbf{#1}}
\usepackage[table,dvipsnames]{xcolor}
\usepackage{enumitem}
\usepackage[T1]{fontenc}

\definecolor{headerblue}{HTML}{EBF3FF}
\definecolor{textred}{HTML}{980000}
\definecolor{textgreen}{HTML}{274E13}
\definecolor{mygray}{gray}{0.5}

\definecolor{lightblue}{RGB}{220,230,241}
\definecolor{graytext}{gray}{0.4}

\definecolor{darkblue}{rgb}{0.0,0.0,0.7}
\definecolor{darkred}{rgb}{0.7,0.0,0.0}

\usepackage{hyperref} 
\hypersetup{
    colorlinks=true,
    linkcolor=black,
    citecolor=darkblue,
    urlcolor=darkblue,
}

\icmltitle{Reward Prediction with Factorized World States}

\icmlauthors{%
  Yijun Shen\textsuperscript{1,*} \quad
  Delong Chen\textsuperscript{2,*} \\
  Xianming Hu\textsuperscript{1} \ 
  Jiaming Mi\textsuperscript{1} \ 
  Hongbo Zhao\textsuperscript{1} \ 
  Kai Zhang\textsuperscript{1,†} \ 
  Pascale Fung\textsuperscript{2}
}
\icmlaffiliations{%
  \textsuperscript{1} East China Normal University \quad 
  \textsuperscript{2} HKUST \quad \\
  * Equal contribution \quad 
  † Corresponding Author \\
  \texttt{kzhang@cs.ecnu.edu.cn}
}
\icmlabstract{

Agents must infer action outcomes and select actions that maximize a reward signal indicating how close the goal is to being reached. Supervised learning of reward models could introduce biases inherent to training data, limiting generalization to novel goals and environments. In this paper, we investigate whether well-defined world state representations alone can enable accurate reward prediction across domains.
To address this, we introduce ‌\textsc{StateFactory}, a factorized representation method that transforms unstructured observations into a hierarchical object-attribute structure using language models. This structured representation allows rewards to be estimated naturally as the semantic similarity between the current state and the goal state under hierarchical constraint.
Overall, the compact representation structure induced by ‌\textsc{StateFactory} enables strong reward generalization capabilities. 
We evaluate on \textsc{RewardPrediction}, a new benchmark dataset spanning five diverse domains and comprising 2,454 unique action-observation trajectories with step-wise ground-truth rewards. 
Our method shows promising zero-shot results against both VLWM-critic and LLM-as-a-Judge reward models, achieving 60\% and 8\% lower EPIC distance, respectively.
Furthermore, this superior reward quality successfully translates into improved agent planning performance, yielding success rate gains of +21.64\% on \texttt{AlfWorld} and +12.40\% on \texttt{ScienceWorld} over reactive system-1 policies and enhancing system-2 agent planning. 
Project Page: \url{https://statefactory.github.io}

}
\icmlrunningtitle{\color{black} Reward Prediction with Factorized World States}
\usepackage{natbib}
\usepackage{hyperref}
\usepackage{microtype}
\usepackage{graphicx}
\usepackage{subfigure}
\usepackage{booktabs} 
\usepackage[toc,page,header]{appendix}
\usepackage{minitoc}
\usepackage{makecell}
\usepackage{dblfloatfix}
\usepackage{amsmath}
\usepackage[capitalize,noabbrev]{cleveref}
\usepackage{mathtools}
\usepackage{cuted} 
\usepackage[most]{tcolorbox}
\tcbuselibrary{theorems}
\usepackage{caption}
\usepackage{enumitem}
\usepackage{listings}

\usepackage{tgheros} 
\definecolor{bg}{HTML}{F8FAFD}
\definecolor{keyword}{HTML}{0077B6}
\definecolor{string}{HTML}{E76F51}
\definecolor{comment}{HTML}{A0AEC0}
\definecolor{number}{HTML}{457B9D}
\definecolor{function}{HTML}{2A9D8F}
\definecolor{class}{HTML}{F4A261}
\definecolor{text}{HTML}{2D3A4A}
\lstdefinestyle{pytorchheros}{
    backgroundcolor=\color{bg},
    basicstyle=\fontfamily{qhv}\selectfont\footnotesize\color{text}, 
    keywordstyle=\color{keyword}\bfseries,
    stringstyle=\color{string},
    commentstyle=\color{comment}\itshape,
    numberstyle=\color{number},
    identifierstyle=\color{text},
    classoffset=1,
    morekeywords={Net},
    keywordstyle=\color{class}\bfseries,
    classoffset=0,
    emph={__init__,forward,print}, emphstyle=\color{function},
    frame=single,
    framerule=0pt,
    rulecolor=\color{bg},
    tabsize=4,
    showstringspaces=false,
    breaklines=true,
    linewidth=\linewidth,
    xleftmargin=0em,
    xrightmargin=0em,
    aboveskip=1em,
    belowskip=1em,
    literate={~}{{\textasciitilde}}1
}
\renewcommand\lstlistingname{Algorithm}
\crefname{lstlisting}{\MakeLowercase\lstlistingname}{\MakeLowercase\lstlistingname s}
\Crefname{lstlisting}{\lstlistingname}{\lstlistingname s}
\usepackage{float} 
\newfloat{lstfloat}{htbp}{lop}
\floatname{lstfloat}{Listing}

\usepackage{amsmath,amsthm}
\usepackage{bm,bbm}

\usepackage{varwidth}

\usepackage{hyperref}
\usepackage{cleveref}
\usepackage{mathtools}
\usepackage{algpseudocode}
\usepackage{mdframed}

\usepackage{multicol}
\usepackage{multirow}
\usepackage{setspace}

\usepackage{colortbl}

\usepackage[svgnames]{xcolor}
\usepackage{framed}

\newcommand{\1}{\mathbf{1}}

\usepackage{tikz,ifthen}
\usetikzlibrary{positioning}
\usetikzlibrary{shapes}
\usetikzlibrary{arrows}
\usetikzlibrary{fit}
\usetikzlibrary{calc}
\usetikzlibrary{shapes.misc}

\crefname{defi}{defn.}{defns.}

\def\1{\bm{1}}

\DeclareMathAlphabet{\mathsfit}{\encodingdefault}{\sfdefault}{m}{sl}
\SetMathAlphabet{\mathsfit}{bold}{\encodingdefault}{\sfdefault}{bx}{n}

\sectionheaderline{
  \seclink{sec:intro}{1}{Intro} | 
  \seclink{sec:benchmark}{2}{Benchmark} | 
  \seclink{sec:methods}{3}{Method} |
  \seclink{sec:experiments}{4}{Experiments} |
  \seclink{sec:related-works}{5}{Related Works} |
  \seclink{sec:conclusion}{6}{Conclusion} |
  \seclink{sec:appendix}{7}{Appx}
}

\begin{document}
\icmlmaketitle

\newtcbtheorem
  [
  crefname={detail}{detail}]
  {detail}
  {Experiment Details}
  {%
    fontupper=\small,
    colback=orange!5,
    colframe=orange!35!black,
    fonttitle=\bfseries,
    boxsep=1pt,
    left=1.5mm,
    right=1.5mm,
    top=2mm,
    bottom=1mm,
  }
  {detail}

\newtcbtheorem
  [
  crefname={def.}{def.}]
  {definition}
  {Definition}
  {%
    fontupper=\small,
    colback=green!5,
    colframe=green!35!black,
    fonttitle=\bfseries,
    boxsep=1pt,
    left=1.5mm,
    right=1.5mm,
    top=2mm,
    bottom=1mm,
  }
  {def}

\newtcbtheorem
  [
  crefname={thm.}{thms.}]
  {theorem}
  {Theorem}
  {%
    fontupper=\small,
    colback=red!5,
    colframe=red!35!black,
    fonttitle=\bfseries,
    boxsep=1pt,
    left=1.5mm,
    right=1.5mm,
    top=2mm,
    bottom=1mm,
  }
  {theorem}
\newtcbtheorem
  [
  crefname={lemma.}{lemmas.}]
  {lemma}
  {Lemma}
  {%
    fontupper=\small,
    colback=blue!3,
    colframe=blue!35!black,
    fonttitle=\bfseries,
    boxsep=1pt,
    left=1.5mm,
    right=1.5mm,
    top=2mm,
    bottom=1mm,
  }
  {lemma}
\newtcbtheorem
  [
  crefname={prop.}{props.}]
  {proposition}
  {Proposition}
  {%
    breakable,
    enhanced,
    fontupper=\small,
    colback=red!5,
    colframe=red!35!black,
    fonttitle=\bfseries,
    boxsep=1pt,
    left=1.5mm,
    right=1.5mm,
    top=2mm,
    bottom=1mm,
  }
  {proposition}
\newtcbtheorem
  [
  crefname={cor.}{corrs.}]
  {corollary}
  {Corollary}
  {%
    breakable,
    enhanced,
    fontupper=\small,
    colback=red!5,
    colframe=red!35!black,
    fonttitle=\bfseries,
    boxsep=1pt,
    left=1.5mm,
    right=1.5mm,
    top=2mm,
    bottom=1mm,
  }
  {corollary}

\medskip
\section{Introduction}
\label{sec:intro}

We are interested in building agents with planning capabilities that generalize across different goals and environments in zero-shot. Many real-world scenarios do not allow agents to perform massive trial-and-error to obtain explicit reward signals, making model-free approaches, \textit{e.g.,} reinforcement learning from verifiable rewards (RLVR) \citep{guo2025deepseek} not applicable. This limitation necessitates \textit{world models} \citep{ha2018world, richens2025general} that predict the consequences of actions as future world states to guide policy learning \citep{hafner2023mastering} or plan search \citep{lecun2022path, hansen2023td}.

To leverage the state predictions of world models, agents need to evaluate \textit{to what extent a predicted state aligns with the goal}, typically via a scalar reward, and then maximize it \citep{m2023model}. Our work focuses on accurately predicting such rewards with effective \textit{generalization}. One direct approach is to train the models \citep{chen2025planning} or reward prediction heads on top of world models \citep{hafner2023mastering} using task-specific supervision. However, explicit supervision may introduce biases and overfitting \citep{di2022goal, gao2023scaling, casper2023open, sharma2023towards}, limiting generalization to the novel environments. A promising alternative is to estimate task progress by measuring the semantic distance between the current state and the desired goal state \citep{liu2022goal, schaul2015universal}, effectively avoiding explicit reward modeling and enabling zero-shot planning \citep{ma2023liv, wang2023optimal}.

There are two bottlenecks in using world-state distances as reward signals. 
First, constructing an abstract state representation space \citep{m2023model} whose geometry accurately reflects task progress \citep{trott2019keeping, wu2018laplacian} is nontrivial. Prior successes leveraged representations from visual foundation models, such as DINO-WM \citep{zhou2411dino}, JEPA-WMs \citep{terver2025drives}, and RoboCLIP \citep{sontakke2023roboclip}. Yet, how to extend from low-level continuous planning to high-level language-based agent planning \citep{wang2024survey, luketina2019survey, cao2024survey}, especially for procedural tasks \citep{chen2025planning} that require stronger semantic and temporal abstraction \citep{sutton1998between}, remains an open challenge.

Second, for agents that operate in text space, rigorous evaluation of reward quality (especially in terms of fine-grained, step-wise proximity to the goal state) is difficult due to the lack of suitable benchmarks. Existing datasets mainly focus on sparse, outcome-oriented rewards, which can make it 
difficult to systematically assess quality of rewards and how they guide the planning process.

We address both challenges by introducing a new \textbf{state representation method} and a \textbf{dedicated benchmark} for rigorous reward evaluation. Specifically, we propose \textsc{StateFactory}, a semantic factorization framework that factorizes world states into hierarchical object–attribute structures. Within this framework, unstructured observations are decomposed into atomic semantic factors using large language models. The reward signals are then naturally estimated by measuring semantic similarity between the current and goal states through the hierarchical routing. This process effectively distills the unstructured observations into compact abstractions, thereby improving the zero-shot reward generalization.

To enable fine-grained evaluation of reward prediction, we introduce \textsc{RewardPrediction}, a benchmark spanning five interactive environments: \texttt{AlfWorld} \citep{shridhar2020alfworld} for robotics planning, \texttt{ScienceWorld} \citep{wang2022scienceworld} for scientific reasoning, \texttt{TextWorld} \citep{cote2018textworld} for text-based games, \texttt{WebShop} \citep{yao2022webshop} for website navigation, and \texttt{BlocksWorld} \citep{valmeekam2023planbench} for classical planning. In total, there are 2,454 unique trajectories, each with step-wise action–observation pairs and scalar rewards, allowing any predicted reward to be evaluated via the EPIC distance \citep{gleave2020quantifying} to the ground-truth rewards.

Overall, the representation structure in \textsc{StateFactory} demonstrates strong reward generalization capabilities. While supervised reward models exhibit a \textbf{138\% increase in reward prediction error} in novel domains on \textsc{RewardPrediction}, \textsc{StateFactory} achieves superior zero-shot performance, reducing EPIC distance by 60\% and 8\% compared to VLWM-critic and LLM-as-a-Judge, respectively. These improvements in reward quality turn directly into enhanced agent planning performance, yielding success rate gains of +21.64\% on AlfWorld and +12.40\% on ScienceWorld over reactive system-1 policies, while consistently strengthening system-2 agent planning.

The main contributions of this work are summarized:
\begin{itemize}
    \item We introduce \textsc{RewardPrediction} benchmark, covering 2,454 unique trajectories with step-wise ground truth reward across five text-based environments.
    \item We propose \textsc{StateFactory}, a new representation method to transform the flat text descriptions to the structured hierarchies for the generalizable reward prediction. 
    \item We demonstrate that \textsc{StateFactory} representation is able to enhance agent planning performance when combined with a world model and action proposal.

\end{itemize}

\begin{figure}[t]
    \centering
    \includegraphics[width=1.0\linewidth]{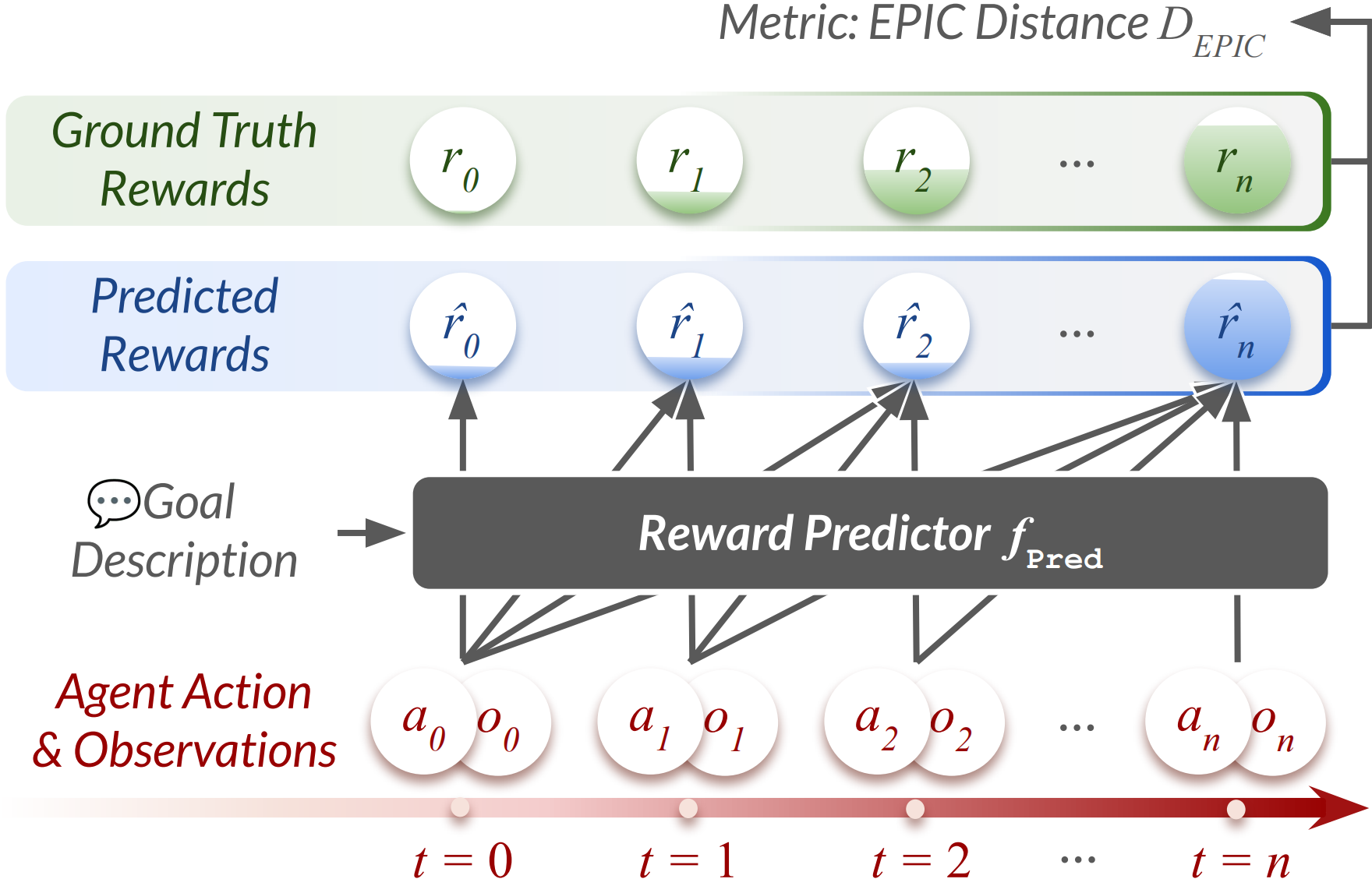}
    \caption{\textbf{\textsc{RewardPrediction} benchmark Overview.} Given a textual goal description, the model computes step-wise progress estimates $[\hat{r}_t]_{t=0}^n$ from sequences of action-observation pairs. The reward predictions are compared against ground-truth $[r_t]_{t=0}^n$ using EPIC distance \citep{gleave2020quantifying} to quantify alignment.}
    \label{fig:benchmark}
\end{figure}

\begin{figure*}
    \centering
    \includegraphics[width=1.0\linewidth]
    {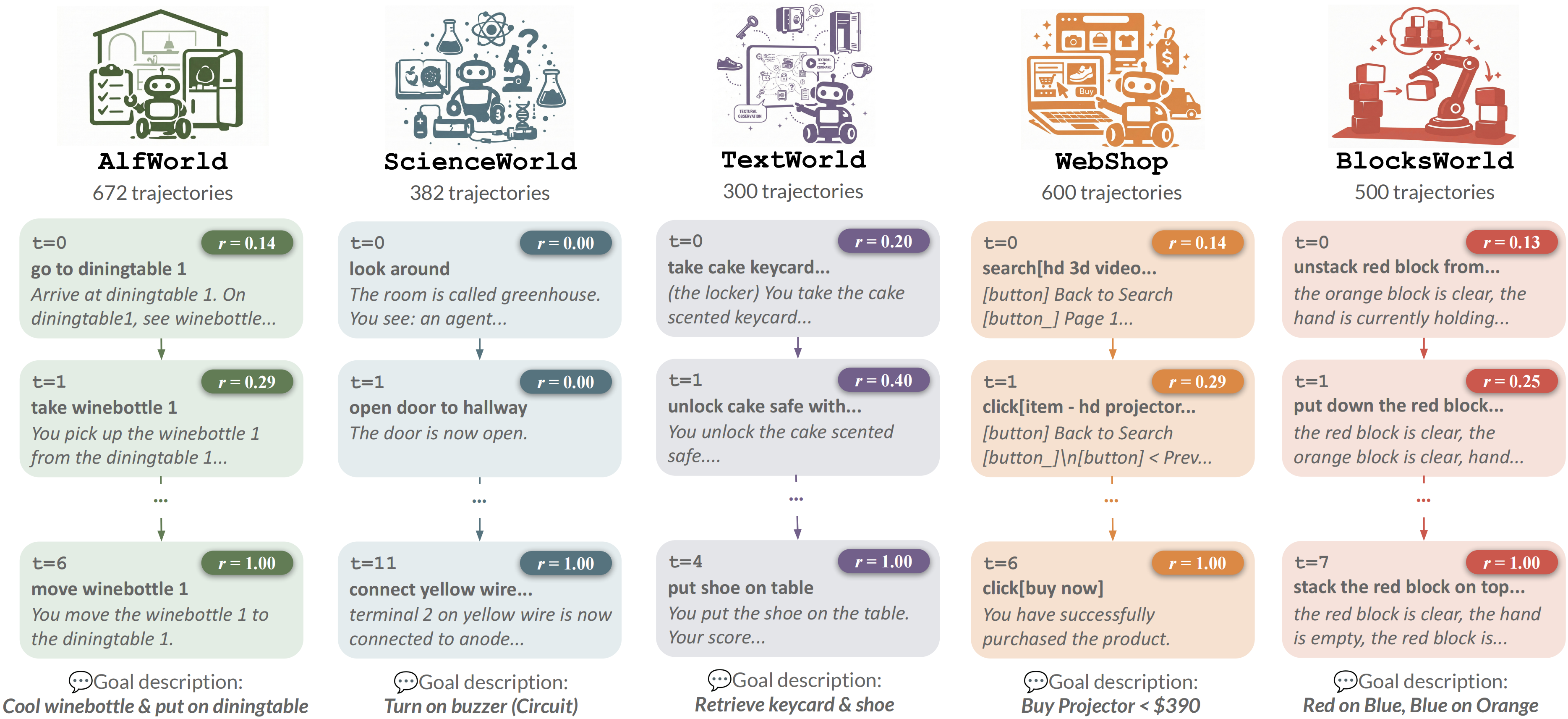}
    \caption{\textbf{\textsc{RewardPrediction} Benchmark Overview.} Representative trajectories across five diverse domains. Each column displays a task instance with actions, observations, and the corresponding ground-truth task progress score ($R \in [0, 1]$) at key time steps.}
    \label{fig:rewardprediction_demo}
\end{figure*}

\section{The \textsc{RewardPrediction} Benchmark}
\label{sec:benchmark}

We introduce the \textsc{RewardPrediction} benchmark to evaluate zero-shot reward prediction across different domains. In the following sections, we first present the formal task formulations (\S\ref{sec:task_formulation}), and then describe implementation details of benchmark construction (\S\ref{sec:trajectory_collection}). Figure~\ref{fig:benchmark} provides an overview of the benchmark task, and Figure~\ref{fig:rewardprediction_demo} presents representative examples of different domains.

\subsection{Formulation}
\label{sec:task_formulation}

Since we are interested in building objective-driven goal-reaching agents, we consider a Goal-Augmented Markov Decision Process (GA-MDP) \citep{liu2022goal} with Partial Observation \citep{kaelbling1998planning} constraints: $\mathcal{M} = \langle \mathcal{S}, \mathcal{A}, \mathcal{O}, \mathcal{T}, \mathcal{R}, \mathcal{G}, \phi,  \Omega \rangle$. Here $\mathcal{S}$, $\mathcal{A}$, and $\mathcal{O}$ respectively denote state, action, and the observation space. $\mathcal{T}$ is the underlying transition function of the environment. The function $\Omega: \mathcal{S}\rightarrow\mathcal{O}$ maps underlying states into agent observations. Each observation $o_t$ usually partially contains the information of the ground truth states $s_t$. $\mathcal{R}$ denotes reward function and $\mathcal{G}$ is the goal space--we consider language goal in this work. Given a textual goal description $g\in\mathcal{G}$ and an agent that reaches $s_t$, the corresponding reward $\mathcal{R}(s_t, g)$ should measure how close is the agent from achieving the goal. With the mapping $\phi: \mathcal{G} \rightarrow \mathcal{S}$ that interprets the goals and translates them into state space, it should have $\mathcal{R}(s_t, g) = 1$ if the goal is reached, \textit{i.e.,} when $||s_t-\phi(g)||\leq\epsilon $ with a small $\epsilon$ \citep{andrychowicz2017hindsight}. To allow representing the richer task progress information, we adopt the \textit{distance-to-goal} reward \citep{trott2019keeping, liu2022goal}:
\begin{equation}
    \mathcal{R}(s_t, g) = 1 - ||s_t-\phi(g)||.
    \label{eq:reaward}
\end{equation}
\textbf{Prediction Task}. At each time $t$, an agent $\pi$ perceives an observation $o_{t}$, takes an action $a_t$, and obtains the reward $r_t$.  We can collect offline trajectories of agent interacting with environments to achieve goals, each expressed as $\langle g,  [o_t, a_t, r_t]_{t=0}^n\rangle$. Based on collected goals and trajectories, we define the \textsc{RewardPrediction} task (Fig.~\ref{fig:benchmark}), which requires a reward predictor $f_\texttt{pred}$ to estimate ground truth rewards based on the goal to achieve, past observations, and action history:
\begin{equation}
    f_\texttt{pred}(g, o_{[0:t]}, a_{[0:t]})\rightarrow \hat{r}_t.
    \label{eq:task}
\end{equation}
Iterating over every step, we obtain a sequence of predicted rewards $[\hat{r}_t]_{t=0}^n$. Prediction of each step could be independent (\textit{stateless}), but alternatively, $f_\texttt{pred}$ is also allowed to keep and access intermediate results from previous steps in the trajectory (\textit{stateful}), \textit{e.g.,} allowing them to maintain a memory of belief states for more efficient prediction.

\textbf{Evaluation Metric}. We adopt the Equivalent Policy-Invariant Comparison (EPIC) distance \citep{gleave2020quantifying, frick2024evaluate} between ground truth rewards and predicted rewards as the metric. EPIC distance is an approach to quantify the difference between reward functions, providing a policy-invariant measure that preserves the fine-grained magnitude information essential for robust planning. Drawing on~\citet{rocamonde2023vision}, which simplifies EPIC to a closed-form function of Pearson correlation for state-only rewards, we formulate the metric as:\looseness=-1
\begin{equation}
    D_{\text{EPIC}} = \frac{1}{\sqrt{2}} \sqrt{1 - \rho(\hat{R}, R)},
    \label{eq:epic_metric}
\end{equation}

where $\rho$ represents Pearson correlation, $\hat{R}=[\hat{r}_t]_{t=0}^n$ are the predicted rewards and $R=[r_t]_{t=0}^n$ are the ground truths. We report average EPIC distance across all trajectories as the final performance.

\subsection{Implementation}
\label{sec:trajectory_collection}

To ensure robustness across diverse physical and semantic landscapes, we construct this benchmark by collecting \textit{task-balanced} interaction trajectories directly from a suite of five real text-based environments.
To mitigate heuristic reward hacking, we employ a paired data construction strategy. Specifically, each instance consists of a positive-negative pair tailored to the same task. The positive anchor is derived from an expert trajectory with deterministic progress gradients\citep{yang2024rank2reward} ($r_t = t/T$)~\citep{ziakas2025test}. Crucially, we augment these sequences with random interaction steps at the boundaries, inherit their native environment rewards, to decouple trajectory length from reward signals. Complementing this, the negative baseline is generated via a random policy to represent failure trajectories ($r_t=0$). This structural coupling forces the model to explicitly distinguish meaningful semantic progress from stochastic noise within the same task context.
The final evaluation set comprises \textbf{2,454 trajectories}, as illustrated in Figure~\ref{fig:rewardprediction_demo}. For the supervised training stage, we scale up data generation to produce approximately 350,000 trajectories.


\textbf{\texttt{AlfWorld}}~\citep{shridhar2020alfworld}. A benchmark for embodied household instruction following requiring agents to map high-level goals (e.g., \textit{put a hot mug in cabinet}) to sequences of executable step-level actions (e.g., \texttt{take mug}, \texttt{heat mug}, and \texttt{put mug in cabinet}). Our dataset covers all six standard interaction types, including \textit{Pick \& Place}, \textit{Examine}, \textit{Cool}, \textit{Heat}, \textit{Clean}, and \textit{Movable}, comprising a total of 672 expert trajectories.

\textbf{\texttt{ScienceWorld}} \citep{wang2022scienceworld}. A complex environment simulating elementary science experiments across topics like thermodynamics and life sciences. It requires agents to perform multi-step causal reasoning, such as \textit{``determine the melting point of lead''} via tool-use actions like \texttt{move lead to stove}, \texttt{turn on stove}, and \texttt{focus on thermometer}. We curated 382 trajectories from the test.
 
\begin{figure}
    \centering
    \includegraphics[width=0.95\linewidth]{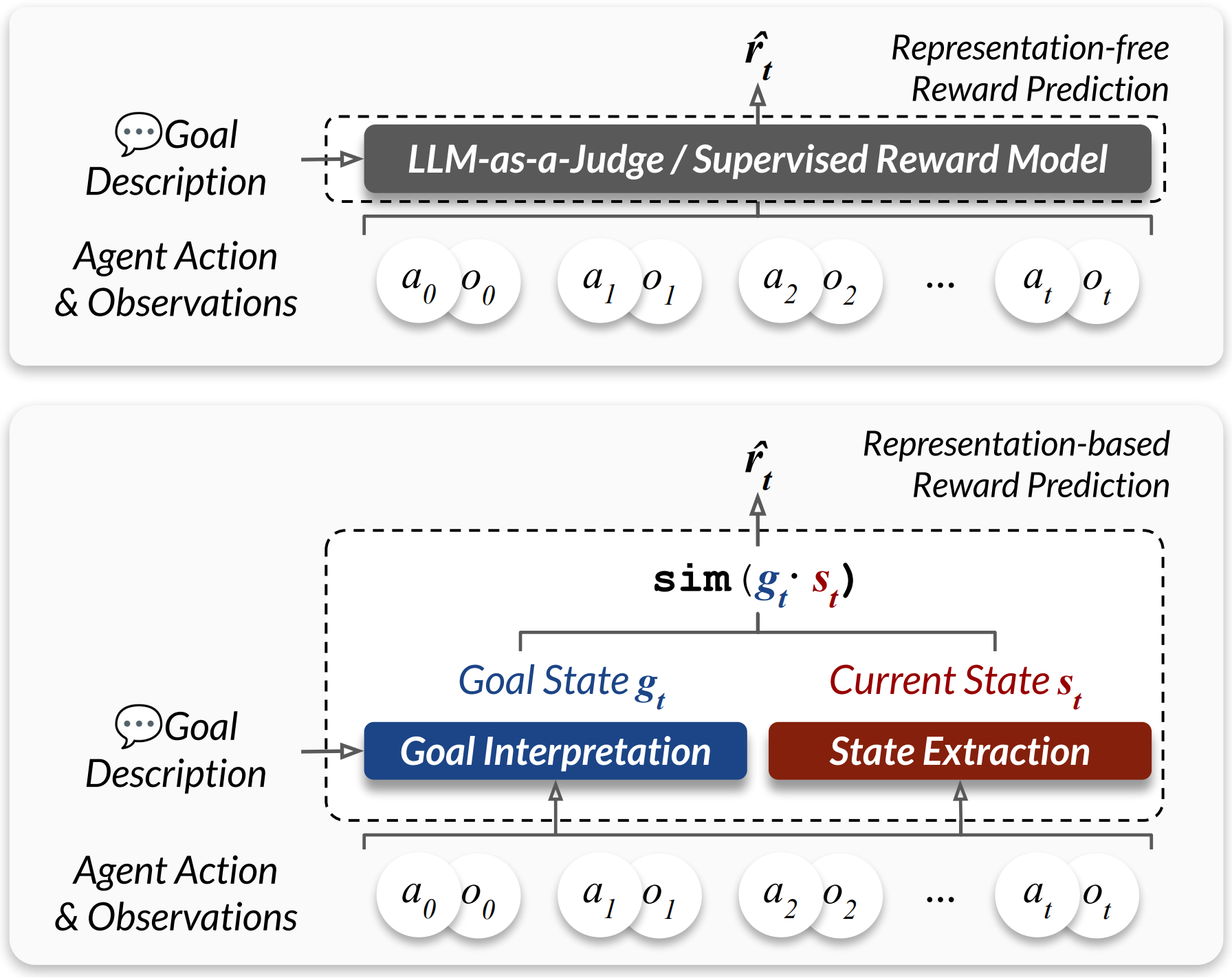}
    \vspace{5pt}
    \caption{\textbf{Two paradigms of reward prediction.} Unlike representation-free methods that regress rewards directly from raw inputs (top), representation-based frameworks derive progress signals by measuring the alignment between factorized state representations $s_t$ and goal interpretations $g_t$ (bottom).}
    \label{fig:formulation}
\end{figure}

\textbf{\texttt{TextWorld}} \citep{cote2018textworld}. A generative framework for procedural text-adventure games requiring multi-step reasoning. We focused on the quests involving complex object manipulation and the lock-and-key puzzles (e.g., \textit{``retrieve the shoe from the safe and place it on the table''}). These tasks require the sequence-dependent actions such as \texttt{take keycard}, \texttt{unlock safe with the keycard}, and \texttt{put shoe on the table}. We utilized the native engine to procedurally generate 300 trajectories focusing on these state-change puzzles.

\begin{figure*}
    \centering
    \includegraphics[width=\linewidth]{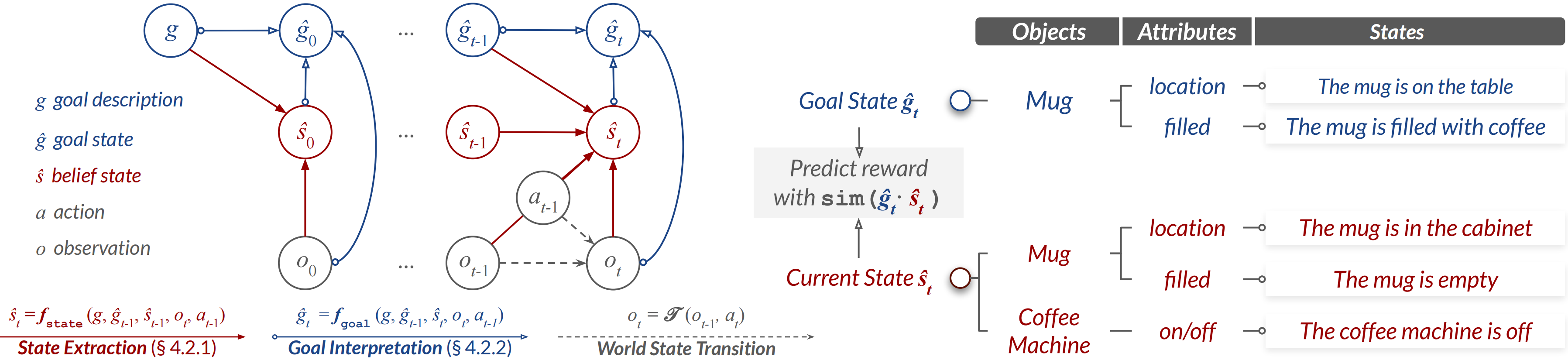}
    \caption{\textbf{\textsc{StateFactory} Framework.} State extraction and goal interpretation are coupled recurrent processes (left). Our approach factorizes states into explicit objects and attributes (right), deriving dense rewards from semantic similarity between $\hat{s}_t$ and $\hat{g}_t$.}
    \label{fig:StateFactory}
\end{figure*}

\textbf{\texttt{WebShop}} \citep{yao2022webshop}. A scalable simulation of e-commerce website navigation requiring agents to find products matching specific attributes (e.g., \textit{``looking for a blue running shoe size 9''}). Actions simulate real browser interactions, including keyword searches and selecting options (e.g., \texttt{search[blue shoe]}, \texttt{click[size-9]}, \texttt{click[buy]}). Unlike other domains, we leveraged \textbf{human demonstrations} to capture realistic navigation nuances. From a pool of successful human trajectories, we subsampled high-quality instances to serve as positive anchors, forming a total of 600 trajectories.
 
\textbf{\texttt{BlocksWorld}} \citep{valmeekam2023planbench}. A classical planning benchmark for evaluating \textbf{multi-step} spatial reasoning and stack manipulation under strict physical constraints. Tasks involve reconfiguring block arrangements to match a target state (e.g., \textit{``stack block A on top of block B''}) using atomic actions like \texttt{pick up}, \texttt{unstack}, and \texttt{put down}. We subsampled official instances and utilized a PDDL-to-text translator to ground symbolic states (e.g., \texttt{on(a,b)}) into dense natural language descriptions. This process bridges the gap between symbolic logic and linguistic reasoning, constructing 500 trajectories that challenge the agent's ability to track long-horizon dependencies.

\section{Reward Prediction Methods}
\label{sec:methods}

In this section, we explore different approaches to solving the \textsc{RewardPrediction} benchmark. We introduce two families of methods: \textit{representation-free} methods and \textit{representation-based} methods, as illustrated in Fig.~\ref{fig:formulation}. The key distinction lies in whether the reward predictor constructs explicit world state representations (belief states) and explicit goal interpretations (approximating $\phi$). Despite architectural differences, methods can also be categorized by whether they leverage the \textsc{RewardPrediction} training set, resulting in two regimes: \textit{supervised} predictors and \textit{zero-shot} predictors.

\subsection{Representation-free Methods}
\label{subsec:rep_free}

\subsubsection{Finetuned Language Models}

Inspired by reward models used in Reinforcement Learning from Human Feedback (RLHF) \citep{ouyang2022training} and the self-supervised critic used in the Vision Language World Model (VLWM) \citep{chen2025planning}, we train language models with linear head as $f_\texttt{pred}$ to predict the reward. Specifically, we generate paired samples from the training set of \textsc{RewardPrediction}, where the positive sample has a reward that is higher than the negative sample. We then train the language model with a contrastive ranking loss \citep{chen2025planning} with a reward centering term \citep{naik2024reward}. The reward model here is fully stateless. Given input goal description and (partial) trajectory at each time step, it predicts the reward independently. 

\subsubsection{LLM-as-a-Judge Prompting}

This approach leverages prompted LLM-as-a-Judge~\citep{zheng2023judging} as generative reward predictor ~\citep{guo2025reward, gunjal2025rubrics, whitehouse2025j1}. We provide goal and trajectory to an LLM, with additional guidance describing the reward prediction task, to instruct it to respond with a scaler reward on a scale of 0 to 1. See Appendix Table \ref{tab:baseline_full_prompts} for the full prompt. In addition to this simple stateless baseline, the flexibility of prompting allow us to implement a stateful variant: the LLM is instructed to construct and manage belief state over time--a summary as memory that tracks the world state.

\subsection{Representation-based Method (\textsc{StateFactory})}
\textsc{StateFactory} predicts task progress by decomposing the GA-MDP $\mathcal{M}$ into three integrated layers: \textit{world state transition}, \textit{state extraction} ($\hat{s}_t$), and \textit{goal interpretation} ($\hat{g}_t$). As illustrated in Fig.~\ref{fig:StateFactory}, raw observations $o_t = \Omega(s_t)$ are distilled into structured object-attribute states $\hat{s}_t$, while the textual goal $g \in \mathcal{G}$ is iteratively grounded into a dynamic goal state $\hat{g}_t$. The reward signal is derived via semantic alignment: $\hat{r}_t = \text{sim}(\hat{g}_t, \hat{s}_t)$. This factorization ensures zero-shot rewards remain grounded in physical transitions while staying responsive to specific goal description.

\subsubsection{State Extraction}
\label{subsec:state_extraction}

Unlike the unstructured representations that retain the task-irrelevant noise \citep{hao2023reasoning, chen2025planning} or simple object-centric approaches that fail to capture fine-grained attribute dynamics \citep{yoneda2024statler, feng2025learning}, our framework explicitly separates entity identity from evolving attributes.
\textsc{StateFactory} treats state extraction as a recurrent, goal-conditioned update process. We employ a tracking function $f_{\texttt{state}}$ to dynamically refine the semantic state $\hat{s}_t$ by integrating the raw observation $o_t$ with both execution history and task progress:
\begin{equation}
    \hat{s}_t = f_{\texttt{state}}(g, \hat{g}_{t-1}, \hat{s}_{t-1}, o_t, a_{t-1}),
    \label{equ:state_update}
\end{equation}
where $a_{t-1}$ and $\hat{s}_{t-1}$ enforce temporal consistency, while the goal context ($g, \hat{g}_{t-1}$) acts as an attentional prior to filter task irrelevant details. The resulting state $\hat{s}_t$ is structured as a set of $N$ object instances $\{e_i\}_{i=1}^N$:
\begin{equation} 
    \quad e_i = \left\langle \mathbf{d}_i, \{ (\boldsymbol{\alpha}_{i,l}, \mathbf{v}_{i,l}) \}_{l=1}^{L_i} \right\rangle, 
    \label{equ:state_structure} 
\end{equation}
where $\mathbf{d}_i$ denotes an identity (\textit{e.g., }\textit{``Mug''}) and $\{(\boldsymbol{\alpha}_{i,l}, \mathbf{v}_{i,l})\}$ represents the dynamic semantic attributes (\textit{e.g.,} \textit{``location: on the table''}).

\begin{table*}[t]
\centering
\small 
\renewcommand{\arraystretch}{1.25}
\setlength{\tabcolsep}{2.5pt} 
\setlength{\aboverulesep}{0pt}
\setlength{\belowrulesep}{0pt}

\definecolor{tableblue}{RGB}{235, 242, 255}
\definecolor{graytext}{RGB}{128, 128, 128}

\caption{\textbf{Main Results:} We compare \textsc{StateFactory} with baselines across RewardPrediction benchmark. \textsc{StateFactory} establishes a new SOTA among zero-shot methods and approaches supervised baselines. \textbf{Bold} and \underline{underline} denote the best and second-best results.}
\label{tab:main_results}

\resizebox{\textwidth}{!}{
    \begin{tabular}{c c c c c | ccccc | c}
    \toprule
    \multirow{2}{*}{\textbf{Method}} & \multirow{2}{*}{\textbf{Backbone}} & \multirow{2}{*}{\shortstack[c]{\textbf{Training Data}}} & \multirow{2}{*}{\textbf{Reasoning}} & \multirow{2}{*}{\textbf{Zero-shot}} & \multicolumn{6}{c}{\textbf{Reward Prediction Error} ($D_{\textsc{EPIC}} \downarrow$)} \\
    \cline{6-11}
    & & & & & \textbf{\texttt{AlfWorld}} & \textbf{\texttt{ScienceWorld}} & \textbf{\texttt{WebShop}} & \textbf{\texttt{BlocksWorld}} & \textbf{\texttt{TextWorld}} & \textbf{Overall} \\ 

    \midrule

    \textit{Monotonic Baseline} & -- & -- & -- & \ding{51} & 0.532 & 0.508 & 0.535 & 0.589 & 0.536 & 0.540 \\
    \midrule

    \multirow{6}{*}{\textit{Supervised RM}}
      & \multirow{6}{*}{\texttt{Qwen2.5-1.5B}} & AlfWorld & \multirow{6}{*}{N/A} & \multirow{6}{*}{\ding{55}} 
      & \color{graytext} \underline{0.212} & \color{graytext} 0.542 & \color{graytext} 0.618 & \color{graytext} 0.596 & \color{graytext} 0.414 & \color{graytext} \underline{0.476} \\
      & & ScienceWorld & & 
      & \color{graytext} 0.506 & \color{graytext} \underline{0.305} & \color{graytext} 0.661 & \color{graytext} 0.706 & \color{graytext} 0.580 & \color{graytext} 0.552 \\
      & & WebShop & & 
      & \color{graytext} 0.707 & \color{graytext} 0.624 & \color{graytext} \textbf{0.242} & \color{graytext} 0.706 & \color{graytext} 0.707 & \color{graytext} 0.597 \\
      & & BlocksWorld & & 
      & \color{graytext} 0.556 & \color{graytext} 0.471 & \color{graytext} 0.596 & \color{graytext} \textbf{0.472} & \color{graytext} 0.556 & \color{graytext} 0.530 \\
      & & TextWorld & & 
      & \color{graytext} 0.523 & \color{graytext} 0.658 & \color{graytext} 0.678 & \color{graytext} 0.604 & \color{graytext} \underline{0.203} & \color{graytext} 0.533 \\
      & & All Domains & & 
      & \color{graytext} \textbf{0.178} & \color{graytext} \textbf{0.283} & \color{graytext} \underline{0.285} & \color{graytext} \underline{0.489} & \color{graytext} \textbf{0.177} & \color{graytext} \textbf{0.282} \\
    
    \midrule

    \textit{VLWM-critic} & \texttt{Llama3.2-1B} & VLWM Trajectories & N/A & \ding{51} & 0.823 & 0.673 & 0.636 & 0.663 & 0.896 & 0.738 \\

    \midrule

    \multirow{4}{*}{\shortstack[c]{\textit{LLM-as-a-Judge}}}
      & \multirow{2}{*}{\texttt{Qwen3-14B}} & \multirow{4}{*}{--} & \ding{55} & \multirow{4}{*}{\ding{51}} & 0.395 & 0.542 & 0.403 & 0.466 & 0.316 & 0.424 \\
      & & & \ding{51} & & 0.370 & \underline{0.371} & \underline{0.356} & 0.436 & 0.211 & 0.349 \\
      & \multirow{2}{*}{\texttt{gpt-oss-20b}} & & Low & & 0.368 & 0.394 & 0.376 & \textbf{0.362} & \underline{0.119} & 0.324 \\
      & & & Medium & & \underline{0.366} & 0.391 & 0.374 & \underline{0.363} & \textbf{0.115} & \underline{0.322} \\

    \hline

    \rowcolor{tableblue}
    \textbf{\textsc{StateFactory}} & \texttt{gpt-oss-20b} & -- & Medium & \ding{51} & \textbf{0.285} & \textbf{0.288} & \textbf{0.286} & 0.427 & 0.201 & \textbf{0.297} \\

    \bottomrule
    \end{tabular}
}
\end{table*}

\subsubsection{Goal Interpretation}
\label{subsec: Goal Interpretation}
Effective goal interpretation must bridge abstract instructions with evolving physical states. Traditional methods typically fix the goal representation at initialization~\citep{mahmoudieh2022zero, xie2023text2reward}, which often creates an ``illusion of progress'' as they fail to adapt to environmental changes during task execution~\citep{xue2025illusion}.

In contrast to static methods, \textsc{StateFactory} treats goal interpretation as an iterative, state-aware process. As illustrated in the blue pathway of Figure~\ref{fig:StateFactory}, we employ a function $f_{\texttt{goal}}$ to dynamically update the goal state $\hat{g}_t$ based on the goal description $g \in \mathcal{G}$ and the current context:
\begin{equation}
    \hat{g}_t = f_{\texttt{goal}}(g, \hat{g}_{t-1}, \hat{s}_t, o_t, a_{t-1}).
    \label{equation_online_goal}
\end{equation}

\subsubsection{Hierarchical Routing}
\label{subsec:reward_estimation}

We derive the reward $\hat{r}_t$ by quantifying the semantic similarity between the goal state $\hat{g}_t$ and world state $\hat{s}_t$. This evaluation proceeds hierarchically, starting from fine-grained attribute alignment to global progress aggregation.

\textbf{Object Matching.}
For each goal object instance $e_k \in \hat{g}_t$, we identify its corresponding physical object instance in the current state $\hat{s}_t$. We search for the candidate $e_i$ that maximizes the joint consistency of identity and attributes:
\begin{equation}
\hat{r}_{k,t} = \max_{e_i \in \hat{s}_t} \left( \text{sim}(\mathbf{d}_k, \mathbf{d}_i) \cdot \psi_{\text{attr}}(e_k, e_i) \right),
\label{eq:object_matching}
\end{equation}
where $\text{sim}(\mathbf{d}_k, \mathbf{d}_i)$ measures identity similarity, ensuring the correct object is selected before evaluating its state.

\textbf{Attribute Matching.}
We compute state satisfaction $\psi_{\text{attr}}$ by averaging the similarity between each goal attribute value $\mathbf{v}_k$ and its semantically aligned counterpart $\hat{\mathbf{v}}_i$ in candidate:
\begin{equation}
\psi_{\text{attr}}(e_k, e_i) = \frac{1}{|A_k|} \sum_{(\boldsymbol{\alpha}_k, \mathbf{v}_k) \in A_k} \text{sim}(\mathbf{v}_k, \hat{\mathbf{v}}_i),
\label{eq:attr_score}
\end{equation}
where $\hat{\mathbf{v}}_i$ denotes the value in $e_i$ associated with the key $\boldsymbol{\alpha}_i$ that maximizes semantic similarity to the goal key $\boldsymbol{\alpha}_k$.

\textbf{Reward Prediction.}
The global reward signal $\hat{r}_t$ is computed by aggregating the local fulfillment scores across all goal object instances to quantify the overall task progress:
\begin{equation}
    \hat{r}_t = \frac{1}{|\hat{g}_t|} \sum_{e_k \in \hat{g}_t} \hat{r}_{k,t}.
    \label{eq:global_reward}
\end{equation}

\begin{figure*}
    \centering
    \includegraphics[width=1.0\linewidth]{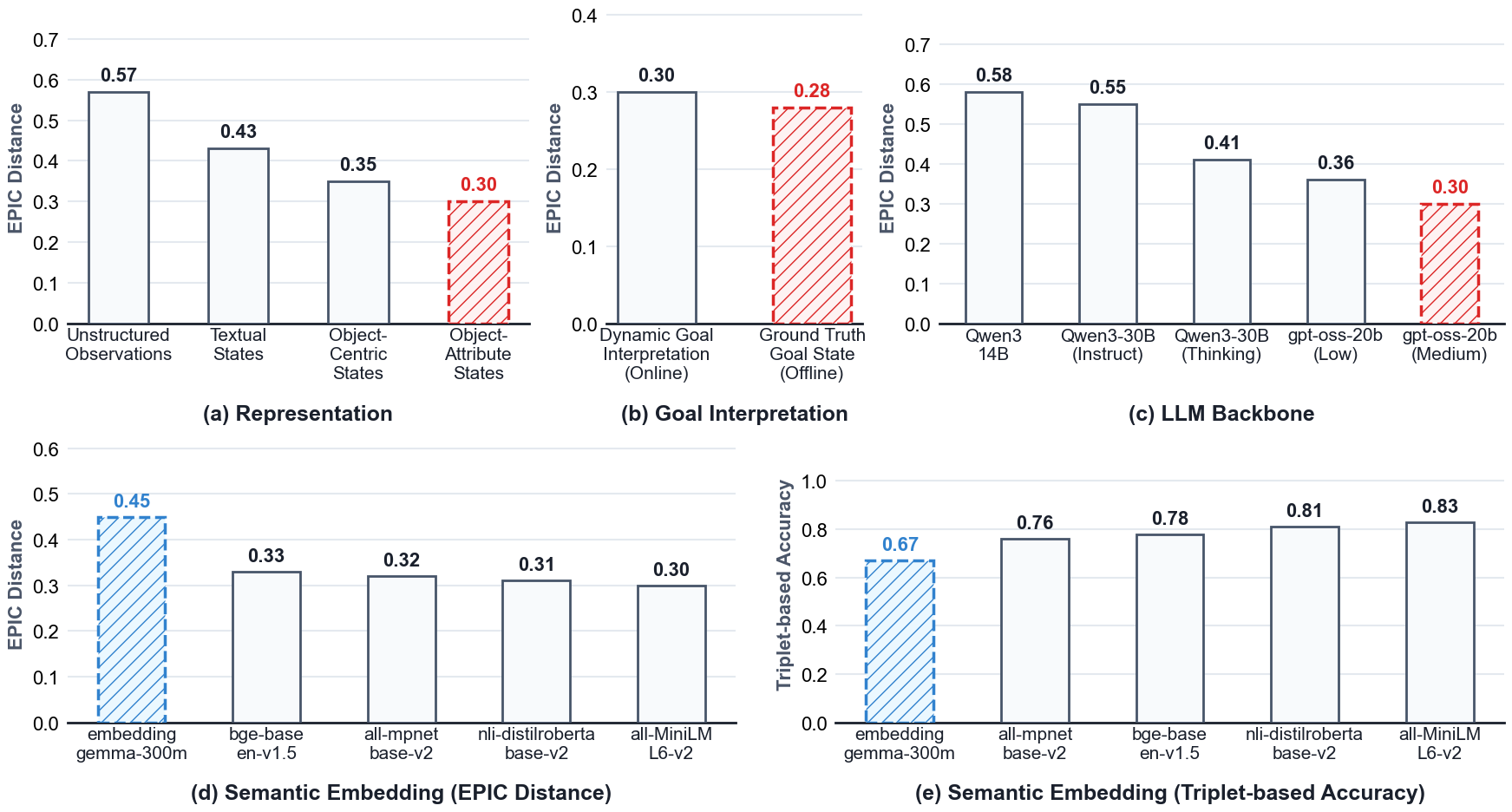}
    \caption{\textsc{StateFactory} ablation experiments. We report the EPIC distance ($D_{\text{EPIC}}$) across settings (a-d) and Triplet-based Accuracy for embedding models (e). For metrics, lower $D_{\text{EPIC}}$ and higher Accuracy indicate better performance. Specifically, Triplet-based Accuracy reflects the model's capability to enforce smaller distances for positive pairs compared to negative ones. Red hatched bars highlight the best performing configurations. Conversely, blue hatched bars indicate the worst performing ones.}
    \label{fig:abolation}
\end{figure*}

\section{Experiments}
\label{sec:experiments}

In this section, we evaluate \textsc{StateFactory} across diverse planning domains to address the following research questions:
1) whether our zero-shot approach provides more accurate reward signals than baselines; 
2) whether it generalizes better to new domains compared to supervised models;
3) how structural granularity contributes to the performance improvement; 
4) how robust the framework is to different design choices (e.g., embeddings, reasoning capacity); and
5) whether these signals can effectively guide online planning to improve success rates.

\subsection{Baselines and Setup}
\label{sec:setup}
We standardize experiments on \textsc{RewardPrediction} using gpt-oss-20b for state extraction and all-MiniLM-L6-v2 for alignment. We compare \textsc{StateFactory} against three distinct categories of baselines: (1) a naive \texttt{Monotonic Baseline}; (2) supervised approaches, including VLWM-critic~\citep{chen2025planning} and other reward models explicitly trained on \textsc{RewardPrediction}; and (3) LLM-as-a-Judge. Full implementation details are provided in Appx.~\ref{app:implementation_details}.

\subsection{Main results on \textsc{RewardPrediction}}

We evaluate reward prediction accuracy using the EPIC distance ($D_{\text{EPIC}}$), where lower values indicate better alignment with ground-truth progress. Table~\ref{tab:main_results} presents a comprehensive comparison of \textsc{StateFactory} against both supervised and representation-free baselines on the \textsc{RewardPrediction} benchmark dataset.

\textbf{Supervised Reward Models.}
Table~\ref{tab:main_results} reveals the severe generalization bottleneck of supervised approaches. While these models achieve exceptional precision within their training domains (e.g., 0.212 in \texttt{AlfWorld}), they exhibit an average error increase of \textbf{138\%} when transferred to unseen tasks. This confirms that supervised critics tend to overfit domain-specific surface patterns rather than learn a generalized notion of progress, rendering them less effective for zero-shot generalization.

\textbf{Representation-free Methods.}
Training-free baselines avoid the overfitting trap but face unique challenges. Inspection of reasoning traces reveals that LLMs implicitly maintain a belief state and extract goals at each step, which improves reward prediction. However, constrained by the probabilistic nature of generation, these implicit reward signals remain discrete and volatile. While they perform slightly better in structured domains like \texttt{TextWorld} (where tasks resemble a ``checklist'') and \texttt{BlocksWorld} (suitable for reverse engineering), this implicit tracking breaks down in open-ended environments.

\textbf{Representation-based Method.}
\textsc{StateFactory} 
addresses the dual challenges of reward generalization and signal continuity. By explicitly decomposing the state space into structured object-attribute representations, our method transforms reward estimation from a discrete generation task into a continuous semantic distance calculation. The fine-grained hierarchical factorization structure not only effectively mitigates noise in flat text observations, but also enables \textsc{StateFactory} to capture subtle progresses that implicit methods could miss, thus yielding a robust estimator  that generalizes across different domains without the need of any parameter update (training). As can be observed, \textsc{StateFactory} achieves an average $D_{\text{EPIC}}$ of \textbf{0.297}, not only superior to the best representation-free baseline (0.322) but also surpassing single-domain supervised models in generalization. Remarkably, this zero-shot performance approaches the fully supervised upper bound trained on the entire dataset (by combining data across all domains).

\subsection{Ablation Studies of \textsc{StateFactory}}
\label{sec:ablation}
\textbf{Importance of Abstraction}. Transitioning from \textit{unstructured observations} to extracted \textit{textual states} yields a significant distance reduction ($0.57 \to 0.43$) in Figure~\ref{fig:abolation}(a). This confirms that raw observations contain excessive distractors, whereas state extraction isolates task-relevant information. However, unstructured text remains susceptible to \textit{score dilution}, where critical state changes are often overshadowed by the total sentence length.

\textbf{Factorization Granularity.} As shown in Figure~\ref{fig:abolation}(a), while \textit{object-centric} states mitigate dilution by isolating objects ($0.35$), they hit a precision ceiling due to \textit{attribute entanglement}. Merging distinct properties into a single embedding (e.g., ``blue mug on the table'') creates interference,  since the model cannot track the \textit{location} change clearly because the \textit{color} information acts as noise. \textsc{StateFactory} overcomes this by refining the granularity to explicit \textit{object-attribute} states. By decomposing observations into \textbf{minimal semantic units}, our method eliminates semantic interference and achieves the best performance ($0.30$).

\textbf{Goal Interpretation.} To evaluate the fidelity of our dynamic goal interpretation, we compare it against an offline (oracle) baseline. We define an offline setting where final state of the expert trajectory serves as the ground truth goal state and compare it with our online setting where goals are interpreted dynamically from interaction history. 
As shown in Figure~\ref{fig:abolation}(b), the distance between real-time inference and the offline baseline is minimal. The average $D_{EPIC}$ increases by only \textbf{0.02} when transitioning from the offline ($0.28$) to the online setting ($0.30$). This stability suggests that our goal interpretation effectively recovers target requirements.

\textbf{LLM Backbone.} We investigate the impact of LLM backbone capabilities on the state factorization, distinguishing between the inference ability (reasoning-enhanced modes) and the parametric scale. As shown in Figure~\ref{fig:abolation}(c), both dimensions closely correlate with alignment precision. First, enabling ``thinking'' modes yields significant gains: \textit{gpt-oss-20b thinking medium} improves from 0.36 to 0.30, and \textit{qwen3-30B-thinking} ($0.41$) substantially outperforms its standard counterpart ($0.55$), confirming that the extended inference capability aids in decomposing complex states. Second, we observe that performance generally improves with parameter count within the same model family, showing that reasoning ability is effectively grounded in model size. Consequently, \textsc{StateFactory} is positioned to scale continuously with future advancements in both model size and reasoning strategies.

\textbf{Semantic Embeddings.} One advantage of \textsc{StateFactory} is that any text-embedding model can be applied on top of the hierarchical object-attribute structure to compute final reward scores. Here, we investigate the correlation between the discriminative power of the embedding and the performance of \textsc{StateFactory}. This discriminative power is measured by the \textbf{triplet-based accuracy}~\citep{dumpala2024sugarcrepe++}, derived from triplets (details in Appx.~\ref{app:embedding_robustness}) comprising a state description (anchor), a semantic paraphrase (positive), and a contradiction (negative). The metric is defined as the success rate where the embedding correctly assigns higher similarity to the positive pair than to the negative one. As shown in Fig.~\ref{fig:abolation}(d) and (e), high triplet accuracy strictly correlates with performance: discriminative models (e.g., \texttt{all-MiniLM}) yield superior alignment ($D_{\text{EPIC}} \approx 0.30$), whereas weaker embedding like \texttt{gemma-300m} (acc$=0.67$) fail ($D_{\text{EPIC}}=0.45$). This finding validates that the effectiveness of \textsc{StateFactory} is driven by the discriminative quality of the embedding space.

\subsection{Utility for Agent Planning}
\label{subsec:planning}
\textbf{Quantitative Analysis.}
We posit that the superior reward prediction capabilities of \textsc{StateFactory} translate directly into enhanced agent planning performance. To verify this, we evaluate the \textsc{ReAct + StateFactory} agent, which integrates our reward signals into the standard ReAct baseline. Unlike the vanilla ReAct agent that relies solely on internal reasoning, our method augments decision-making by explicitly ranking top-$K$ action candidates.
As summarized in Table~\ref{tab:performance}, this approach yields robust and consistent gains across all distinct tested domains. Notably, the success rate in \texttt{AlfWorld} improves by 20+ percentage points ($34.33\% \to 55.97\%$), with similar steady climbs observed in \texttt{BlocksWorld} and \texttt{ScienceWorld}. Crucially, we find that explicit reward signals empower the agent to break out of reasoning ``deadlocks'', thereby successfully overcoming the ambiguity of long-horizon planning where pure logic alone typically fails.   

\begin{table}[t] 
\centering
\small
\caption{Success Rate on \texttt{AlfWorld}, \texttt{BlocksWorld}, and \texttt{ScienceWorld}.
Results show that  ReAct + \textsc{StateFactory} consistently improves performance in following domains.}
\label{tab:performance}
\resizebox{\columnwidth}{!}{
    \begin{tabular}{@{} lcc @{}} 
    \toprule
    \textbf{Domain} & \textbf{Method} & \textbf{Success Rate} (\%) \\ \midrule
    
    \multirow{2}{*}{\texttt{AlfWorld}}    & ReAct~\citep{yao2022react}   & 34.33 \\
                                 & ReAct + \textsc{StateFactory} & \textbf{55.97} \\ \midrule
    
    \multirow{2}{*}{\texttt{BlocksWorld}} & ReAct~\citep{yao2022react}   & 85.00 \\
                                 & ReAct + \textsc{StateFactory} & \textbf{93.00} \\ \midrule
    
    \multirow{2}{*}{\texttt{ScienceWorld}}& ReAct~\citep{yao2022react}   & 22.63 \\
                                 & ReAct + \textsc{StateFactory} & \textbf{35.03} \\ 
    \bottomrule
    \end{tabular}
}
\vspace{-8pt}
\end{table}

\textbf{Qualitative Analysis.}
As exemplified by the preliminary case study in Table~\ref{tab:system2planning-case1}, we further investigate whether accurate reward prediction facilitates complex planning within a system-2 framework. We implement this by integrating LLM-based action proposals, a learned world model~\citep{li2025word}, and \textsc{StateFactory} into Monte Carlo Tree Search (MCTS). In this setup, \textsc{StateFactory} acts as a structured heuristic, guiding exploration via a continuous reward signal. This formulation establishes a clear reward signal, evidenced by consistent value increases as the agent approaches targets (e.g., the ``CD''), enabling successful navigation even without sparse success signals. Crucially, this reward signal is underpinned by our dynamic goal interpretation and state extraction, which refine abstract intents into granular requirements and trigger instantaneous reward spikes upon object discovery, ensuring simulated trajectories remain grounded in physical evidence.
\section{Related Works}
\label{sec:related-works}

\textbf{Goal Conditioned Reinforcement Learning}. 
Accurately measuring task advancement is essential for policy optimization \citep{yu2025reward}. Early approaches relying on manual engineering \citep{lillicrap2015continuous, singh2019end, andrychowicz2020learning} are labor-intensive and notoriously susceptible to the reward hacking \citep{amodei2016concrete, kaufmann2024survey}. To alleviate these issues, recent methods leverage human feedback, including the demonstrations and preferences \citep{christiano2017deep, ouyang2022training, hong2024orpo, xiao2025connection}, or LLM-synthesized rewards \citep{ma2023eureka, xie2023text2reward, wang2025vagen, xue2025illusion}. However, these methods struggle to generalize to novel domains due to their dependence on task-specific supervision \citep{wolf2025reward}. 
Consequently, the field has shifted towards unsupervised paradigms that extract dense signals directly from pre-trained representations. These approaches can be broadly categorized into two streams based on how they utilize the feature space.
The first stream focuses on explicit reward modeling, where scalar signals are derived by quantifying task progress or goal probability. This includes methods such as vision-language alignment \citep{rocamonde2023vision, sontakke2023roboclip, alakuijala2024video}, discriminative classification likelihood \citep{sermanet2016unsupervised, warde2018unsupervised} or temporal distances in passive videos \citep{sermanet2018time, liu2025timerewarder}. The second stream employs latent world models for implicit reward prediction \citep{zhou2411dino, terver2025drives, sobal2025learning}. Instead of relying on explicit scorers, these methods derive signals by formulating the latent distance to the goal as an intrinsic cost, effectively transforming predictive dynamics into a dense progress indicator for planning.

\textbf{State Representation in World Models}.
The performance of language-based agent learning heavily relies on the fidelity of world representation \citep{valmeekam2023planning, wang2023describe}.
Existing approaches address this challenge with varying degrees of structural assumption. Initial works employ \textit{unstructured representations}, typically maintaining the state as raw observations \citep{yao2022react, hao2023reasoning}. However, this lack of filtering retains excessive task-irrelevant noise. This information overload obscures critical state changes, making it difficult to track the status of objects consistently. \citep{valmeekam2023planbench}. 
To better organize these observations, recent \textit{object-centric frameworks} decompose environments into discrete entities \citep{feng2025learning, yoneda2024statler}. Yet, these approaches typically treat objects as simple units, failing to separate the object's identity from its changing attributes.
Consequently, it becomes difficult to track specific changes (e.g., the rising temperature of an ingredient).
This lack of detail hinders the measurement of precise progress required for effective learning \citep{wang2024can}.
To capture these fine-grained dynamics, recent studies have explored object-attribute representations, which factorize entities into structured attribute-value pairs to support more accurate state tracking \citep{zhu2023ghost, rozanov2025stateact}.

\textbf{Language-based Agent Learning}.
Similar to physical control, the scope of sequential decision-making extends to digital environments. Here, interactions such as web navigation \citep{zhou2023webarena, deng2023mind2web}, operating system control \citep{zhang2025appagent, xie2024osworld} and text-based games \citep{cote2018textworld, hausknecht2020interactive} are formulated as control problems governed by comparable dynamics.
Traditionally, agents operate as \textit{model-free policies}, mapping observations to actions via in-context reasoning \citep{wei2022chain, yao2022react, shinn2023reflexion, park2023generative}, yet often struggle with long-horizon consistency and generalization \citep{wang2024survey, madaan2023self, wang2023learning}.
To address this, recent work adopts the \textit{Model-based RL framework}, validating LLMs as implicit world models that simulate text or code transitions \citep{zhuge2023mindstorms, li2025word}. This predictive capability enables search-based planning (e.g., MCTS) in web agents, outperforming reactive prompting by simulating future states before execution \citep{gu2411your, hao2023reasoning, yao2023tree}. Furthermore, CoSPlan~\citep{grover2025cosplan} demonstrates that incorporating incremental scene graph updates allows agents to explicitly track state transitions, enabling them to detect and rectify errors in complex visual sequences.
Concurrently, to mitigate exploration bottlenecks, the paradigm is shifting to \textit{offline reinforcement learning} \citep{kumar2020conservative, levine2020offline}. Recent Approaches \citep{zhang2025agent, nakano2021webgpt} refine policies directly from sub-optimal offline data without extensive online interaction. This leverages implicit value frameworks \citep{snell2022offline} to distill optimal behaviors from mediocre traces, a principle now extended to self-generated agent experiences \citep{chebotar2023q}.

\section{Conclusion}
\label{sec:conclusion}

In this work, we demonstrate that structured text-based world state representations alone can enable accurate and generalizable reward prediction. We introduce the \textsc{RewardPrediction} benchmark and propose \textsc{StateFactory}, a zero-shot framework that achieves superior prediction accuracy by factorizing observations into hierarchical object-attribute structures. Our results show that \textsc{StateFactory} exhibits clear advantages over supervised models and LLM-as-a-Judge baselines across diverse domains. Furthermore, our state-driven signals yield success rate improvements of up to 21.64\% for system-1 policies , while qualitative analysis validates the utility of \textsc{StateFactory} for system-2 agent planning by effectively guiding structured search grounded in physical evidence.

\sectionheaderline{
  \seclink{sec:intro}{1}{Intro} | 
  \seclink{sec:benchmark}{2}{Benchmark} | 
  \seclink{sec:methods}{3}{Method} |
  \seclink{sec:experiments}{4}{Experiments} |
  \seclink{sec:related-works}{5}{Related Works} |
  \textbf{\hyperref[sec:conclusion]{Sec 6: Conclusion}} |
  \hyperref[sec:appendix]{Sec 7: Appx}
}

\bibliography{bibliography}

\begin{thebibliography}{97}
\providecommand{\natexlab}[1]{#1}
\providecommand{\url}[1]{\texttt{#1}}
\expandafter\ifx\csname urlstyle\endcsname\relax
  \providecommand{\doi}[1]{doi: #1}\else
  \providecommand{\doi}{doi: \begingroup \urlstyle{rm}\Url}\fi

\bibitem[Alakuijala et~al.(2024)Alakuijala, McLean, Woungang, Farsad, Kaski, Marttinen, and Yuan]{alakuijala2024video}
Minttu Alakuijala, Reginald McLean, Isaac Woungang, Nariman Farsad, Samuel Kaski, Pekka Marttinen, and Kai Yuan.
\newblock Video-language critic: Transferable reward functions for language-conditioned robotics.
\newblock \emph{arXiv preprint arXiv:2405.19988}, 2024.

\bibitem[Amodei et~al.(2016)Amodei, Olah, Steinhardt, Christiano, Schulman, and Man{\'e}]{amodei2016concrete}
Dario Amodei, Chris Olah, Jacob Steinhardt, Paul Christiano, John Schulman, and Dan Man{\'e}.
\newblock Concrete problems in ai safety.
\newblock \emph{arXiv preprint arXiv:1606.06565}, 2016.

\bibitem[Andrychowicz et~al.(2017)Andrychowicz, Wolski, Ray, Schneider, Fong, Welinder, McGrew, Tobin, Pieter~Abbeel, and Zaremba]{andrychowicz2017hindsight}
Marcin Andrychowicz, Filip Wolski, Alex Ray, Jonas Schneider, Rachel Fong, Peter Welinder, Bob McGrew, Josh Tobin, OpenAI Pieter~Abbeel, and Wojciech Zaremba.
\newblock Hindsight experience replay.
\newblock \emph{Advances in neural information processing systems}, 30, 2017.

\bibitem[Andrychowicz et~al.(2020)Andrychowicz, Baker, Chociej, Jozefowicz, McGrew, Pachocki, Petron, Plappert, Powell, Ray, et~al.]{andrychowicz2020learning}
OpenAI:~Marcin Andrychowicz, Bowen Baker, Maciek Chociej, Rafal Jozefowicz, Bob McGrew, Jakub Pachocki, Arthur Petron, Matthias Plappert, Glenn Powell, Alex Ray, et~al.
\newblock Learning dexterous in-hand manipulation.
\newblock \emph{The International Journal of Robotics Research}, 39\penalty0 (1):\penalty0 3--20, 2020.

\bibitem[Cao et~al.(2024)Cao, Zhao, Cheng, Shu, Chen, Liu, Liang, Zhao, Yan, and Li]{cao2024survey}
Yuji Cao, Huan Zhao, Yuheng Cheng, Ting Shu, Yue Chen, Guolong Liu, Gaoqi Liang, Junhua Zhao, Jinyue Yan, and Yun Li.
\newblock Survey on large language model-enhanced reinforcement learning: Concept, taxonomy, and methods.
\newblock \emph{IEEE Transactions on Neural Networks and Learning Systems}, 2024.

\bibitem[Casper et~al.(2023)Casper, Davies, Shi, Gilbert, Scheurer, Rando, Freedman, Korbak, Lindner, Freire, et~al.]{casper2023open}
Stephen Casper, Xander Davies, Claudia Shi, Thomas~Krendl Gilbert, J{\'e}r{\'e}my Scheurer, Javier Rando, Rachel Freedman, Tomasz Korbak, David Lindner, Pedro Freire, et~al.
\newblock Open problems and fundamental limitations of reinforcement learning from human feedback.
\newblock \emph{arXiv preprint arXiv:2307.15217}, 2023.

\bibitem[Chebotar et~al.(2023)Chebotar, Vuong, Hausman, Xia, Lu, Irpan, Kumar, Yu, Herzog, Pertsch, et~al.]{chebotar2023q}
Yevgen Chebotar, Quan Vuong, Karol Hausman, Fei Xia, Yao Lu, Alex Irpan, Aviral Kumar, Tianhe Yu, Alexander Herzog, Karl Pertsch, et~al.
\newblock Q-transformer: Scalable offline reinforcement learning via autoregressive q-functions.
\newblock In \emph{Conference on Robot Learning}, pages 3909--3928. PMLR, 2023.

\bibitem[Chen et~al.(2025)Chen, Moutakanni, Chung, Bang, Ji, Bolourchi, and Fung]{chen2025planning}
Delong Chen, Theo Moutakanni, Willy Chung, Yejin Bang, Ziwei Ji, Allen Bolourchi, and Pascale Fung.
\newblock Planning with reasoning using vision language world model.
\newblock \emph{arXiv preprint arXiv:2509.02722}, 2025.

\bibitem[Christiano et~al.(2017)Christiano, Leike, Brown, Martic, Legg, and Amodei]{christiano2017deep}
Paul~F Christiano, Jan Leike, Tom Brown, Miljan Martic, Shane Legg, and Dario Amodei.
\newblock Deep reinforcement learning from human preferences.
\newblock \emph{Advances in neural information processing systems}, 30, 2017.

\bibitem[C{\^o}t{\'e} et~al.(2018)C{\^o}t{\'e}, K{\'a}d{\'a}r, Yuan, Kybartas, Barnes, Fine, Moore, Hausknecht, El~Asri, Adada, et~al.]{cote2018textworld}
Marc-Alexandre C{\^o}t{\'e}, Akos K{\'a}d{\'a}r, Xingdi Yuan, Ben Kybartas, Tavian Barnes, Emery Fine, James Moore, Matthew Hausknecht, Layla El~Asri, Mahmoud Adada, et~al.
\newblock Textworld: A learning environment for text-based games.
\newblock In \emph{Workshop on Computer Games}, pages 41--75. Springer, 2018.

\bibitem[Deng et~al.(2023)Deng, Gu, Zheng, Chen, Stevens, Wang, Sun, and Su]{deng2023mind2web}
Xiang Deng, Yu~Gu, Boyuan Zheng, Shijie Chen, Sam Stevens, Boshi Wang, Huan Sun, and Yu~Su.
\newblock Mind2web: Towards a generalist agent for the web.
\newblock \emph{Advances in Neural Information Processing Systems}, 36:\penalty0 28091--28114, 2023.

\bibitem[Di~Langosco et~al.(2022)Di~Langosco, Koch, Sharkey, Pfau, and Krueger]{di2022goal}
Lauro~Langosco Di~Langosco, Jack Koch, Lee~D Sharkey, Jacob Pfau, and David Krueger.
\newblock Goal misgeneralization in deep reinforcement learning.
\newblock In \emph{International Conference on Machine Learning}, pages 12004--12019. PMLR, 2022.

\bibitem[Dumpala et~al.(2024)Dumpala, Jaiswal, Shama~Sastry, Milios, Oore, and Sajjad]{dumpala2024sugarcrepe++}
Sri~Harsha Dumpala, Aman Jaiswal, Chandramouli Shama~Sastry, Evangelos Milios, Sageev Oore, and Hassan Sajjad.
\newblock Sugarcrepe++ dataset: Vision-language model sensitivity to semantic and lexical alterations.
\newblock \emph{Advances in Neural Information Processing Systems}, 37:\penalty0 17972--18018, 2024.

\bibitem[Feng et~al.(2025)Feng, Lippe, and Magliacane]{feng2025learning}
Fan Feng, Phillip Lippe, and Sara Magliacane.
\newblock Learning interactive world model for object-centric reinforcement learning.
\newblock \emph{arXiv preprint arXiv:2511.02225}, 2025.

\bibitem[Frick et~al.(2024)Frick, Li, Chen, Chiang, Angelopoulos, Jiao, Zhu, Gonzalez, and Stoica]{frick2024evaluate}
Evan Frick, Tianle Li, Connor Chen, Wei-Lin Chiang, Anastasios~N Angelopoulos, Jiantao Jiao, Banghua Zhu, Joseph~E Gonzalez, and Ion Stoica.
\newblock How to evaluate reward models for rlhf.
\newblock \emph{arXiv preprint arXiv:2410.14872}, 2024.

\bibitem[Gao et~al.(2023)Gao, Schulman, and Hilton]{gao2023scaling}
Leo Gao, John Schulman, and Jacob Hilton.
\newblock Scaling laws for reward model overoptimization.
\newblock In \emph{International Conference on Machine Learning}, pages 10835--10866. PMLR, 2023.

\bibitem[Gleave et~al.(2020)Gleave, Dennis, Legg, Russell, and Leike]{gleave2020quantifying}
Adam Gleave, Michael Dennis, Shane Legg, Stuart Russell, and Jan Leike.
\newblock Quantifying differences in reward functions.
\newblock \emph{arXiv preprint arXiv:2006.13900}, 2020.

\bibitem[Grover et~al.(2025)Grover, Pathak, Kumar, Vineet, and Rawat]{grover2025cosplan}
Shresth Grover, Priyank Pathak, Akash Kumar, Vibhav Vineet, and Yogesh~S Rawat.
\newblock Cosplan: Corrective sequential planning via scene graph incremental updates.
\newblock \emph{arXiv preprint arXiv:2512.10342}, 2025.

\bibitem[Gu et~al.(2024)Gu, Zhang, Ning, Zheng, Gou, Xue, Chang, Srivastava, Xie, Qi, et~al.]{gu2411your}
Yu~Gu, Kai Zhang, Yuting Ning, Boyuan Zheng, Boyu Gou, Tianci Xue, Cheng Chang, Sanjari Srivastava, Yanan Xie, Peng Qi, et~al.
\newblock Is your llm secretly a world model of the internet? model-based planning for web agents.
\newblock \emph{arXiv preprint arXiv:2411.06559}, 2024.

\bibitem[Gunjal et~al.(2025)Gunjal, Wang, Lau, Nath, He, Liu, and Hendryx]{gunjal2025rubrics}
Anisha Gunjal, Anthony Wang, Elaine Lau, Vaskar Nath, Yunzhong He, Bing Liu, and Sean Hendryx.
\newblock Rubrics as rewards: Reinforcement learning beyond verifiable domains.
\newblock \emph{arXiv preprint arXiv:2507.17746}, 2025.

\bibitem[Guo et~al.(2025{\natexlab{a}})Guo, Yang, Zhang, Song, Zhang, Xu, Zhu, Ma, Wang, Bi, et~al.]{guo2025deepseek}
Daya Guo, Dejian Yang, Haowei Zhang, Junxiao Song, Ruoyu Zhang, Runxin Xu, Qihao Zhu, Shirong Ma, Peiyi Wang, Xiao Bi, et~al.
\newblock Deepseek-r1: Incentivizing reasoning capability in llms via reinforcement learning.
\newblock \emph{arXiv preprint arXiv:2501.12948}, 2025{\natexlab{a}}.

\bibitem[Guo et~al.(2025{\natexlab{b}})Guo, Chi, Dong, Dong, Wu, Huang, and Wei]{guo2025reward}
Jiaxin Guo, Zewen Chi, Li~Dong, Qingxiu Dong, Xun Wu, Shaohan Huang, and Furu Wei.
\newblock Reward reasoning model.
\newblock \emph{arXiv preprint arXiv:2505.14674}, 2025{\natexlab{b}}.

\bibitem[Ha and Schmidhuber(2018)]{ha2018world}
David Ha and J{\"u}rgen Schmidhuber.
\newblock World models.
\newblock \emph{arXiv preprint arXiv:1803.10122}, 2\penalty0 (3), 2018.

\bibitem[Hafner et~al.(2023)Hafner, Pasukonis, Ba, and Lillicrap]{hafner2023mastering}
Danijar Hafner, Jurgis Pasukonis, Jimmy Ba, and Timothy Lillicrap.
\newblock Mastering diverse domains through world models.
\newblock \emph{arXiv preprint arXiv:2301.04104}, 2023.

\bibitem[Hansen et~al.(2023)Hansen, Su, and Wang]{hansen2023td}
Nicklas Hansen, Hao Su, and Xiaolong Wang.
\newblock Td-mpc2: Scalable, robust world models for continuous control.
\newblock \emph{arXiv preprint arXiv:2310.16828}, 2023.

\bibitem[Hao et~al.(2023)Hao, Gu, Ma, Hong, Wang, Wang, and Hu]{hao2023reasoning}
Shibo Hao, Yi~Gu, Haodi Ma, Joshua Hong, Zhen Wang, Daisy Wang, and Zhiting Hu.
\newblock Reasoning with language model is planning with world model.
\newblock In \emph{Proceedings of the 2023 Conference on Empirical Methods in Natural Language Processing}, pages 8154--8173, 2023.

\bibitem[Hausknecht et~al.(2020)Hausknecht, Ammanabrolu, C{\^o}t{\'e}, and Yuan]{hausknecht2020interactive}
Matthew Hausknecht, Prithviraj Ammanabrolu, Marc-Alexandre C{\^o}t{\'e}, and Xingdi Yuan.
\newblock Interactive fiction games: A colossal adventure.
\newblock In \emph{Proceedings of the AAAI Conference on Artificial Intelligence}, volume~34, pages 7903--7910, 2020.

\bibitem[Hong et~al.(2024)Hong, Lee, and Thorne]{hong2024orpo}
Jiwoo Hong, Noah Lee, and James Thorne.
\newblock Orpo: Monolithic preference optimization without reference model.
\newblock \emph{arXiv preprint arXiv:2403.07691}, 2024.

\bibitem[Hu et~al.(2022)Hu, Shen, Wallis, Allen-Zhu, Li, Wang, Wang, Chen, et~al.]{hu2022lora}
Edward~J Hu, Yelong Shen, Phillip Wallis, Zeyuan Allen-Zhu, Yuanzhi Li, Shean Wang, Lu~Wang, Weizhu Chen, et~al.
\newblock Lora: Low-rank adaptation of large language models.
\newblock \emph{ICLR}, 1\penalty0 (2):\penalty0 3, 2022.

\bibitem[Hui et~al.(2024)Hui, Yang, Cui, Yang, Liu, Zhang, Liu, Zhang, Yu, Lu, et~al.]{hui2024qwen2}
Binyuan Hui, Jian Yang, Zeyu Cui, Jiaxi Yang, Dayiheng Liu, Lei Zhang, Tianyu Liu, Jiajun Zhang, Bowen Yu, Keming Lu, et~al.
\newblock Qwen2. 5-coder technical report.
\newblock \emph{arXiv preprint arXiv:2409.12186}, 2024.

\bibitem[Kaelbling et~al.(1998)Kaelbling, Littman, and Cassandra]{kaelbling1998planning}
Leslie~Pack Kaelbling, Michael~L Littman, and Anthony~R Cassandra.
\newblock Planning and acting in partially observable stochastic domains.
\newblock \emph{Artificial intelligence}, 101\penalty0 (1-2):\penalty0 99--134, 1998.

\bibitem[Kaufmann et~al.(2024)Kaufmann, Weng, Bengs, and H{\"u}llermeier]{kaufmann2024survey}
Timo Kaufmann, Paul Weng, Viktor Bengs, and Eyke H{\"u}llermeier.
\newblock A survey of reinforcement learning from human feedback.
\newblock 2024.

\bibitem[Kumar et~al.(2020)Kumar, Zhou, Tucker, and Levine]{kumar2020conservative}
Aviral Kumar, Aurick Zhou, George Tucker, and Sergey Levine.
\newblock Conservative q-learning for offline reinforcement learning.
\newblock \emph{Advances in neural information processing systems}, 33:\penalty0 1179--1191, 2020.

\bibitem[LeCun(2022)]{lecun2022path}
Yann LeCun.
\newblock A path towards autonomous machine intelligence version 0.9. 2, 2022-06-27.
\newblock \emph{Open Review}, 62\penalty0 (1):\penalty0 1--62, 2022.

\bibitem[Levine et~al.(2020)Levine, Kumar, Tucker, and Fu]{levine2020offline}
Sergey Levine, Aviral Kumar, George Tucker, and Justin Fu.
\newblock Offline reinforcement learning: Tutorial, review, and perspectives on open problems.
\newblock \emph{arXiv preprint arXiv:2005.01643}, 2020.

\bibitem[Li et~al.(2025)Li, Wang, Qiu, Yin, Zhang, Qian, Li, Ma, Chen, Ji, et~al.]{li2025word}
Yixia Li, Hongru Wang, Jiahao Qiu, Zhenfei Yin, Dongdong Zhang, Cheng Qian, Zeping Li, Pony Ma, Guanhua Chen, Heng Ji, et~al.
\newblock From word to world: Can large language models be implicit text-based world models?
\newblock \emph{arXiv preprint arXiv:2512.18832}, 2025.

\bibitem[Lillicrap et~al.(2015)Lillicrap, Hunt, Pritzel, Heess, Erez, Tassa, Silver, and Wierstra]{lillicrap2015continuous}
Timothy~P Lillicrap, Jonathan~J Hunt, Alexander Pritzel, Nicolas Heess, Tom Erez, Yuval Tassa, David Silver, and Daan Wierstra.
\newblock Continuous control with deep reinforcement learning.
\newblock \emph{arXiv preprint arXiv:1509.02971}, 2015.

\bibitem[Liu et~al.(2022)Liu, Zhu, and Zhang]{liu2022goal}
Minghuan Liu, Menghui Zhu, and Weinan Zhang.
\newblock Goal-conditioned reinforcement learning: Problems and solutions.
\newblock \emph{arXiv preprint arXiv:2201.08299}, 2022.

\bibitem[Liu et~al.(2025)Liu, Wen, Hu, Jayaraman, and Gao]{liu2025timerewarder}
Yuyang Liu, Chuan Wen, Yihang Hu, Dinesh Jayaraman, and Yang Gao.
\newblock Timerewarder: Learning dense reward from passive videos via frame-wise temporal distance.
\newblock \emph{arXiv preprint arXiv:2509.26627}, 2025.

\bibitem[Luketina et~al.(2019)Luketina, Nardelli, Farquhar, Foerster, Andreas, Grefenstette, Whiteson, and Rockt{\"a}schel]{luketina2019survey}
Jelena Luketina, Nantas Nardelli, Gregory Farquhar, Jakob Foerster, Jacob Andreas, Edward Grefenstette, Shimon Whiteson, and Tim Rockt{\"a}schel.
\newblock A survey of reinforcement learning informed by natural language.
\newblock \emph{arXiv preprint arXiv:1906.03926}, 2019.

\bibitem[M.~Moerland et~al.(2023)M.~Moerland, Broekens, Plaat, and M.~Jonker]{m2023model}
Thomas M.~Moerland, Joost Broekens, Aske Plaat, and Catholijn M.~Jonker.
\newblock Model-based reinforcement learning: A survey.
\newblock \emph{Foundations and Trends in Machine Learning}, 16\penalty0 (1):\penalty0 1--118, 2023.

\bibitem[Ma et~al.(2023{\natexlab{a}})Ma, Kumar, Zhang, Bastani, and Jayaraman]{ma2023liv}
Yecheng~Jason Ma, Vikash Kumar, Amy Zhang, Osbert Bastani, and Dinesh Jayaraman.
\newblock Liv: Language-image representations and rewards for robotic control.
\newblock In \emph{International Conference on Machine Learning}, pages 23301--23320. PMLR, 2023{\natexlab{a}}.

\bibitem[Ma et~al.(2023{\natexlab{b}})Ma, Liang, Wang, Huang, Bastani, Jayaraman, Zhu, Fan, and Anandkumar]{ma2023eureka}
Yecheng~Jason Ma, William Liang, Guanzhi Wang, De-An Huang, Osbert Bastani, Dinesh Jayaraman, Yuke Zhu, Linxi Fan, and Anima Anandkumar.
\newblock Eureka: Human-level reward design via coding large language models.
\newblock \emph{arXiv preprint arXiv:2310.12931}, 2023{\natexlab{b}}.

\bibitem[Madaan et~al.(2023)Madaan, Tandon, Gupta, Hallinan, Gao, Wiegreffe, Alon, Dziri, Prabhumoye, Yang, et~al.]{madaan2023self}
Aman Madaan, Niket Tandon, Prakhar Gupta, Skyler Hallinan, Luyu Gao, Sarah Wiegreffe, Uri Alon, Nouha Dziri, Shrimai Prabhumoye, Yiming Yang, et~al.
\newblock Self-refine: Iterative refinement with self-feedback.
\newblock \emph{Advances in Neural Information Processing Systems}, 36:\penalty0 46534--46594, 2023.

\bibitem[Mahmoudieh et~al.(2022)Mahmoudieh, Pathak, and Darrell]{mahmoudieh2022zero}
Parsa Mahmoudieh, Deepak Pathak, and Trevor Darrell.
\newblock Zero-shot reward specification via grounded natural language.
\newblock In \emph{International Conference on Machine Learning}, pages 14743--14752. PMLR, 2022.

\bibitem[Naik et~al.(2024)Naik, Wan, Tomar, and Sutton]{naik2024reward}
Abhishek Naik, Yi~Wan, Manan Tomar, and Richard~S Sutton.
\newblock Reward centering.
\newblock \emph{arXiv preprint arXiv:2405.09999}, 2024.

\bibitem[Nakano et~al.(2021)Nakano, Hilton, Balaji, Wu, Ouyang, Kim, Hesse, Jain, Kosaraju, Saunders, et~al.]{nakano2021webgpt}
Reiichiro Nakano, Jacob Hilton, Suchir Balaji, Jeff Wu, Long Ouyang, Christina Kim, Christopher Hesse, Shantanu Jain, Vineet Kosaraju, William Saunders, et~al.
\newblock Webgpt: Browser-assisted question-answering with human feedback.
\newblock \emph{arXiv preprint arXiv:2112.09332}, 2021.

\bibitem[Ouyang et~al.(2022)Ouyang, Wu, Jiang, Almeida, Wainwright, Mishkin, Zhang, Agarwal, Slama, Ray, et~al.]{ouyang2022training}
Long Ouyang, Jeffrey Wu, Xu~Jiang, Diogo Almeida, Carroll Wainwright, Pamela Mishkin, Chong Zhang, Sandhini Agarwal, Katarina Slama, Alex Ray, et~al.
\newblock Training language models to follow instructions with human feedback.
\newblock \emph{Advances in neural information processing systems}, 35:\penalty0 27730--27744, 2022.

\bibitem[Park et~al.(2023)Park, O'Brien, Cai, Morris, Liang, and Bernstein]{park2023generative}
Joon~Sung Park, Joseph O'Brien, Carrie~Jun Cai, Meredith~Ringel Morris, Percy Liang, and Michael~S Bernstein.
\newblock Generative agents: Interactive simulacra of human behavior.
\newblock In \emph{Proceedings of the 36th annual acm symposium on user interface software and technology}, pages 1--22, 2023.

\bibitem[Richens et~al.(2025)Richens, Everitt, and Abel]{richens2025general}
Jonathan Richens, Tom Everitt, and David Abel.
\newblock General agents need world models.
\newblock In \emph{Forty-second International Conference on Machine Learning}, 2025.

\bibitem[Rocamonde et~al.(2023)Rocamonde, Montesinos, Nava, Perez, and Lindner]{rocamonde2023vision}
Juan Rocamonde, Victoriano Montesinos, Elvis Nava, Ethan Perez, and David Lindner.
\newblock Vision-language models are zero-shot reward models for reinforcement learning.
\newblock \emph{arXiv preprint arXiv:2310.12921}, 2023.

\bibitem[Rozanov and Rei(2025)]{rozanov2025stateact}
Nikolai Rozanov and Marek Rei.
\newblock Stateact: Enhancing llm base agents via self-prompting and state-tracking.
\newblock In \emph{Proceedings of the 1st Workshop for Research on Agent Language Models (REALM 2025)}, pages 367--385, 2025.

\bibitem[Schaul et~al.(2015)Schaul, Horgan, Gregor, and Silver]{schaul2015universal}
Tom Schaul, Daniel Horgan, Karol Gregor, and David Silver.
\newblock Universal value function approximators.
\newblock In \emph{International conference on machine learning}, pages 1312--1320. PMLR, 2015.

\bibitem[Sermanet et~al.(2016)Sermanet, Xu, and Levine]{sermanet2016unsupervised}
Pierre Sermanet, Kelvin Xu, and Sergey Levine.
\newblock Unsupervised perceptual rewards for imitation learning.
\newblock \emph{arXiv preprint arXiv:1612.06699}, 2016.

\bibitem[Sermanet et~al.(2018)Sermanet, Lynch, Chebotar, Hsu, Jang, Schaal, Levine, and Brain]{sermanet2018time}
Pierre Sermanet, Corey Lynch, Yevgen Chebotar, Jasmine Hsu, Eric Jang, Stefan Schaal, Sergey Levine, and Google Brain.
\newblock Time-contrastive networks: Self-supervised learning from video.
\newblock In \emph{2018 IEEE international conference on robotics and automation (ICRA)}, pages 1134--1141. IEEE, 2018.

\bibitem[Sharma et~al.(2023)Sharma, Tong, Korbak, Duvenaud, Askell, Bowman, Cheng, Durmus, Hatfield-Dodds, Johnston, et~al.]{sharma2023towards}
Mrinank Sharma, Meg Tong, Tomasz Korbak, David Duvenaud, Amanda Askell, Samuel~R Bowman, Newton Cheng, Esin Durmus, Zac Hatfield-Dodds, Scott~R Johnston, et~al.
\newblock Towards understanding sycophancy in language models.
\newblock \emph{arXiv preprint arXiv:2310.13548}, 2023.

\bibitem[Shinn et~al.(2023)Shinn, Cassano, Gopinath, Narasimhan, and Yao]{shinn2023reflexion}
Noah Shinn, Federico Cassano, Ashwin Gopinath, Karthik Narasimhan, and Shunyu Yao.
\newblock Reflexion: Language agents with verbal reinforcement learning.
\newblock \emph{Advances in Neural Information Processing Systems}, 36:\penalty0 8634--8652, 2023.

\bibitem[Shridhar et~al.(2020)Shridhar, Yuan, C{\^o}t{\'e}, Bisk, Trischler, and Hausknecht]{shridhar2020alfworld}
Mohit Shridhar, Xingdi Yuan, Marc-Alexandre C{\^o}t{\'e}, Yonatan Bisk, Adam Trischler, and Matthew Hausknecht.
\newblock Alfworld: Aligning text and embodied environments for interactive learning.
\newblock \emph{arXiv preprint arXiv:2010.03768}, 2020.

\bibitem[Singh et~al.(2019)Singh, Yang, Hartikainen, Finn, and Levine]{singh2019end}
Avi Singh, Larry Yang, Kristian Hartikainen, Chelsea Finn, and Sergey Levine.
\newblock End-to-end robotic reinforcement learning without reward engineering.
\newblock \emph{arXiv preprint arXiv:1904.07854}, 2019.

\bibitem[Snell et~al.(2022)Snell, Kostrikov, Su, Yang, and Levine]{snell2022offline}
Charlie Snell, Ilya Kostrikov, Yi~Su, Mengjiao Yang, and Sergey Levine.
\newblock Offline rl for natural language generation with implicit language q learning.
\newblock \emph{arXiv preprint arXiv:2206.11871}, 2022.

\bibitem[Sobal et~al.(2025)Sobal, Zhang, Cho, Balestriero, Rudner, and LeCun]{sobal2025learning}
Vlad Sobal, Wancong Zhang, Kyunghyun Cho, Randall Balestriero, Tim~GJ Rudner, and Yann LeCun.
\newblock Learning from reward-free offline data: A case for planning with latent dynamics models.
\newblock \emph{arXiv preprint arXiv:2502.14819}, 2025.

\bibitem[Sontakke et~al.(2023)Sontakke, Zhang, Arnold, Pertsch, B{\i}y{\i}k, Sadigh, Finn, and Itti]{sontakke2023roboclip}
Sumedh Sontakke, Jesse Zhang, S{\'e}b Arnold, Karl Pertsch, Erdem B{\i}y{\i}k, Dorsa Sadigh, Chelsea Finn, and Laurent Itti.
\newblock Roboclip: One demonstration is enough to learn robot policies.
\newblock \emph{Advances in Neural Information Processing Systems}, 36:\penalty0 55681--55693, 2023.

\bibitem[Sutton(1998)]{sutton1998between}
Richard~S Sutton.
\newblock Between mdps and semi-mdps: Learning, planning, and representing knowledge at multiple temporal scales.
\newblock 1998.

\bibitem[Terver et~al.(2025)Terver, Yang, Ponce, Bardes, and LeCun]{terver2025drives}
Basile Terver, Tsung-Yen Yang, Jean Ponce, Adrien Bardes, and Yann LeCun.
\newblock What drives success in physical planning with joint-embedding predictive world models?
\newblock \emph{arXiv preprint arXiv:2512.24497}, 2025.

\bibitem[Trott et~al.(2019)Trott, Zheng, Xiong, and Socher]{trott2019keeping}
Alexander Trott, Stephan Zheng, Caiming Xiong, and Richard Socher.
\newblock Keeping your distance: Solving sparse reward tasks using self-balancing shaped rewards.
\newblock \emph{Advances in Neural Information Processing Systems}, 32, 2019.

\bibitem[Valmeekam et~al.(2023{\natexlab{a}})Valmeekam, Marquez, Olmo, Sreedharan, and Kambhampati]{valmeekam2023planbench}
Karthik Valmeekam, Matthew Marquez, Alberto Olmo, Sarath Sreedharan, and Subbarao Kambhampati.
\newblock Planbench: An extensible benchmark for evaluating large language models on planning and reasoning about change.
\newblock \emph{Advances in Neural Information Processing Systems}, 36:\penalty0 38975--38987, 2023{\natexlab{a}}.

\bibitem[Valmeekam et~al.(2023{\natexlab{b}})Valmeekam, Sreedharan, Marquez, Olmo, and Kambhampati]{valmeekam2023planning}
Karthik Valmeekam, Sarath Sreedharan, Matthew Marquez, Alberto Olmo, and Subbarao Kambhampati.
\newblock On the planning abilities of large language models (a critical investigation with a proposed benchmark).
\newblock \emph{arXiv preprint arXiv:2302.06706}, 2023{\natexlab{b}}.

\bibitem[Wang and Li(2023)]{wang2023learning}
Danqing Wang and Lei Li.
\newblock Learning from mistakes via cooperative study assistant for large language models.
\newblock \emph{arXiv preprint arXiv:2305.13829}, 2023.

\bibitem[Wang et~al.(2025)Wang, Zhang, Wang, Gao, Li, Wang, Chen, Wan, Lu, Yang, et~al.]{wang2025vagen}
Kangrui Wang, Pingyue Zhang, Zihan Wang, Yaning Gao, Linjie Li, Qineng Wang, Hanyang Chen, Chi Wan, Yiping Lu, Zhengyuan Yang, et~al.
\newblock Vagen: Reinforcing world model reasoning for multi-turn vlm agents.
\newblock \emph{arXiv preprint arXiv:2510.16907}, 2025.

\bibitem[Wang et~al.(2024{\natexlab{a}})Wang, Ma, Feng, Zhang, Yang, Zhang, Chen, Tang, Chen, Lin, et~al.]{wang2024survey}
Lei Wang, Chen Ma, Xueyang Feng, Zeyu Zhang, Hao Yang, Jingsen Zhang, Zhiyuan Chen, Jiakai Tang, Xu~Chen, Yankai Lin, et~al.
\newblock A survey on large language model based autonomous agents.
\newblock \emph{Frontiers of Computer Science}, 18\penalty0 (6):\penalty0 186345, 2024{\natexlab{a}}.

\bibitem[Wang et~al.(2022)Wang, Jansen, C{\^o}t{\'e}, and Ammanabrolu]{wang2022scienceworld}
Ruoyao Wang, Peter Jansen, Marc-Alexandre C{\^o}t{\'e}, and Prithviraj Ammanabrolu.
\newblock Scienceworld: Is your agent smarter than a 5th grader?
\newblock \emph{arXiv preprint arXiv:2203.07540}, 2022.

\bibitem[Wang et~al.(2024{\natexlab{b}})Wang, Todd, Xiao, Yuan, C{\^o}t{\'e}, Clark, and Jansen]{wang2024can}
Ruoyao Wang, Graham Todd, Ziang Xiao, Xingdi Yuan, Marc-Alexandre C{\^o}t{\'e}, Peter Clark, and Peter Jansen.
\newblock Can language models serve as text-based world simulators?
\newblock \emph{arXiv preprint arXiv:2406.06485}, 2024{\natexlab{b}}.

\bibitem[Wang et~al.(2023{\natexlab{a}})Wang, Torralba, Isola, and Zhang]{wang2023optimal}
Tongzhou Wang, Antonio Torralba, Phillip Isola, and Amy Zhang.
\newblock Optimal goal-reaching reinforcement learning via quasimetric learning.
\newblock In \emph{International Conference on Machine Learning}, pages 36411--36430. PMLR, 2023{\natexlab{a}}.

\bibitem[Wang et~al.(2023{\natexlab{b}})Wang, Cai, Chen, Liu, Ma, and Liang]{wang2023describe}
Zihao Wang, Shaofei Cai, Guanzhou Chen, Anji Liu, Xiaojian Ma, and Yitao Liang.
\newblock Describe, explain, plan and select: Interactive planning with large language models enables open-world multi-task agents.
\newblock \emph{arXiv preprint arXiv:2302.01560}, 2023{\natexlab{b}}.

\bibitem[Warde-Farley et~al.(2018)Warde-Farley, Van~de Wiele, Kulkarni, Ionescu, Hansen, and Mnih]{warde2018unsupervised}
David Warde-Farley, Tom Van~de Wiele, Tejas Kulkarni, Catalin Ionescu, Steven Hansen, and Volodymyr Mnih.
\newblock Unsupervised control through non-parametric discriminative rewards.
\newblock \emph{arXiv preprint arXiv:1811.11359}, 2018.

\bibitem[Wei et~al.(2022)Wei, Wang, Schuurmans, Bosma, Xia, Chi, Le, Zhou, et~al.]{wei2022chain}
Jason Wei, Xuezhi Wang, Dale Schuurmans, Maarten Bosma, Fei Xia, Ed~Chi, Quoc~V Le, Denny Zhou, et~al.
\newblock Chain-of-thought prompting elicits reasoning in large language models.
\newblock \emph{Advances in neural information processing systems}, 35:\penalty0 24824--24837, 2022.

\bibitem[Whitehouse et~al.(2025)Whitehouse, Wang, Yu, Li, Weston, Kulikov, and Saha]{whitehouse2025j1}
Chenxi Whitehouse, Tianlu Wang, Ping Yu, Xian Li, Jason Weston, Ilia Kulikov, and Swarnadeep Saha.
\newblock J1: Incentivizing thinking in llm-as-a-judge via reinforcement learning.
\newblock \emph{arXiv preprint arXiv:2505.10320}, 2025.

\bibitem[Wolf et~al.(2025)Wolf, Kirk, and Musolesi]{wolf2025reward}
Lorenz Wolf, Robert Kirk, and Mirco Musolesi.
\newblock Reward model overoptimisation in iterated rlhf.
\newblock \emph{arXiv preprint arXiv:2505.18126}, 2025.

\bibitem[Wu et~al.(2018)Wu, Tucker, and Nachum]{wu2018laplacian}
Yifan Wu, George Tucker, and Ofir Nachum.
\newblock The laplacian in rl: Learning representations with efficient approximations.
\newblock \emph{arXiv preprint arXiv:1810.04586}, 2018.

\bibitem[Xiao et~al.(2025)Xiao, Yuan, Li, Chen, and Honavar]{xiao2025connection}
Teng Xiao, Yige Yuan, Mingxiao Li, Zhengyu Chen, and Vasant~G Honavar.
\newblock On a connection between imitation learning and rlhf.
\newblock \emph{arXiv preprint arXiv:2503.05079}, 2025.

\bibitem[Xie et~al.(2023)Xie, Zhao, Wu, Liu, Luo, Zhong, Yang, and Yu]{xie2023text2reward}
Tianbao Xie, Siheng Zhao, Chen~Henry Wu, Yitao Liu, Qian Luo, Victor Zhong, Yanchao Yang, and Tao Yu.
\newblock Text2reward: Reward shaping with language models for reinforcement learning.
\newblock \emph{arXiv preprint arXiv:2309.11489}, 2023.

\bibitem[Xie et~al.(2024)Xie, Zhang, Chen, Li, Zhao, Cao, Hua, Cheng, Shin, Lei, et~al.]{xie2024osworld}
Tianbao Xie, Danyang Zhang, Jixuan Chen, Xiaochuan Li, Siheng Zhao, Ruisheng Cao, Toh~J Hua, Zhoujun Cheng, Dongchan Shin, Fangyu Lei, et~al.
\newblock Osworld: Benchmarking multimodal agents for open-ended tasks in real computer environments.
\newblock \emph{Advances in Neural Information Processing Systems}, 37:\penalty0 52040--52094, 2024.

\bibitem[Xue et~al.(2025)Xue, Qi, Shi, Song, Gou, Song, Sun, and Su]{xue2025illusion}
Tianci Xue, Weijian Qi, Tianneng Shi, Chan~Hee Song, Boyu Gou, Dawn Song, Huan Sun, and Yu~Su.
\newblock An illusion of progress? assessing the current state of web agents.
\newblock \emph{arXiv preprint arXiv:2504.01382}, 2025.

\bibitem[Yang et~al.(2024)Yang, Tjia, Berg, Damen, Agrawal, and Gupta]{yang2024rank2reward}
Daniel Yang, Davin Tjia, Jacob Berg, Dima Damen, Pulkit Agrawal, and Abhishek Gupta.
\newblock Rank2reward: Learning shaped reward functions from passive video.
\newblock In \emph{2024 IEEE International Conference on Robotics and Automation (ICRA)}, pages 2806--2813. IEEE, 2024.

\bibitem[Yao et~al.(2022{\natexlab{a}})Yao, Chen, Yang, and Narasimhan]{yao2022webshop}
Shunyu Yao, Howard Chen, John Yang, and Karthik Narasimhan.
\newblock Webshop: Towards scalable real-world web interaction with grounded language agents.
\newblock \emph{Advances in Neural Information Processing Systems}, 35:\penalty0 20744--20757, 2022{\natexlab{a}}.

\bibitem[Yao et~al.(2022{\natexlab{b}})Yao, Zhao, Yu, Du, Shafran, Narasimhan, and Cao]{yao2022react}
Shunyu Yao, Jeffrey Zhao, Dian Yu, Nan Du, Izhak Shafran, Karthik~R Narasimhan, and Yuan Cao.
\newblock React: Synergizing reasoning and acting in language models.
\newblock In \emph{The eleventh international conference on learning representations}, 2022{\natexlab{b}}.

\bibitem[Yao et~al.(2023)Yao, Yu, Zhao, Shafran, Griffiths, Cao, and Narasimhan]{yao2023tree}
Shunyu Yao, Dian Yu, Jeffrey Zhao, Izhak Shafran, Tom Griffiths, Yuan Cao, and Karthik Narasimhan.
\newblock Tree of thoughts: Deliberate problem solving with large language models.
\newblock \emph{Advances in neural information processing systems}, 36:\penalty0 11809--11822, 2023.

\bibitem[Yoneda et~al.(2024)Yoneda, Fang, Li, Zhang, Jiang, Lin, Picker, Yunis, Mei, and Walter]{yoneda2024statler}
Takuma Yoneda, Jiading Fang, Peng Li, Huanyu Zhang, Tianchong Jiang, Shengjie Lin, Ben Picker, David Yunis, Hongyuan Mei, and Matthew~R Walter.
\newblock Statler: State-maintaining language models for embodied reasoning.
\newblock In \emph{2024 IEEE International Conference on Robotics and Automation (ICRA)}, pages 15083--15091. IEEE, 2024.

\bibitem[Yu et~al.(2025)Yu, Wan, Wang, Gao, Gan, Zhang, and Zhan]{yu2025reward}
Rui Yu, Shenghua Wan, Yucen Wang, Chen-Xiao Gao, Le~Gan, Zongzhang Zhang, and De-Chuan Zhan.
\newblock Reward models in deep reinforcement learning: A survey.
\newblock \emph{arXiv preprint arXiv:2506.15421}, 2025.

\bibitem[Zhang et~al.(2025{\natexlab{a}})Zhang, Yang, Liu, Li, Han, Chen, Huang, Fu, and Yu]{zhang2025appagent}
Chi Zhang, Zhao Yang, Jiaxuan Liu, Yanda Li, Yucheng Han, Xin Chen, Zebiao Huang, Bin Fu, and Gang Yu.
\newblock Appagent: Multimodal agents as smartphone users.
\newblock In \emph{Proceedings of the 2025 CHI Conference on Human Factors in Computing Systems}, pages 1--20, 2025{\natexlab{a}}.

\bibitem[Zhang et~al.(2025{\natexlab{b}})Zhang, Chen, Liu, Xue, Liao, Liu, Wang, Ning, Chen, Fu, et~al.]{zhang2025agent}
Kai Zhang, Xiangchao Chen, Bo~Liu, Tianci Xue, Zeyi Liao, Zhihan Liu, Xiyao Wang, Yuting Ning, Zhaorun Chen, Xiaohan Fu, et~al.
\newblock Agent learning via early experience.
\newblock \emph{arXiv preprint arXiv:2510.08558}, 2025{\natexlab{b}}.

\bibitem[Zheng et~al.(2023)Zheng, Chiang, Sheng, Zhuang, Wu, Zhuang, Lin, Li, Li, Xing, et~al.]{zheng2023judging}
Lianmin Zheng, Wei-Lin Chiang, Ying Sheng, Siyuan Zhuang, Zhanghao Wu, Yonghao Zhuang, Zi~Lin, Zhuohan Li, Dacheng Li, Eric Xing, et~al.
\newblock Judging llm-as-a-judge with mt-bench and chatbot arena.
\newblock \emph{Advances in neural information processing systems}, 36:\penalty0 46595--46623, 2023.

\bibitem[Zhou et~al.(2024)Zhou, Pan, LeCun, and Pinto]{zhou2411dino}
Gaoyue Zhou, Hengkai Pan, Yann LeCun, and Lerrel Pinto.
\newblock Dino-wm: World models on pre-trained visual features enable zero-shot planning.
\newblock \emph{arXiv preprint arXiv:2411.04983}, 2024.

\bibitem[Zhou et~al.(2023)Zhou, Xu, Zhu, Zhou, Lo, Sridhar, Cheng, Ou, Bisk, Fried, et~al.]{zhou2023webarena}
Shuyan Zhou, Frank~F Xu, Hao Zhu, Xuhui Zhou, Robert Lo, Abishek Sridhar, Xianyi Cheng, Tianyue Ou, Yonatan Bisk, Daniel Fried, et~al.
\newblock Webarena: A realistic web environment for building autonomous agents.
\newblock \emph{arXiv preprint arXiv:2307.13854}, 2023.

\bibitem[Zhu et~al.(2023)Zhu, Chen, Tian, Tao, Su, Yang, Huang, Li, Lu, Wang, et~al.]{zhu2023ghost}
Xizhou Zhu, Yuntao Chen, Hao Tian, Chenxin Tao, Weijie Su, Chenyu Yang, Gao Huang, Bin Li, Lewei Lu, Xiaogang Wang, et~al.
\newblock Ghost in the minecraft: Generally capable agents for open-world environments via large language models with text-based knowledge and memory.
\newblock \emph{arXiv preprint arXiv:2305.17144}, 2023.

\bibitem[Zhuge et~al.(2023)Zhuge, Liu, Faccio, Ashley, Csord{\'a}s, Gopalakrishnan, Hamdi, Hammoud, Herrmann, Irie, et~al.]{zhuge2023mindstorms}
Mingchen Zhuge, Haozhe Liu, Francesco Faccio, Dylan~R Ashley, R{\'o}bert Csord{\'a}s, Anand Gopalakrishnan, Abdullah Hamdi, Hasan Abed Al~Kader Hammoud, Vincent Herrmann, Kazuki Irie, et~al.
\newblock Mindstorms in natural language-based societies of mind.
\newblock \emph{arXiv preprint arXiv:2305.17066}, 2023.

\bibitem[Ziakas and Russo(2025)]{ziakas2025test}
Christos Ziakas and Alessandra Russo.
\newblock Test-time adaptation for generalizable task progress estimation.
\newblock \emph{arXiv preprint arXiv:2506.10085}, 2025.

\end{thebibliography}
\bibliographystyle{plainnat}

\newpage
\appendix
\onecolumn

\sectionheaderline{
  \seclink{sec:intro}{1}{Intro} | 
  \seclink{sec:benchmark}{2}{Benchmark} | 
  \seclink{sec:methods}{3}{Method} |
  \seclink{sec:experiments}{4}{Experiments} |
  \seclink{sec:related-works}{5}{Related Works} |
  \seclink{sec:conclusion}{6}{Conclusion} |
  \textbf{\hyperref[sec:appendix]{Sec 7: Appx}} 
}

\phantomsection
\section*{\centerline{Appendix}}
\label{sec:appendix}

\begin{itemize}[leftmargin=*, label={}, itemsep=2em plus 0.5em]
    
    \item \hyperref[app:benchmark]{\textbf{\Large A. Benchmark Details and Data Construction}}
    \begin{itemize}[leftmargin=1.5em, label=--, itemsep=0.8em plus 0.2em, topsep=0.5em]
        
        \item {\large \hyperref[app:four_stage_pipeline]{A.1 Four-Stage Data Generation Pipeline}} \\
        \textcolor{black!70}{\textit{\normalsize Details the rigorous four-stage pipeline for generating positive/negative trajectories and assigning hybrid ground-truth rewards.}}
        
        \item {\large \hyperref[app:task_environments]{A.2 Evaluation Task Environments}} \\
        \textcolor{black!70}{\textit{\normalsize Introduces the five evaluation environments, detailing their core mechanics and respective reasoning challenges.}}
        
        \item {\large \hyperref[app:sourcing_preprocessing]{A.3 Domain-Specific Sourcing and Preprocessing}} \\
        \textcolor{black!70}{\textit{\normalsize Explains the tailored sampling, filtering, and procedural generation strategies across the five evaluated task domains.}}
        
        \item {\large \hyperref[app:dataset_stats]{A.4 Dataset Statistics}} \\
        \textcolor{black!70}{\textit{\normalsize Provides a comprehensive statistical summary of trajectory pairs and native reward types included in the benchmark.}}
        
        \item {\large \hyperref[app:data_visualization]{A.5 Data Sample Visualization}} \\
        \textcolor{black!70}{\textit{\normalsize Presents detailed timestep-wise trajectory tables visualizing representative positive and negative interaction samples to clearly illustrate the mechanics of our hybrid reward labeling protocol.}}
    
    \end{itemize}

    \item \hyperref[app:hierarchical_matching_details]{\textbf{\Large B. Hierarchical Routing}} \\
    \vspace{0.2em} 
    \textcolor{black!70}{\textit{\normalsize Details the global semantic alignment process, including dynamic key alignment, attribute satisfaction scoring, and joint object selection for dense reward derivation.}}

    \item \hyperref[app:implementation_details]{\textbf{\Large C. Experimental Implementation Details}}
    \begin{itemize}[leftmargin=1.5em, label=--, itemsep=0.8em plus 0.2em, topsep=0.5em]
        \item {\large \hyperref[app:monotonic_baseline]{C.1 Monotonic Baseline (Rule-based Heuristic)}} \\
        \textcolor{black!70}{\textit{\normalsize Describes the non-parametric, rule-based reference baseline that models task progress via naive temporal linear interpolation.}}
        
        \item {\large \hyperref[app:supervised_rm]{C.2 Supervised Reward Model}} \\
        \textcolor{black!70}{\textit{\normalsize Details the preference dataset curation, model architecture, LoRA configuration, and training hyperparameters.}}
        
        \item {\large \hyperref[app:baseline_prompts]{C.3 LLM-as-a-Judge}} \\
        \textcolor{black!70}{\textit{\normalsize Outlines the direct full-history trajectory evaluation protocol, detailed prompt-level reasoning ablations (CoT vs. Zero-Shot), and specific test-time compute scaling strategies designed for robust reward estimation.}}
        
        \item {\large \hyperref[app:statefactory_details]{C.4 StateFactory}} \\
        \textcolor{black!70}{\textit{\normalsize Provides hyperparameters, semantic embedding model selections, and comprehensive prompt templates for the inference pipeline.}}

        \item {\large \hyperref[app:ablation]{C.5 Ablation}} \\
        \textcolor{black!70}{\textit{\normalsize Presents detailed ablation studies on state representation granularity, goal interpretation modalities, semantic embeddings, and LLM backbones to validate each core component's contribution.}}
        
        \item {\large \hyperref[app:planning]{C.6 Utility for Agent Planning}} \\
        \textcolor{black!70}{\textit{\normalsize Details the implementation of the System-1 ReAct policy, its dense-reward enhancement via the StateFactory engine, and System-2 MCTS planning with World Models, including a detailed trajectory trace.}}

    \end{itemize}
    
    \item \hyperref[app:reward_prediction_examples]{\textbf{\Large D. RewardPrediction Examples}} \\
    \vspace{0.2em}
    \textcolor{black!70}{\textit{\normalsize Provides extended visual case studies of positive and negative evaluation trajectories across multiple simulated environments.}}

\end{itemize}

\vfill 
\clearpage

\section{Benchmark Details and Data Construction}
\label{app:benchmark}

\subsection{Four-Stage Data Generation Pipeline}
\label{app:four_stage_pipeline}
To ensure high-quality supervision for reward modeling and to mitigate the sparsity of environmental feedback, we constructed the \textsc{RewardPrediction} benchmark using a rigorous four-stage data generation pipeline derived strictly from native environment feedback.

\begin{itemize}[leftmargin=*]
    \item \textbf{Positive Trajectory Synthesis with Random Augmentation.} 
    Positive trajectories are primarily grounded in absolute expert demonstrations provided by native environments or human experts. To enhance the diversity of the dataset and introduce temporal robustness, we implement a \textit{Random Padding} mechanism: expert sequences are padded with $k \in [0, 3]$ random interaction steps, which are stochastically distributed at both the initiation and termination of the trajectory. This augmentation simulates realistic scenarios where an agent might perform sub-optimal exploration before or after achieving the core task.

    \item \textbf{Ground-Truth Reward Derivation via Hybrid Labeling.} 
    (i) \textbf{Expert Segment Densification:} The core expert segment is assigned dense rewards via linear interpolation, formulated as $r_t = t/T_{\text{expert}}$. This captures the monotonic progress and the increasing value of states as they approach the goal. 
    (ii) \textbf{Random Step Grounding:} Conversely, the appended random steps strictly inherit their original environmental reward values (typically zero). This ensures the model learns a sharp contrast between the dense progress of expert actions and the random exploration.

    \item \textbf{Negative Sample Generation \& Strict Filtering Protocol.}
    Negative trajectories are generated via a stochastic random policy to represent non-attainment of goals. While random exploration predominantly yields no progress, we enforce a \textit{Strict Filtering Protocol} to eliminate label noise: any random trajectory that inadvertently overlaps with, or matches any sub-segment of, an expert demonstration is immediately discarded. By mitigating the risk of "accidental successes" being labeled as failures, we justify a strict ground truth of $r_t \neq 1$ for all steps in the negative set, establishing a robust, flat non-successful baseline.

    \item \textbf{Native Observation, Action Space and Reward.} 
    To ensure the \textbf{maximal fidelity and empirical rigor} of the \textsc{RewardPrediction} benchmark, we strictly adhere to a principle of minimal intervention across all data modalities. All observations are extracted directly from the native environment without any form of manual modification. Similarly, we maintain the absolute completeness of the action space by requiring the model to operate within the full, unaltered set of admissible candidate actions provided by the environment at each discrete time step, rather than relying on simplified or heuristic-based action subsets. Furthermore, all foundational reward signals utilized in our labeling pipeline are sourced entirely from the environment's intrinsic feedback logic. This preservation of unedited signals ensures that our benchmark reflects the authentic complexity of the original task domain.
\end{itemize}

\subsection{Evaluation Task Environments}
\label{app:task_environments}
To evaluate our method across a diverse set of challenges, ranging from household navigation to scientific reasoning, we incorporated five distinct interactive environments:

\begin{itemize}
    \item \textbf{\texttt{AlfWorld}:} \texttt{AlfWorld} is an embodied text-based environment covering tasks like \texttt{cleaning, heating, and cooling objects}, challenging agents to ground high-level language instructions into sequential navigation and manipulation actions within simulated domestic scenes.
    
    \item \textbf{\texttt{ScienceWorld}:} \texttt{ScienceWorld} simulates scientific experiments across interconnected rooms, demanding a deep understanding of physics, scientific causality, and multi-step logical reasoning to interact with diverse objects.
    
    \item \textbf{\texttt{WebShop}:} \texttt{WebShop} is a massive simulated e-commerce environment built on real-world Amazon data, testing the agent's ability to process noisy web text, match complex multi-attribute user intents, and perform long-horizon navigation to successfully complete purchases.
    
    \item \textbf{\texttt{TextWorld}:} \texttt{TextWorld} is a procedural text-adventure framework with partially observable environments, requiring agents to utilize memory and map abstract textual clues to sequential puzzle-solving strategies.
    
    \item \textbf{\texttt{BlocksWorld}:} \texttt{BlocksWorld} is a fundamental planning environment where the agent must logically unstack and restack colored blocks on a table to reach a specific target configuration, testing basic spatial and sequential reasoning under strict physical constraints.
    
\end{itemize}

\subsection{Domain-Specific Sourcing and Preprocessing}
\label{app:sourcing_preprocessing}
We tailored the sampling and filtering strategy for each domain to ensure dataset quality and balance:

\begin{itemize}
    \item \textbf{\texttt{AlfWorld} (Uniform Sampling):} We randomly sampled \textbf{336} tasks from the training set, representing approximately 10\% of the total data. To prevent task bias, we ensured a uniform distribution across the six distinct task types, resulting in exactly 56 positive-negative trajectory pairs per task type.
    
    \item \textbf{\texttt{ScienceWorld} (Validity Filtering):} We selected \textbf{191} trajectories from the test set. During selection, we explicitly filtered out specific tasks (e.g., \textit{task-4-grow-fruit} and \textit{task-4-grow-plant}) where the native oracle could not reliably reach a task progress of 1.0 due to environmental limitations. Additionally, we clipped terminal failure scores (e.g., -100) to 0 to maintain reward boundedness.
    
    \item \textbf{\texttt{WebShop} (Human Demonstrations):} We leveraged human demonstrations as the expert policy. From a total pool of 711 successful human trajectories, we randomly subsampled \textbf{300} high-quality instances to serve as positive samples, ensuring a diverse coverage of e-commerce navigation behaviors.
    
    \item \textbf{\texttt{TextWorld} (Procedural Generation):} We utilized the native engine to procedurally generate \textbf{150} unique text-adventure instances, focusing on cooking and treasure-hunting quests that require multi-step reasoning.
    
    \item \textbf{\texttt{BlocksWorld} (Subsampling \& PDDL):} Sourced from the \texttt{BlocksWorld} suite, we randomly subsampled \textbf{250} instances from the official 500-instance dataset. We utilized the authorized PDDL-to-text translator to convert symbolic states (e.g., \texttt{on(a, b)}) into natural language descriptions, ensuring consistency with other environments.
\end{itemize}

\subsection{Dataset Statistics}
\label{app:dataset_stats}

\begin{table}[H]
\centering
\small
\caption{\textbf{Statistics of the \textsc{RewardPrediction} Benchmark.} We report the total number of positive and negative trajectory pairs, the type of native reward signal provided by the environment, and the nature of tasks across different domains.}
\label{tab:benchmark_stats}

\begin{tabularx}{\textwidth}{@{} l >{\hsize=0.8\hsize\centering\arraybackslash}X >{\hsize=1.0\hsize\raggedright\arraybackslash}X >{\hsize=1.2\hsize\raggedright\arraybackslash}X @{}} 
\toprule
\textbf{Domain} & \textbf{Pairs (Pos / Neg)} & \textbf{Native Reward Type} & \textbf{Task Description} \\ \midrule
\textbf{\texttt{AlfWorld}} & 336 / 336 & Sparse (Binary) & Embodied household instruction following \\
\textbf{\texttt{ScienceWorld}} & 191 / 191 & Dense (0-100 Score) & Interactive scientific experiments \\
\textbf{\texttt{WebShop}} & 300 / 300 & Dense (Attribute Match) & E-commerce navigation (Subsampled human demos) \\
\textbf{\texttt{BlocksWorld}} & 250 / 250 & Sparse (Binary) & Classical planning with spatial reasoning \\
\textbf{\texttt{TextWorld}} & 150 / 150 & Sparse (Binary) & Text-adventure games via native engine \\ \bottomrule
\end{tabularx}
\end{table}

\subsection{Data Sample Visualization}
\label{app:data_visualization}

To provide a granular view of our hybrid data construction and labeling strategy, we transform raw environment interactions into structured, timestep-wise trajectory tables. Table~\ref{tab:sample_pos} and Table~\ref{tab:sample_neg} visualize a representative positive-negative trajectory pair from the \texttt{AlfWorld} domain, illustrating how our pipeline converts sparse environmental feedback into a dense, semantically-rich supervision signal.

\noindent\textbf{Case Analysis: Positive Trajectory.} 
Table~\ref{tab:sample_pos} visualizes a positive trajectory consisting of agent actions ($a_t$), environment observations ($o_t$), and derived ground-truth rewards ($R^*_t$). This sample exemplifies our hybrid supervision approach:
\begin{itemize}[leftmargin=*]
    \item \textbf{Expert Segments ($t=0 \text{ to } 5$):} The core sequence follows an optimal expert policy. As per our methodology, these steps are assigned \textbf{monotonically increasing dense rewards} (from $0.17$ to $1.00$) via linear interpolation, providing a granular gradient that captures task progress.
    \item \textbf{Random Augmentation ($t=6$):} The final step is an appended random action (\texttt{random\_post}). Crucially, this step strictly inherits the native environment success signal ($R=1.0$), ensuring the model learns to maintain the success state even amidst non-expert exploration.
\end{itemize}

\noindent\textbf{Case Analysis: Negative Trajectory.} 
In contrast, Table~\ref{tab:sample_neg} displays a negative trajectory generated via a stochastic random policy. Despite the agent's active interaction with the environment (e.g., opening drawers), the actions fail to advance the task state toward the goal. Following our \textit{Strict Filtering Protocol}, these steps are uniformly assigned a \textbf{flat ground truth of $0$}, establishing a robust non-progressing baseline that represents a clean failure case.

\begin{table}[H]
\centering
\caption{\textbf{Visualized Positive Sample (\texttt{AlfWorld}).} Task: \textit{"put a hot potato in diningtable."} The expert segment ($t=0\text{-}5$) receives interpolated dense rewards, while the final random augmentation ($t=6$) inherits the native success signal.}
\label{tab:sample_pos}
\resizebox{\textwidth}{!}{
    \begin{tabularx}{1.1\textwidth}{@{}c l X c@{}}
    \toprule
    \textbf{Step} & \textbf{Action ($a_t$)} & \textbf{Observation ($o_t$)} & \textbf{GT ($R_t$)} \\ \midrule
    0 & go to countertop 3 & You arrive at countertop 3. On the countertop 3, you see a bread 1... a potato 1... & 0.17 \\
    1 & take potato 1... & You pick up the potato 1 from the countertop 3. & 0.33 \\
    2 & go to microwave 1 & You arrive at microwave 1. The microwave 1 is closed. & 0.50 \\
    3 & heat potato 1... & You heat the potato 1 using the microwave 1. & 0.67 \\
    4 & go to diningtable 1 & You arrive at diningtable 1. On the diningtable 1, you see a bowl 3... & 0.83 \\
    5 & move potato 1... & You move the potato 1 to the diningtable 1. & 1.00 \\ \midrule
    \textbf{6} & \textbf{go to toaster 1} & \textbf{You arrive at toaster 1. On the toaster 1, you see nothing.} & \textbf{1.00} \\ \bottomrule
    \end{tabularx}
}
\end{table}

\begin{table}[H]
\centering
\caption{\textbf{Visualized Negative Sample (\texttt{AlfWorld}).} Task: \textit{"put a hot potato in diningtable."} A completely random trajectory where actions fail to advance the task state, resulting in a consistent zero reward.}
\label{tab:sample_neg}
\resizebox{\textwidth}{!}{
    \begin{tabularx}{1.1\textwidth}{@{}c l X c@{}}
    \toprule
    \textbf{Step} & \textbf{Action ($a_t$)} & \textbf{Observation ($o_t$)} & \textbf{GT ($R_t$)} \\ \midrule
    0 & go to stoveburner 1 & You arrive at stoveburner 1. On the stoveburner 1, you see a pot 1. & 0.00 \\
    1 & go to cabinet 7 & You arrive at cabinet 7. The cabinet 7 is closed. & 0.00 \\
    2 & go to drawer 6 & You arrive at drawer 6. The drawer 6 is closed. & 0.00 \\
    3 & open drawer 6 & You open the drawer 6. The drawer 6 is open. In it, you see nothing. & 0.00 \\
    4 & go to cabinet 8 & You arrive at cabinet 8. The cabinet 8 is closed. & 0.00 \\
    5 & open cabinet 8 & You open the cabinet 8. The cabinet 8 is open. In it, you see nothing. & 0.00 \\
    6 & go to toaster 1 & You arrive at toaster 1. On the toaster 1, you see nothing. & 0.00 \\ \bottomrule
    \end{tabularx}
}
\end{table}

\section{Hierarchical Routing}
\label{app:hierarchical_matching_details}

As conceptualized in Section~\ref{subsec:reward_estimation}, the dense reward $\hat{r}_t$ is derived through a global semantic alignment process. We search for the physical instantiation that maximizes a composite {Identity-Attribute score. This process is detailed below.

\textbf{Semantic Key Alignment (The Inner Loop).}
To address potential linguistic discrepancies between the goal description and environment observations, we do not assume fixed key correspondences. For a required attribute pair $(\boldsymbol{\alpha}_k, \mathbf{v}_k)$ in goal entity $e_k$, and a candidate state entity $e_i$, we dynamically identify the most relevant key $\boldsymbol{\alpha}^*$ in $e_i$ via hard-max alignment:
\begin{equation}
    \boldsymbol{\alpha}^* = \underset{\boldsymbol{\alpha}' \in \text{keys}(e_i)}{\operatorname{arg\,max}} \ \text{sim}(\boldsymbol{\alpha}_k, \boldsymbol{\alpha}').
\end{equation}
The value $\mathbf{v}_i^*$ referenced in Eq.~\ref{eq:attr_score} is strictly defined as the value associated with this optimal key $\boldsymbol{\alpha}^*$. This ensures we compare semantically equivalent properties regardless of naming conventions.

\textbf{Attribute Satisfaction Scoring.}
With the keys aligned, we compute the attribute satisfaction $\psi_{\text{attr}}(e_k, e_i)$ by averaging the semantic similarity between the goal values and the aligned state values:
\begin{equation}
    \psi_{\text{attr}}(e_k, e_i) = \frac{1}{|A_k|} \sum_{(\boldsymbol{\alpha}_k, \mathbf{v}_k) \in A_k} \text{sim}(\mathbf{v}_k, \mathbf{v}_i^*).
\end{equation}

\textbf{Joint Object Selection.}
To determine the local fulfillment $\hat{r}_{k,t}$, we evaluate every candidate entity $e_i$ in the current state $\hat{s}_t$. We define a composite alignment score $\Phi(e_k, e_i)$ that couples identity matching with state:
\begin{equation}
    \Phi(e_k, e_i) = \underbrace{\text{sim}(\mathbf{d}_k, \mathbf{d}_i)}_{\text{Identity}} \cdot \underbrace{\psi_{\text{attr}}(e_k, e_i)}_{\text{Attribute}}.
\end{equation}
This multiplicative interaction acts as a soft logic gate:
\begin{itemize}
    \item If the object is wrong (Low Identity), the total score is suppressed even if attributes match by chance.
    \item If the object is correct but the state is wrong (Low Attribute), the total score remains low.
\end{itemize}
The system selects the single candidate that maximizes this joint probability: $\hat{r}_{k,t} = \max_{e_i} \Phi(e_k, e_i)$.

\textbf{Global Aggregation.}
Finally, the total reward $\hat{r}_t$ is the mean of these maximized local scores across all $K$ goal entities, as defined in Eq.~\ref{eq:global_reward}.

\section{Experimental Implementation Details}
\label{app:implementation_details}

\subsection{Monotonic Baseline (Rule-based Heuristic)} 
\label{app:monotonic_baseline}
The \textit{Monotonic Baseline} serves as a non-parametric, rule-based reference to evaluate the necessity of semantic reward modeling. In this configuration, we bypass environmental feedback and instead apply a naive temporal heuristic to both positive and negative trajectories. Specifically, for any trajectory of length $T$, the reward at each time step $t$ is assigned via linear interpolation: $r_t = t/T$. By enforcing this strictly increasing progress signal regardless of the actual state-action semantics, this baseline tests whether a simple "time-equals-progress" assumption can suffice for the task. It provides a lower bound for our evaluation, highlighting the limitations of ignoring the distinction between expert goal-oriented actions and aimless exploration.

\subsection{Supervised Reward Model} 
\label{app:supervised_rm}
In this section, we detail the dataset curation, model architecture, and training protocols employed for the reward model (RM) baselines.

\subsubsection{Dataset Construction and Input Format}
We curated preference datasets across five distinct interactive environments: \texttt{AlfWorld}, \texttt{TextWorld}, \texttt{ScienceWorld}, \texttt{Blocks World}, and \texttt{WebShop}. We collected 35,000 preference pairs for each environment, yielding a cumulative dataset of 175,000 pairs.

Each sample comprises a raw text trajectory containing a high-level goal, a sequence of actions, and corresponding state observations. Deviating from standard instruction-tuning formats, we eschew separate system prompts; instead, the input is fed as a flattened interaction history. The input format is structured as follows:

\begin{quote}
\small
\texttt{<GOAL> [Task Description] </GOAL> --- <ACTION> [Action$_1$] </ACTION> <STATE> [Observation$_1$] </STATE> ...}
\end{quote}

We trained six distinct reward models: five task-specific models (35k pairs each) and one comprehensive model (``All-Combined'') trained on the union of all datasets (175k pairs).

\subsubsection{Model Architecture and LoRA Configuration}
We employ \textbf{Qwen2.5-1.5B} \citep{hui2024qwen2} as the backbone architecture. To manage computational costs efficiently and mitigate potential catastrophic forgetting, we utilize \textbf{Low-Rank Adaptation (LoRA)} \citep{hu2022lora} in lieu of full-parameter fine-tuning.

The reward model is implemented by appending a scalar classification head to the base model. LoRA adapters are applied to all linear layers with a rank of $r=16$ and an alpha scaling factor of $\alpha=32$. The training objective minimizes the negative log-likelihood of the preferred trajectory rankings:
\begin{equation}
\mathcal{L}(\theta) = -\mathbb{E}_{(y_w, y_l) \sim \mathcal{D}} \left[ \log \sigma \left( r_\theta(y_w) - r_\theta(y_l) \right) \right]
\end{equation}
where $y_w$ and $y_l$ denote the chosen and rejected trajectories, respectively.

\subsubsection{Training Setup and Hyperparameters}
All models were implemented using the Hugging Face \texttt{trl} library. To accommodate the extensive context requirements of agent interaction histories, we set the maximum sequence length to \textbf{8192} tokens and utilized mixed-precision training (BF16) with gradient checkpointing to optimize memory efficiency.

As detailed in Table \ref{tab:specific_details}, the hardware infrastructure varied due to resource availability, ranging from dual NVIDIA RTX A6000 configurations to clusters of NVIDIA A100 GPUs. Consequently, while the effective batch size and training duration varied by task, the core optimization hyperparameters (Table \ref{tab:common_hyperparams}) were kept constant to ensure fair comparison. Models were generally trained for 3 epochs, with the exception of \texttt{WebShop} and the \texttt{Combined} model, which were trained for 2 epochs.

Table \ref{tab:common_hyperparams} outlines the shared configuration, while Table \ref{tab:specific_details} provides the specific hardware setup, batch sizes, and training duration for each task.

\begin{table}[H] 
\centering
\caption{\textbf{Common Hyperparameters.} These settings were fixed across all experiments.}
\label{tab:common_hyperparams}
\begin{tabular}{ll}
\toprule
\textbf{Parameter} & \textbf{Value} \\
\midrule
Base Model & \texttt{Qwen2.5-1.5B} \\
Method & LoRA (\texttt{SEQ\_CLS}) \\
Precision & \texttt{BF16} \\
Max Length & 8192 \\
\midrule
\multicolumn{2}{c}{\textit{LoRA Configuration}} \\
Rank ($r$) / Alpha ($\alpha$) & 16 / 32 \\
Target Modules & All Linear Layers \\
\midrule
\multicolumn{2}{c}{\textit{Optimization}} \\
Optimizer & \texttt{AdamW} \\
Learning Rate & $1 \times 10^{-5}$ \\
LR Scheduler & Cosine \\
Batch Size & 16 / 8 / 2 \\
\bottomrule
\end{tabular}
\end{table}

\begin{table}[H]
\centering
\caption{\textbf{Task-Specific Training Details.} Comparison of dataset scale, compute resources, and training duration across tasks. Global batch size (BS) scales with GPU count.}
\label{tab:specific_details}
\definecolor{tableblue}{RGB}{235, 242, 255}

\begin{tabularx}{\linewidth}{@{} l >{\centering\arraybackslash}X c >{\centering\arraybackslash}X c c @{}}
\toprule
\multirow{2}{*}{\textbf{Task}} & \multicolumn{2}{c}{\textbf{Data Config}} & \multicolumn{2}{c}{\textbf{Compute Setup}} & \textbf{Cost} \\
\cmidrule(lr){2-3} \cmidrule(lr){4-5} \cmidrule(l){6-6}
 & \textbf{Size} & \textbf{Ep.} & \textbf{GPUs (Mem)} & \textbf{BS} & \textbf{Hours} \\
\midrule
\texttt{AlfWorld}     & 35k & 3 & 2$\times$A6000 (80G) & 16 & $\sim$4h \\
\texttt{TextWorld}    & 35k & 3 & 2$\times$A6000 (48G) & 4  & $\sim$4.5h \\
\texttt{ScienceWorld} & 35k & 3 & 2$\times$A100 (80G)  & 16 & $\sim$6h \\
\texttt{BlocksWorld}  & 35k & 3 & 2$\times$A100 (80G)  & 32 & $\sim$7h \\
\texttt{WebShop}      & 35k & 2 & 2$\times$A100 (80G)  & 16 & $\sim$10h \\
\midrule
\rowcolor{tableblue!50} 
\textbf{Combined} & \textbf{175k} & \textbf{2} & \textbf{4$\times$A100 (80G)} & \textbf{32} & \textbf{$\sim$20h} \\
\bottomrule
\end{tabularx}
\end{table}

\subsection{LLM-as-a-Judge} 
\label{app:baseline_prompts}

To ensure the \textbf{maximal fidelity and empirical rigor} of the evaluation, we implement a direct trajectory evaluation protocol where the \textit{Reward Evaluator} is provided with the full, uncompressed interaction history. All evaluating models are configured with a \textbf{context window of 8,192 tokens} to accommodate long-horizon trajectories without information loss. Furthermore, we set the \textbf{sampling temperature to $0.01$} across all variants to enforce near-deterministic outputs, thereby ensuring the high stability and reproducibility of our reward signals. To systematically investigate the impact of reasoning on reward judgment, we employ two distinct control mechanisms:

\noindent\textbf{1. Prompt-Level Reasoning Ablation (e.g., Qwen-2.5-14B).} 

For general-purpose LLMs, we modulate reasoning capabilities through explicit system-level instructions, allowing us to isolate the effects of internal Chain-of-Thought (CoT) processes. Table~\ref{tab:baseline_full_prompts} provides the detailed prompt templates used for these variants:

\begin{itemize}[leftmargin=*]

    \item \textbf{Intuitive Mode (No-Thinking):} The system prompt imposes strict negative constraints, forcing the model to provide an immediate judgment based on "first-order intuition" without intermediate analysis (e.g., \textit{"Directly output the score. Do NOT analyze, reason, or explain"}).

    \item \textbf{Analytical Mode (CoT):} All negative constraints are removed, and the model is explicitly instructed to perform a comprehensive review of the trajectory history and justify its scoring logic through step-by-step reasoning before arriving at a final value.

\end{itemize}

\noindent\textbf{2. Test-Time Compute Scaling (Reasoning Effort Control).} 
For models with native reasoning-specialized architectures (e.g., GPT-OSS-20B), we modulate the depth of internal verification and search processes by adjusting the \texttt{reasoning\_effort} parameter. This mechanism allows for the scaling of the \textbf{test-time compute budget} independently of the input prompt structure. In our experiments, we compare a \textit{Baseline} (Low reasoning effort) against an \textit{Enhanced} (Medium reasoning effort) configuration. Notably, as these models generate reasoning within a native ``reasoning content'' accessible via metadata, we can evaluate the logical derivation without requiring explicit ``analysis'' fields in the final text response. This allows us to maintain a clean, standardized output format—consisting solely of the reward score—across all models while still capturing the full depth of latent reasoning. By systematically varying the ``reasoning effort'' parameter, we can observe how increased latent search steps and self-correction cycles contribute to a more precise grounding of the reward signal, effectively decoupling the model's evaluative accuracy from its prompt-based instructions.

\begin{table}[H]
\centering
\caption{\textbf{Detailed Baseline Prompt Templates.} We strictly follow the provided prompt structures for both variants under the \textbf{Full-History Protocol}. The CoT variant (Part I) uses a standard helper system prompt, while the Zero-Shot variant (Part II) explicitly injects a \textbf{[Without Thinking]} constraint in the system prompt to suppress reasoning.}
\begin{tcolorbox}[
    colback=white,
    colframe=blue!50!black,
    title=\textbf{Detailed Prompt Templates for Baseline Implementation},
    fonttitle=\bfseries\small,
    boxrule=0.8pt, arc=3pt,
    left=8pt, right=8pt, top=8pt, bottom=8pt
]
\footnotesize

\promptsection{blue}{PART I: Chain-of-Thought (CoT) Variant \hfill (Reasoning Mode)}

\begin{itemize}[leftmargin=1.2em, nosep, label={}]
    \item \inlbl{System Prompt:} You are a helpful assistant. Please output the final answer in JSON format.

    \item \inlbl{User Prompt:} You are an expert in task progress evaluation. Please judge the agent's progress (0-100) based on the full interaction history.
    
    \item \textbf{[Task Description]} \\
    \var{\{task\_description\}}

    \item \textbf{[Interaction History]} \\
    The following is the chronological log of the agent's past actions and observations: \\
    \var{\{history\_log\}}

    \item \textbf{[Evaluation Requirements]}
    \begin{enumerate}[nosep, leftmargin=1.5em]
        \item \textbf{Analysis:} Review the History to evaluate the progress strictly.
        \item \textbf{Scoring:} Provide an integer score (0-100).
    \end{enumerate}

    \item \inlbl{Output strictly in JSON:} \\
    \jsonbox{\{"reward": 50\}}
\end{itemize}

\promptsection{gray}{PART II: Zero-Shot (No-Thinking) Variant \hfill (Direct Output)}

\begin{itemize}[leftmargin=1.2em, nosep, label={}]
    \item \inlbl{System Prompt:} You are a helpful assistant. Please output the final answer in JSON format. \\
    \textbf{[Without Thinking]}

    \item \inlbl{User Prompt:} You are an expert in task progress evaluation. Judge the agent's progress (0-100) based on the full interaction history provided below.
    
    \item \textbf{[Task Description]} \\
    \var{\{task\_description\}}

    \item \textbf{[Full Interaction History (From Start to Present)]} \\
    The following is the chronological log of all agent actions and observations up to the current moment: \\
    \var{\{history\_log\}}

    \item \textbf{[Evaluation Requirements]}
    \begin{itemize}[nosep, leftmargin=1.5em, label=$\bullet$]
        \item Determine the progress score (0-100) based on the final state in the history.
        \item \textbf{Do NOT analyze, reason, or explain.}
        \item Rely on your intuition to give a score immediately.
    \end{itemize}

    \item \textbf{[Output Format]} \\
    Output strictly in JSON format: \\
    \jsonbox{\{"reward": 50\}}
\end{itemize}

\end{tcolorbox}
\label{tab:baseline_full_prompts}
\end{table}

\subsection{\textsc{StateFactory}}
\label{app:statefactory_details}

\noindent\textbf{Hyperparameters.} 
To ensure experimental reproducibility and minimize stochastic variance, we utilize \textbf{gpt-oss-20b} as the primary reasoning backbone for all evaluative tasks. The model is deployed via the vLLM inference engine, with the sampling temperature fixed at a near-deterministic $T=0.01$. We maintain a consistent context window of 8,192 tokens.

\noindent\textbf{Embedding Model.} 
Semantic similarity measures within \textsc{StateFactory} are computed using \textbf{all-MiniLM-L6-v2}. This specific encoder was selected based on our systematic ablation study (see Figure~\ref{fig:abolation}), where it demonstrated superior alignment fidelity for environment-specific state representations. Despite its compact 384-dimensional vector space, it outperformed larger alternatives in capturing the underlying semantic distance.

\subsubsection{\textsc{StateFactory} Inference Pipeline}
\label{app:statefactory_prompts}

Our proposed method transforms raw observations into structured states via a multi-stage pipeline. 
Table~\ref{tab:semantic_state_factorization} details the \textbf{Semantic State Factorization} process (Modules A \& B). 
Table~\ref{tab:dynamic_goal_interpretation} details the \textbf{Dynamic Goal Interpretation} (Module C).

\begin{table}[H]
\centering
\caption{\textbf{Semantic State Factorization Templates.} This pipeline constitutes the first half of the StateFactory method. Module A grounds raw observations into structured states, while Module B filters and refines these states to maintain a coherent, task-relevant history.}
\begin{tcolorbox}[
    fit,
    height=\dimexpr\textheight-3.5cm\relax,
    colback=white,
    colframe=teal!50!black,
    title=\textbf{StateFactory Pipeline Part I: Semantic State Factorization},
    fonttitle=\bfseries\small,
    boxrule=0.8pt, arc=3pt,
    left=6pt, right=6pt, top=6pt, bottom=6pt
]

\modulename{teal}{MODULE A: World State Extraction Template \hfill (Perception \& Grounding)}

\begin{itemize}[leftmargin=1.2em, nosep, label={}]
    \item \lbl{System Instruction:} \var{\{system\_instruction\}}
    \item \lbl{Role:} You are a state extraction assistant. Your goal is to extract the current state of the agent based on the provided information, following the rules below:
    \item \lbl{Rules:}
        \begin{enumerate}[nosep, leftmargin=1.5em]
            \item \textbf{Output Format:} Output a SINGLE JSON object. It must contain two keys: "\textsf{thinking}" and "\textsf{current\_state}".
            \begin{itemize}[nosep]
                \item "\textsf{thinking}": "For each object, cite the exact phrase from 'Observation' that justifies its state. Verify that the object actually exists in 'Observation' or 'Previous States'. EXPLICITLY REJECT any information that appears ONLY in 'Previous Goal State'."
                \item "\textsf{current\_state}": \var{\{output\_format\_des\}}
            \end{itemize}
            \item \textbf{Relevance:} Only extract objects that are present in the Observation, or Previous States, and that are directly relevant to the Task Description. Include only those objects that actively contribute to task progress or are essential for understanding the current task state. Do not extract from Task Description or Previous Goal State.
            \item \textbf{Previous Goal State:} This is provided ONLY for context verification. IT IS A FORBIDDEN SOURCE. You must NOT copy, infer, or hallucinate any object or state from 'Previous Goal State' into 'current\_state'. If an object is in the goal but hasn't been observed yet, ignore it.
            \item \textbf{Action Observation:} The immediate feedback observed by the agent after performing an action. This may be None if no observable outcome occurs.
            \item \textbf{Previous States:} Represents the accumulated task-relevant states from previous steps. Use it to maintain continuity in the agent's world model:
            \begin{itemize}[nosep, label=-]
                \item If an object from Previous States appears in the current Observation, update its state with the most recent information.
                \item If an object from Previous States does not appear in the current Observation, retain its previous state in Current State, assuming it remains unchanged.
                \item Only include objects in Current State if they are either (a) newly observed and task-relevant, or (b) carried over from Previous States.
                \item Do not include objects that are neither present in the current Observation nor in Previous States.
            \end{itemize}
        \end{enumerate}
    \item \lbl{Input:} Task Description: \var{\{task\_description\}}, Previous Goal State: \var{\{prev\_goal\_state\}}, Action: \var{\{last\_action\}}, Observation: \var{\{observation\}}, Previous States: \var{\{prev\_states\}}
    \item \lbl{Output Example:} \var{\{output\_format\}} \var{\{output\_format\_des\}}
\end{itemize}

\modulename{teal}{MODULE B: Task-Relevant World State Extraction Template \hfill (Filtering \& Refinement)}

\begin{itemize}[leftmargin=1.2em, nosep, label={}]
    \item \lbl{System Instruction:} \var{\{system\_instruction\}}
    \item \lbl{Role:} You are a task-relevant state extraction assistant. Given the historical task-related states (Previous States), the current step's new states (Current State), the actions taken (Action History), and the Task Description, extract and update the relevant world states according to the following rules:
    \item \lbl{Rules:}
        \begin{enumerate}[nosep, leftmargin=1.5em]
            \item Extract task-relevant states from the Current State. Exclude non-factual statements. Do not extract from Task Description. A state is considered strongly relevant only if it directly or clearly indirectly affects whether the task can be completed. If uncertain about a state's relevance, exclude it.
            \item Combine the results from Rule 1 with the previous historical states (Previous States) to form an updated set of task-relevant historical states. Note that Previous States may be None, in which case the output should be based exclusively on the results from Rule 1.
            \item Refine the combined state set by preserving all states that are relevant to the current task, even if they are intermediate or partially superseded. Only remove or update a state if it represents an outdated or incorrect version of the same information about a specific object. In such cases, retain the most accurate and up-to-date state while discarding its obsolete counterparts. Avoid eliminating states merely because more specific or complete information has emerged, as long as the original states contribute to the task context or support traceability of reasoning.
            \item \textbf{Output Format:} Only output the JSON content—no additional text, explanation, or formatting. Follow the specified JSON format below exactly: \var{\{output\_format\}} \var{\{output\_format\_des\}}
        \end{enumerate}
    \item \lbl{Input:} Task Description: \var{\{task\_description\}}, Previous States: \var{\{prev\_states\}}, Current State: \var{\{current\_state\}}, Action History: \var{\{action\_history\}}
\end{itemize}

\end{tcolorbox}
\label{tab:semantic_state_factorization}
\end{table}

\begin{table}[H]
\centering
\caption{\textbf{Dynamic Goal Interpretation Template.} As the second stage of the StateFactory method, this component maintains a persistent yet evolving definition of success, strictly distinguishing between the agent's ``plans'' and the task's ``goals''.}
\begin{tcolorbox}[
    colback=white,
    colframe=teal!50!black,
    title=\textbf{StateFactory Pipeline Part II: Dynamic Goal Interpretation},
    fonttitle=\bfseries\small,
    boxrule=0.8pt, arc=3pt,
    left=6pt, right=6pt, top=6pt, bottom=6pt
]
\footnotesize

\modulename{teal}{MODULE C: Dynamic Goal Interpretation Template \hfill (Goal State Management)}

\begin{itemize}[leftmargin=1.2em, nosep, label={}]
    \item \lbl{System Instruction:} \var{\{system\_instruction\}}
    \item \lbl{Role:} You are an expert in task goal state extraction. Your mission is to generate an Evolving Goal State: a minimal, accurate JSON specification that defines only the FINAL, STATIC BLUEPRINT FOR TASK SUCCESS, derived exclusively from the Task Description.
    
    \item \lbl{Core Principle: Goal vs. Plan (CRITICAL)}
    \begin{itemize}[nosep, leftmargin=1em, label=$\bullet$]
        \item \textbf{Goal (Legal):} A concrete, factual state that must be objectively true at the moment of success AND is explicitly or implicitly required by the Task Description. This includes all explicit milestones in a multi-step task (e.g., "First do A, then do B").
        \item \textbf{Plan / Blocker (Illegal):} An intermediate step, prerequisite, enabling action, or blocker that YOU (the LLM) deduce is necessary, but is NOT one of the final goals listed in the Task Description.
        \item \textbf{Constraint:} Your Goal State MUST ONLY contain Goals. It MUST NEVER contain Plans or Blockers.
    \end{itemize}

    \item \lbl{Rules \& Instructions:}
        \begin{enumerate}[nosep, leftmargin=1.5em]
            \item \textbf{Output Format:} You must output only the JSON content. Do not include any additional text, explanations, comments, or markdown formatting. Adhere strictly to this JSON structure:
            \begin{itemize}[nosep]
                \item Output a SINGLE JSON object containing two keys:
                \item "\textsf{thinking}": A step-by-step analysis string. First, analyze the 'Task Description' to identify the ultimate success conditions. Second, verify that every goal attribute comes from the Task, NOT just because it exists in 'Current State'. REJECT any state that appears solely because it is currently true in 'Current State'.
                \item "\textsf{goal\_state}": \var{\{output\_format\_des\}}
                \item Example of desired output structure: \var{\{output\_format\}} \var{\{output\_format\_des\}}
            \end{itemize}

            \item \textbf{Core Logic: Blueprint Creation (Step 0)}
            \begin{itemize}[nosep]
                \item \textbf{a. Step 0: Complete Task Translation:}
                \begin{itemize}[nosep, label=-]
                    \item At Step 0, you MUST translate ALL final goal verbs and required milestones from the Task Description into their final physical states.
                    \item If a task has multiple required parts (e.g., "First do A, then do B"), ALL parts must be translated into final states and included in the blueprint from the beginning.
                    \item This blueprint is the COMPLETE and FINAL definition of success.
                \end{itemize}
                \item \textbf{b. State-Change Filter:} Use this only at Step 0 to exclude "trivial truths".
            \end{itemize}

            \item \textbf{Core Logic: Blueprint Evolution (Step $\ge$ 1)}
            \begin{itemize}[nosep]
                \item \textbf{a. GOAL PERSISTENCE:}
                \begin{itemize}[nosep, label=-]
                    \item The Goal State created at Step 0 is the static blueprint for success.
                    \item Once a goal is added at Step 0, it MUST REMAIN for the entire episode.
                    \item DO NOT remove goals just because they are achieved in the Current State.
                \end{itemize}
                \item \textbf{b. STRICT IMMUTABILITY:}
                \begin{itemize}[nosep, label=-]
                    \item You are STRICTLY FORBIDDEN from adding new goals after Step 0.
                    \item The blueprint from Step 0 is considered complete. Do not add goals for plans, prerequisites, or blockers that you deduce.
                \end{itemize}
                \item \textbf{c. LEGAL EVOLUTION:} The ONLY two legal modifications to the Goal State after Step 0 are:
                \begin{enumerate}[nosep, label=\arabic*.]
                    \item \textbf{Task Milestone Addition:}
                    \begin{itemize}[nosep, label=$\cdot$]
                        \item \textit{Condition:} The Task Description contains a multi-step requirement (e.g., "First do A, then do B"), and the Observation confirms that step A is now complete.
                        \item \textit{Action:} You are now allowed to add the goal for step B to the Goal State. This is legal because the goal for B originates from the task description.
                    \end{itemize}
                    \item \textbf{Refinement / Anchoring:}
                    \begin{itemize}[nosep, label=$\cdot$]
                        \item \textit{Condition:} A goal in the Goal State is generic, and the Observation first identifies its specific instance.
                        \item \textit{Action:} You MUST update the Goal State to "anchor" the generic goal to the specific, observed object. WARNING: Do not copy irrelevant attributes from Current State unless the Task specifically requires them.
                    \end{itemize}
                \end{enumerate}
            \end{itemize}
            
            \item \textbf{Sources of Truth (Inputs)}
            \begin{itemize}[nosep, label=$\cdot$]
                \item \textbf{Task Description:} Sole source of ultimate intent.
                \item \textbf{Observation:} Used only to ground generics or reveal implicit preconditions.
                \item \textbf{Previous Goal State:} The baseline to refine—never discard or contradict it without task-level justification.
                \item \textbf{Current State \& Action History:} REFERENCE ONLY. Provide contextual world facts and recent action history that may inform phrasing, object references, or state granularity, but they must not introduce new goal requirements.
            \end{itemize}
        \end{enumerate}

    \item \lbl{Input:} Task Description: \var{\{task\_description\}}, Current State: \var{\{current\_state\}}, Observation: \var{\{observation\}}, Action History: \var{\{action\_history\}}, Previous Goal State: \var{\{prev\_goal\_state\}}

    \item \lbl{Command:} Now generate the updated GoalState based on the above rules.
\end{itemize}

\end{tcolorbox}
\label{tab:dynamic_goal_interpretation}
\end{table}

\subsection{Ablation}
\label{app:ablation}
\subsubsection{Representation}
\label{app:granularity_details}

To rigorously evaluate the impact of state representation, we implemented a \textbf{Representation-Controlled Ablation}. We established a baseline using the raw environment output and compared it against three levels of LLM-processed abstractions. Crucially, strictly controlling the representation format allows us to decouple the benefits of \textit{denoising} from \textit{structuring}.

The four levels of granularity are defined as follows:

\begin{itemize}
    \item \textbf{Level 1: Unstructured Observations (Raw Input).}
    This baseline utilizes the raw, noisy output from the environment directly as the state representation. It serves as the control group for "zero processing."
    
    \item \textbf{Level 2: Textual States (Flat Denoising).}
    The model outputs a list of independent sentences. \textbf{Critically, this level maintains the same unstructured format as Level 1} (flat natural language) but applies the ``relevance filtering`` constraint. Comparing Level 1 and Level 2 isolates the performance gain attributed solely to \textbf{semantic denoising} without the introduction of hierarchical structure.
    
    \item \textbf{Level 3: Object-Centric States (Nested List).}
    The model groups sentences under specific object keys. This introduces \textbf{entity awareness} and tests the value of hierarchical grouping, but retains the ``entangled attributes`` limitation within the descriptions.
    
    \item \textbf{Level 4: Object-Attribute States (Full Factorization).}
    The model strictly separates descriptions into atomic Key-Value pairs. This is the standard \textsc{StateFactory} configuration, which tests the value of complete \textbf{attribute disentanglement}.
\end{itemize}

Table~\ref{tab:granularity_ablation_prompts} presents the specific representation schemas used to enforce these constraints for Levels 1, 2, 3, and 4.

\begin{table}[H]
\centering
\caption{\textbf{Prompt Variations for representation Structural Granularity.} We utilized the same backbone and input context across all levels, varying only the \textit{Output Schema} and \textit{Description Constraints} to isolate the effect of structural depth.}
\begin{tcolorbox}[
    colback=teal!2!white,
    colframe=teal!40!black,
    title=\textbf{Ablation Study: Output Representation Schemas},
    fonttitle=\bfseries\small,
    boxrule=0.8pt, arc=3pt,
    left=6pt, right=6pt, top=6pt, bottom=6pt
]
\footnotesize

\schema{Level 1\&2: Unstructured Observations (Raw Input) \& Textual States (Flat Denoising)} \hfill \textit{(Denoising only, no hierarchy)}

\par\noindent\hrulefill 

\begin{itemize}[leftmargin=1.2em, nosep, topsep=4pt, label={}]
    \item \textbf{Description Constraint:} Every element must be a grammatically complete English sentence that states only observable, factual, and specific information—no speculation or abstraction. Each sentence must explicitly name its subject to ensure clarity and self-containment.
    \item \textbf{Output Schema:}
    \par
    \jsontext{
        "\placeholder{world\_state\_description}"
    }
\end{itemize}

\par\bigskip

\schema{Level 2: Object-Centric States (Nested List)} \hfill \textit{(Entity grouping, entangled attributes)}

\par\noindent\hrulefill

\begin{itemize}[leftmargin=1.2em, nosep, topsep=4pt, label={}]
    \item \textbf{Object Definition:} Refers to specific or abstract entities identified in the observations. Objects must be directly grounded in the observed input.
    \item \textbf{Description Constraint:} A list of distinct attributes or facts associated with the identified object. Each item must be a standalone, grammatically complete English sentence that includes its own subject.
    \item \textbf{Output Schema:}
    \par
    \jsontext{
    \jsonsym{[}\\
    \hspace*{1.5em}\jsonsym{\{} "object": \jsonsym{\{} "\placeholder{obj\_desc}": \jsonsym{[} "\placeholder{world\_state\_desc}", ... \jsonsym{]} \jsonsym{\}} \jsonsym{\}},\\
    \hspace*{1.5em}\jsonsym{\{} "object": \jsonsym{\{} "\placeholder{obj\_desc}": \jsonsym{[} "\placeholder{world\_state\_desc}" \jsonsym{]} \jsonsym{\}} \jsonsym{\}}\\
    \jsonsym{]}
    }
\end{itemize}

\par\bigskip

\schema{Level 3: Object-Attribute States (Full Factorization)} \hfill \textit{(Ours: \textsc{StateFactory})}

\par\noindent\hrulefill

\begin{itemize}[leftmargin=1.2em, nosep, topsep=4pt, label={}]
    \item \textbf{Object Definition:} Refers to specific or abstract entities identified in the observations. Only entities that are explicitly present or clearly implied should be included.
    \item \textbf{Description Constraint:} Use key-value pairs to describe object attributes. The \textbf{"value"} (\placeholder{world\_state\_desc}) must NOT assume the key is part of the sentence. It must be a complete sentence with subject and verb. The \textbf{"key"} (\placeholder{state\_name}) must be short and summative.
    \item \textbf{Output Schema:}
    \par
    \jsontext{
    \jsonsym{[}\\
    \hspace*{1.5em}\jsonsym{\{} "object": \jsonsym{\{} "\placeholder{obj\_desc}": \jsonsym{[} \\
    \hspace*{4.0em}\jsonsym{\{} "\placeholder{state\_key}": "\placeholder{world\_state\_desc}" \jsonsym{\}},\\
    \hspace*{4.0em}\jsonsym{\{} "\placeholder{state\_key}": "\placeholder{world\_state\_desc}" \jsonsym{\}} \\
    \hspace*{2.5em}\jsonsym{]} \jsonsym{\}} \\
    \hspace*{1.5em}\jsonsym{\}} \\
    \jsonsym{]}
    }
\end{itemize}

\end{tcolorbox}
\label{tab:granularity_ablation_prompts}
\end{table}

\subsubsection{LLM Backbone}
\label{app:backbone_ablation}

To isolate the contribution of the underlying LLM's capabilities, we conducted an ablation study by swapping the core backbone of the \textsc{StateFactory} pipeline while keeping all other distinct modules (e.g., prompt constraints, extraction protocols) constant. We evaluated distinct configurations varying in two dimensions: \textbf{parametric scale} (e.g., 14B vs. 30B) and \textbf{inference strategy} (Standard Instruct vs. Reasoning-Enhanced/Thinking modes).

As shown in Table~\ref{tab:backbone_ablation_full}, both dimensions strongly correlate with alignment precision (measured by EPIC Distance $\downarrow$).
\begin{itemize}
    \item \textbf{Impact of Reasoning:} Enabling ``thinking'' modes yields significant gains. For instance, shifting from \textit{GPT-OSS-20B (Low)} to \textit{(Medium)} improves the average EPIC distance from 0.36 to 0.30. Similarly, \textit{Qwen3-30B-Thinking} ($0.41$) substantially outperforms its standard instruction-tuned counterpart ($0.55$), confirming that extended inference capability is crucial for decomposing complex environmental states into structured factors.
    \item \textbf{Impact of Scale:} Performance generally improves with parameter count within the same model family. Comparing \textit{Qwen3-14B} ($0.58$) to \textit{Qwen3-30B} ($0.55$), we observe a natural improvement grounded in model size.
\end{itemize}

Consequently, \textsc{StateFactory} is positioned to scale continuously with future advancements in both model size and reasoning strategies.

\subsubsection{Semantic Embedding}
\label{app:embedding_robustness}

As discussed in the main text, the reliability of \textsc{StateFactory} hinges on the embedding model's ability to maintain \textbf{Semantic Discriminability}—the capacity to cluster semantically equivalent states while segregating distinct or contradictory ones. To quantify this, we utilize the \textbf{Triplet-based Accuracy} metric.

\textbf{Probing Dataset Construction.}
We derived a synthetic probing dataset, $\mathcal{D}_{probe}$, from \textsc{StateFactory}'s intermediate outputs. For each unique state $s_{anc}$ (Anchor), we generated a triplet $(s_{anc}, s_{pos}, s_{neg})$:

\begin{itemize}
    \item \textbf{Anchor ($s_{anc}$):} A raw factorized state description (e.g., \textit{"The apple is sliced"}).
    \item \textbf{Positive ($s_{pos}$):} A semantically equivalent but syntactically distinct paraphrase generated via an LLM. We explicitly instruct the model to vary vocabulary and structure while strictly preserving the truth value (e.g., \textit{"Slices of apple are present"}).
    \item \textbf{Negative ($s_{neg}$):} A state sampled from the same episode's history describing a different object or a contradictory attribute (e.g., \textit{"The apple is whole"} or \textit{"The table is dirty"}). This serves as a distractor to test discriminative power.
\end{itemize}

\textbf{Metric Definition: Triplet-based Accuracy.}
Let $\mathcal{D}$ denote the set of collected triplets, where each triplet $t = (s_{anc}, s_{pos}, s_{neg})$ consists of an anchor, a positive paraphrase, and a negative distractor. Let $d(\cdot)$ be the embedding function and $\text{sim}(\cdot, \cdot)$ denote cosine similarity. The accuracy is defined as:

\begin{equation}
\begin{split}
    \text{Accuracy} &= \frac{1}{|\mathcal{D}|} \sum_{t \in \mathcal{D}} \mathbb{I} \big[ \text{sim}(d(s_{anc}), d(s_{pos})) \\
    &\quad > \text{sim}(d(s_{anc}), d(s_{neg})) \big]
\end{split}
\end{equation}
where $\mathbb{I}[\cdot]$ is the indicator function.

\textbf{Qualitative Case Studies: Decision Boundaries.}
\label{app:qualitative_cases}

To provide deeper insight into performance disparities, we analyze three representative cases from our probing dataset (Table~\ref{tab:case_study_examples}). These examples illustrate the operational boundaries of different embedding architectures, focusing on the binary \textbf{Success} criterion ($sim(anc, pos) > sim(anc, neg)$) versus \textbf{Failure}.

\textbf{Geometric vs. Co-occurrence (Case 1).} 
In spatial tasks, the negative sample often shares high lexical overlap with the anchor (e.g., ``blue block'' and ``orange block'' appear in both). Discriminative models trained on NLI data robustly identify the contradiction between ``on'' and ``standing over''. Models like \textit{embeddinggemma-300m} also pass this test, though they struggle to establish a confident separation compared to discriminative baselines.

\textbf{Topic Clustering vs. Truth Value (Case 2).}
Case 2 illustrates a critical divergence. The negative sample describes the same object (``tissuebox'') but in a contradictory location (``on sidetable''). Embeddings like \textit{embeddinggemma-300m} fail here, likely because they cluster sentences based on shared topics rather than truth conditions. In contrast, discriminative models successfully reject the distractor, strictly adhering to the semantic state.

\begin{table}[H]
\centering
\caption{\textbf{Polarity Decision Boundaries.} We evaluate whether the embedding models correctly rank the semantic paraphrase higher than the distractor. While most models robustly handle spatial contradictions (Case 1), failures in Topic Distraction (Case 2) reveal that some architectures prioritize lexical co-occurrence over factual truth conditions.}
\begin{tcolorbox}[
    colback=gray!2!white,
    colframe=gray!50!black,
    title=\textbf{Qualitative Analysis: Success vs. Failure on Challenging Triplets},
    fonttitle=\bfseries\small,
    boxrule=0.8pt, arc=3pt,
    left=4pt, right=4pt, top=4pt, bottom=4pt
]
\footnotesize
\renewcommand{\arraystretch}{1.3}
\newcommand{\pass}{\textcolor{teal}{\ding{51} \textbf{Success}}}
\newcommand{\fail}{\textcolor{red}{\ding{55} \textbf{Failure}}}

\begin{tabularx}{\textwidth}{@{}p{0.15\textwidth} X p{0.4\textwidth}@{}}
    \toprule
    \textbf{Category} & \textbf{Triplet (Anchor / Positive / Negative)} & \textbf{Polarity Check ($Sim_{pos} > Sim_{neg}$?)} \\ 
    \midrule
    
    \textbf{Case 1: Spatial Relations} \newline \textit{(\texttt{BlocksWorld})} & 
    \textbf{Anc:} ``The blue block is on the orange block.'' \newline
    \textbf{Pos:} ``The blue block is positioned on top of...'' \newline
    \textbf{Neg:} ``The agent is clutching the blue block while standing over the orange block.'' & 
    \pass \ \textit{all-MiniLM-L6-v2}, \textit{all-mpnet-base-v2}, \textit{nli-distilroberta-base-v2} \newline
    \pass \ \textit{embeddinggemma-300m} \\ 
    \midrule
    
    \textbf{Case 2: Topic Distraction} \newline \texttt{(ALFWorld)} & 
    \textbf{Anc:} ``The tissuebox 1 is examined...'' \newline
    \textbf{Pos:} ``...holding the tissuebox under the lamp'' \newline
    \textbf{Neg:} ``The tissuebox 1 is on sidetable 1'' & 
    \pass \ \textit{all-MiniLM-L6-v2}, \textit{all-mpnet-base-v2} \newline
    \fail \ \textit{embeddinggemma-300m} \newline \textit{(Confused by topic overlap)} \\ 
    \bottomrule
\end{tabularx}
\end{tcolorbox}
\label{tab:case_study_examples}
\end{table}

\subsubsection{Comprehensive Ablation Study Ablation Results}
\begin{table}[H]
\centering
\small
\caption{\textbf{Ablation Study I: Representation.} We compare varying state representation granularity across five domains. The results demonstrate that structured representations significantly outperform unstructured text, with our \textbf{Object-Attribute} approach achieving the lowest EPIC Distance ($\downarrow$).}
\label{tab:ablation_granularity}
\resizebox{\textwidth}{!}{%
\small
\begin{tabular}{lcccccc}
\toprule
\multirow{2}{*}{\textbf{State Representation}} & \multicolumn{6}{c}{\textbf{EPIC Distance} ($\downarrow$)} \\
\cmidrule(l){2-7} 
 & \textbf{\texttt{AlfWorld}} & \textbf{\texttt{ScienceWorld}} & \textbf{WebShop} & \textbf{\texttt{BlocksWorld}} & \textbf{\texttt{TextWorld}} & \textbf{Overall} \\ 
\midrule
Unstructured Observations (Raw) & 0.49 & 0.45 & 0.52 & 0.88 & 0.50 & 0.57 \\
Textual States (Flat Denoising) & 0.31 & 0.35 & 0.42 & 0.72 & 0.37 & 0.43 \\
Object-Centric States (Nested)  & 0.33 & 0.33 & 0.40 & 0.45 & 0.26 & 0.35 \\
\rowcolor{blue!5} \textbf{Object-Attribute States (Ours)} & \textbf{0.28} & \textbf{0.29} & \textbf{0.29} & \textbf{0.43} & \textbf{0.20} & \textbf{0.30} \\
\bottomrule
\end{tabular}%
}
\end{table}

\begin{table}[H]
\centering
\small
\caption{\textbf{Ablation Study II: Goal Interpretation.} Comparison between \textbf{Offline (Oracle)} settings using ground-truth goal states and \textbf{Online (Dynamic)} setting using inferred ones. The minimal gap indicates our Dynamic Goal Interpretation is robust without requiring privileged information.}
\label{tab:ablation_modality}
\resizebox{\textwidth}{!}{%
\begin{tabular}{lcccccc}
\toprule
\multirow{2}{*}{\textbf{Goal Source Setting}} & \multicolumn{6}{c}{\textbf{EPIC Distance} ($\downarrow$)} \\
\cmidrule(l){2-7} 
 & \textbf{\texttt{AlfWorld}} & \textbf{\texttt{ScienceWorld}} & \textbf{WebShop} & \textbf{\texttt{BlocksWorld}} & \textbf{\texttt{TextWorld}} & \textbf{Overall} \\ 
\midrule
Offline (Oracle Goal State)          & 0.26 & 0.27 & 0.26 & 0.40 & 0.22 & 0.28 \\
\rowcolor{blue!5} \textbf{Online (Dynamic Goal State)}          & 0.28 & 0.29 & 0.29 & 0.43 & 0.20 & 0.30 \\
\bottomrule
\end{tabular}%
}
\end{table}

\begin{table}[H]
\centering
\small
\caption{\textbf{Ablation Study III: Semantic Embeddings.} We evaluate \textsc{StateFactory} using five different embedding backbones. We report the \textbf{EPIC Distance} ($\downarrow$) on downstream tasks and the \textbf{Triplet Accuracy} ($\uparrow$) on our synthetic probing dataset. Results show a strong correlation: models with higher Triplet Accuracy (better semantic discriminability) consistently yield lower EPIC distances.}
\label{tab:embedding_robustness}
\resizebox{\textwidth}{!}{%
\begin{tabular}{l c ccccc c}
\toprule
\multirow{2}{*}{\textbf{Embedding Backbone}} & \textbf{Probing Metric} & \multicolumn{6}{c}{\textbf{Downstream Performance (EPIC} $\downarrow$\textbf{)}} \\
\cmidrule(lr){2-2} \cmidrule(l){3-8} 
 & \textbf{Triplet Acc.} ($\uparrow$) & \textbf{\texttt{AlfWorld}} & \textbf{\texttt{ScienceWorld}} & \textbf{\texttt{WebShop}} & \textbf{Blocks} & \textbf{\texttt{TextWorld}} & \textbf{Overall} \\ 
\midrule
nli-distilroberta-base-v2 & 0.83 & \textbf{0.28} & \textbf{0.29} & \textbf{0.29} & 0.47 & \textbf{0.20} & 0.31 \\
all-mpnet-base-v2         & 0.76 & \textbf{0.28} & \textbf{0.29} & \textbf{0.29} & 0.54 & \textbf{0.20} & 0.32 \\
all-MiniLM-L6-v2          & 0.81 & \textbf{0.28} & \textbf{0.29} & \textbf{0.29} & \textbf{0.43} & \textbf{0.20} & \textbf{0.30} \\
bge-base-en-v1.5          & 0.78 & 0.32 & 0.34 & 0.31 & 0.41 & 0.25 & 0.33 \\
embedding-gemma-300m      & 0.67 & 0.45 & 0.47 & 0.42 & 0.52 & 0.38 & 0.45 \\
\bottomrule
\end{tabular}%
}
\end{table}

\begin{table}[H]
\centering
\small 
\caption{\textbf{Ablation Study IV: LLM Backbone.} We compare different LLM backbones driving the \textsc{StateFactory} pipeline. \textbf{Lower} EPIC distance ($\downarrow$) indicates better alignment with ground-truth progress. The results highlight that ``Thinking'' variants (Reasoning-Enhanced) and larger parameter scales consistently improve reward precision.}
\label{tab:backbone_ablation_full}

\begin{tabularx}{\textwidth}{l >{\centering\arraybackslash}X >{\centering\arraybackslash}X >{\centering\arraybackslash}X >{\centering\arraybackslash}X >{\centering\arraybackslash}X >{\centering\arraybackslash}X}
\toprule
\multirow{2}{*}{\textbf{Backbone Model}} & \multicolumn{6}{c}{\textbf{EPIC Distance} ($\downarrow$)} \\
\cmidrule(l){2-7} 
 & \textbf{\texttt{AlfWorld}} & \textbf{ScienceWorld} & \textbf{\texttt{WebShop}} & \textbf{Blocks} & \textbf{\texttt{TextWorld}} & \textbf{Overall} \\ 
\midrule
\multicolumn{7}{l}{\textit{Qwen Family (Impact of Scale \& Thinking)}} \\
Qwen3-14B                        & 0.53 & 0.57 & 0.59 & 0.68 & 0.53 & 0.58 \\
Qwen3-30B-A3B-Instruct-2507               & 0.53 & 0.59 & 0.58 & 0.53 & 0.52 & 0.55 \\
\rowcolor{gray!10} Qwen3-30B-A3B-Thinking-2507 & 0.36 & 0.46 & 0.40 & 0.52 & 0.31 & 0.41 \\ 
\midrule
\multicolumn{7}{l}{\textit{GPT-OSS Family (Impact of Reasoning Effort)}} \\
GPT-OSS-20B (Low)                & 0.37 & 0.35 & 0.34 & 0.47 & 0.26 & 0.36 \\
\rowcolor{blue!5} GPT-OSS-20B (Medium) & \textbf{0.28} & \textbf{0.29} & \textbf{0.29} & \textbf{0.43} & \textbf{0.20} & \textbf{0.30} \\ 
\bottomrule
\end{tabularx}
\end{table}

\subsection{Utility for Agent Planning}
\label{app:planning}
\subsubsection{System-1 Policy Implementation Details}
To rigorously evaluate the efficacy of \textsc{StateFactory}, we implemented a standard ReAct (Reasoning + Acting) agent as the primary baseline and the foundational System-1 component of our framework. The implementation follows the standard interleaving of thought generation and action execution, adapted for the specific constraints of \texttt{AlfWorld}, \texttt{BlocksWorld}, and \texttt{ScienceWorld}.

\textbf{Action Parsing and Error Recovery.}
The Language Model is queried to generate a response in a strict format: \texttt{Thought: <reasoning> \textbackslash n Action: <action>}. We employ a robust regular expression parser to extract the command:
\begin{itemize}
    \item \textbf{Syntactic Error Feedback:} If the output format is malformed (e.g., missing the ``Action:'' prefix), the system intercepts the error and returns a synthetic observation: \textit{``Observation: Invalid Format. Please strictly follow the format Thought:... Action:...''}.
    \item \textbf{Semantic Error Feedback:} If the action is syntactically correct but physically impossible (e.g., trying to pick up a table), the environment's native error message is fed back to the agent, allowing for semantic self-correction.
\end{itemize}

\textbf{Action Space Exploration.}
To facilitate efficient exploration, we explicitly provide the agent with the list of currently valid actions (e.g., \texttt{info['admissible\_commands']}) at each step.


\begin{tcolorbox}[
    breakable,
    colback=white,
    colframe=gray!50!black,
    title=\textbf{ReAct Baseline Prompt: \textsc{AlfWorld}} \hfill \textit{(Household Tasks)},
    fonttitle=\bfseries\small,
    boxrule=0.8pt, arc=3pt,
    left=6pt, right=6pt, top=6pt, bottom=6pt,
    before skip=10pt, 
    after skip=10pt   
]
\footnotesize
Interact with a household to solve a task. Imagine you are an intelligent agent in a household environment and your target is to perform actions to complete the task goal. At the beginning of your interactions, you will be given the detailed description of the current environment and your goal to accomplish and Task Knowledge to guide you to finish the task.

For each of your turns, you will be given the observation of the last turn and the valid actions you can take right now (Valid Actions). You should first think about the current condition and plan for your future actions, then output your action in this turn. Your output must strictly follow this format: "Thought: your thoughts.\textbackslash nAction: your next action"

Your response should use the following format:

Thought: <your thoughts>\\
Action: <your next action>

Here is an example:\\
You are in the middle of a room. Looking quickly around you, you see a bathtubbasin 1, a cabinet 2, a cabinet 1, a countertop 1, a garbagecan 1, a handtowelholder 1, a sinkbasin 1, a toilet 1, a toiletpaperhanger 1, and a towelholder 1.\\
Your task is to: put a toiletpaper in toiletpaperhanger.\\
Thought: I need to place a toilet paper roll on the toilet paper hanger. First, I should check if there's already a roll on the hanger or if it's empty.\\
Action: go to toiletpaperhanger 1

Observation: On the toiletpaperhanger 1, you see nothing.

Thought: The toilet paper hanger is empty, so I need to find a new roll of toilet paper to put on it.\\
Action: go to toilet 1

Observation: On the toilet 1, you see a soapbottle 1, and a toiletpaper 1.

Thought: I've found a toilet paper roll on the toilet. I should pick it up so I can place it on the hanger.\\
Action: take toiletpaper 1 from toilet 1

Observation: You pick up the toiletpaper 1 from the toilet 1.

Thought: Now that I have the toilet paper roll, I need to return to the toilet paper hanger to install it.\\
Action: go to toiletpaperhanger 1

Observation: On the toiletpaperhanger 1, you see nothing.

Thought: I'm back at the toilet paper hanger with the roll in hand. It's time to put the toilet paper on the hanger.\\
Action: put toiletpaper 1 in/on toiletpaperhanger 1

----------------------------------------------------------------\\
Now, the real task begins.
\end{tcolorbox}

\vspace{1em}

\begin{tcolorbox}[
    breakable,
    colback=white,
    colframe=gray!50!black,
    title=\textbf{ReAct Baseline Prompt: \textsc{BlocksWorld}} \hfill \textit{(Spatial Logic)},
    fonttitle=\bfseries\small,
    boxrule=0.8pt, arc=3pt,
    left=6pt, right=6pt, top=6pt, bottom=6pt
]
\footnotesize
Interact with a blocks world to solve a stacking task. Imagine you are an intelligent planning agent in a blocks world environment and your target is to perform actions to complete the task goal. At the beginning of your interactions, you will be given the detailed description of the current block configuration and your goal to accomplish.

For each of your turns, you will be given the observation of the last turn and the valid actions you can take right now (Valid Actions). Each valid action is shown as:\\
\hspace*{1em} <ACTION\_ID> -> <human-readable explanation>

You should:\\
1. Use the explanation to understand what each action does.\\
2. Output ONLY the <ACTION\_ID> (the part before "->") in your "Action:" line.

Your response should use the following format:\\
Thought: <your thoughts>\\
Action: <ACTION\_ID>

Here is an example:\\
Task: the green block is on top of the red block\\
Observation: the green block is clear, the red block is on the table, the hand is empty\\
Valid Actions:\\
UNSTACK\_A\_B -> unstack the green block from on top of the red block\\
PICK-UP\_C -> pick up the blue block

Thought: I need to remove the green block from the red block. The action UNSTACK\_A\_B does exactly that.\\
Action: UNSTACK\_A\_B

----------------------------------------------------------------\\
Now, the real task begins.
\end{tcolorbox}

\vspace{1em}

\begin{tcolorbox}[
    breakable,
    colback=white,
    colframe=gray!50!black,
    title=\textbf{ReAct Baseline Prompt: \textsc{ScienceWorld}} \hfill \textit{(Scientific Discovery)},
    fonttitle=\bfseries\small,
    boxrule=0.8pt, arc=3pt,
    left=6pt, right=6pt, top=6pt, bottom=6pt
]
\footnotesize
You are an intelligent agent conducting a scientific experiment in a simulated environment. Your goal is to complete the given task by performing a sequence of valid actions.

The environment contains several rooms: kitchen, foundry, workshop, bathroom, outside, living room, bedroom, greenhouse, art studio, and hallway.

For each of your turns, you will be given the observation of the last turn and the valid actions you can take right now (Valid Actions). You should first think about the current condition and plan for your future actions, then output your action in this turn.

Your response should use the following format:\\
Thought: <your thoughts>\\
Action: <your next action>

Example:\\
Observation: You are in the kitchen. On the counter, you see a beaker 1 and a note 1.\\
Valid Actions: ['look around', 'examine beaker 1', 'read note 1', 'teleport to greenhouse', 'pick up beaker 1']

Thought: I should read the note first—it might contain instructions for the experiment.\\
Action: read note 1

----------------------------------------------------------------\\
Now, the real task begins.
\end{tcolorbox}

\subsubsection{System-1 Policy Enhancement via \textsc{StateFactory}}
\label{app:policy_enhancement}

To overcome the ambiguity of sparse environmental signals in the standard ReAct baseline, we augment the System-1 policy with a \textbf{dense state-aware reward mechanism}. Unlike the original ReAct loop which relies solely on textual observations, our framework explicitly quantifies the ``progress'' of each step via \textsc{StateFactory}. This allows the agent to evaluate the potential utility of candidate actions before finalizing its decision. The streamlined process operates as follows:

\textbf{Candidate Evaluation \& State Abstraction.} 
At each time step $t$, rather than blindly selecting an action, the agent first identifies a set of candidate actions. For each candidate, we leverage the \textsc{StateFactory} engine to project the potential outcome into a structured semantic state $s_{t+1}$. By using \textit{Dynamic Goal Interpretation} prompts, we filter out environmental noise and focus on task-relevant state changes.

\textbf{Differential Reward Calculation.} 
For each candidate transition, we compute an immediate, dense reward $r_t$ that reflects its contribution to the final goal $g$. The reward is defined by the gain in semantic proximity:
\begin{equation}
    r_t = \text{Score}(s_{t+1}, g) - \text{Score}(s_t, g) - \lambda \cdot \mathbb{I}(\text{Repetition})
\end{equation}
This differential score captures the \textit{Relative Gain}, positively rewarding actions that satisfy critical sub-goals (e.g., locating a required tool) while penalizing redundant or cyclical behavior.

\textbf{Reward-Guided Decision Making.} 
The calculated rewards $\{r_t\}$ are injected back into the ReAct context. The prompt is thus augmented from a standard observation to a \texttt{(Candidate Action, Predicted Reward)} assessment. This explicit feedback signal empowers the LLM to perform ``informed'' selection---prioritizing actions with high relative gain and pivoting away from unproductive strategies (low or negative reward) without requiring expensive lookahead simulations.

\vspace{0.8em}
We provide the prompts used to filter and propose the \textbf{Top-$K$ candidate actions} in the subsequent sections. These prompts assist the ReAct policy in identifying logical next steps before simulation. Specifically, to balance exploration and efficiency, we set $K=3$ for \textit{AlfWorld} and \textit{BlocksWorld}, and $K=5$ for \textit{ScienceWorld} due to its significantly larger and more complex action space.


\begin{tcolorbox}[
    breakable,
    colback=white,           
    colframe=OrangeRed!40!black,  
    title=\textbf{Action Selection Prompt: AlfWorld} \hfill \textit{(Household Tasks)},
    fonttitle=\bfseries\small,
    boxrule=0.8pt, arc=2pt,
    left=6pt, right=6pt, top=6pt, bottom=6pt
]
\small\sffamily\raggedright
You are an expert agent playing a text-based adventure game. \par
Your goal is to complete the following Task: \var{\{task\}} \par

\vspace{0.4em}
\textbf{Current Observation:} \par \var{\{obs\}} \par
\textbf{Interaction History (Actions \& Results):} \par \var{\{history\}} \par
\textbf{Admissible Actions (Candidates):} \par \var{\{candidates\_with\_ids\}} \par

\vspace{0.6em}
Your task is to identify the top \var{\{k\}} actions that are most likely to advance the game state. \par

\textbf{Selection Rules:} \par
1. \textbf{Analyze History}: Check "Interaction History". Avoid actions that resulted in "Nothing happens" or revealed empty containers. Avoid loops. \par
2. \textbf{Logical Continuity}: Select actions that logically follow the observation (e.g., see target $\rightarrow$ take it). \par
3. \textbf{Exploration}: If blocked, prioritize navigating to new areas. \par
4. \textbf{Index Only}: Return the ID numbers (e.g., 0, 1) only. \par

\vspace{0.6em}
\textbf{Return JSON Format:} \par
\texttt{\{ \\
\hspace*{1em} "reasoning": "...", \\
\hspace*{1em} "selected\_ids": [id1, id2, ...] \\
\}}
\end{tcolorbox}

\vspace{1em}

\begin{tcolorbox}[
    breakable,
    colback=white,           
    colframe=OrangeRed!40!black,  
    title=\textbf{Action Selection Prompt: ScienceWorld} \hfill \textit{(Scientific Discovery)},
    fonttitle=\bfseries\small,
    boxrule=0.8pt, arc=2pt,
    left=6pt, right=6pt, top=6pt, bottom=6pt
]
\small\sffamily\raggedright
You are a strategic planner for a scientific agent. \par

\textbf{CONTEXT} \par
- Goal: \var{\{task\}} \par
- Current Observation: \var{\{obs\}} \par
- Recent History: \var{\{history\}} \par

\textbf{CANDIDATE ACTIONS} \par
\var{\{candidates\_with\_ids\}} \par

\textbf{SELECTION STRATEGY} \par
1. \textbf{Analyze History}: Do not repeat failed actions. If not in target room, prioritize navigation. \par
2. \textbf{STRICT FOCUS PROTOCOL}: \\
- RISK: Focus on wrong object triggers -100 penalty. \\
- SEQUENCE: Follow "First X, then Y" logic. Do not focus on Y if X is not done. \par
3. \textbf{Filter Noise}: Ignore objects not mentioned in current goal step. \par

\vspace{0.6em}
\textbf{OUTPUT FORMAT} \par
\texttt{\{ \\
\hspace*{1em} "reasoning": "Explain sequence logic...", \\
\hspace*{1em} "selected\_ids": [id1, id2, ...] \\
\}}
\end{tcolorbox}

\vspace{1em}

\begin{tcolorbox}[
    breakable,
    colback=white,           
    colframe=OrangeRed!40!black,  
    title=\textbf{Action Selection Prompt: BlocksWorld} \hfill \textit{(Spatial Logic)},
    fonttitle=\bfseries\small,
    boxrule=0.8pt, arc=2pt,
    left=6pt, right=6pt, top=6pt, bottom=6pt
]
\small\sffamily\raggedright
You are an expert agent solving a Blocks World planning task. \par
Your goal is to complete the following Task: \var{\{task\}} \par

\textbf{Context:} \par
- Observation: \var{\{obs\}} \par
- History: \var{\{history\}} \par
- Candidates: \var{\{candidates\_with\_ids\}} \par

\vspace{0.4em}
Format: \texttt{[ID]: [ACTION\_COMMAND] $\rightarrow$ [description]} \par

\textbf{Your Task:} \par
Select top \var{\{k\}} IDs based on: \par
1. \textbf{Feasibility}: Respect stacking rules. \par
2. \textbf{Relevance}: Move blocks toward target. \par
3. \textbf{Repetition}: Avoid failed actions. \par
4. \textbf{Strategy}: Free blocked blocks first. \par

\vspace{0.6em}
\textbf{Return JSON Format:} \par
\texttt{\{ \\
\hspace*{1em} "reasoning": "...", \\
\hspace*{1em} "selected\_ids": [id1, id2, ...] \\
\}}
\end{tcolorbox}

\subsubsection{System-2 Planning via MCTS and World Models}
\label{app:system2_planning}

While the dense reward mechanism enhances the local policy, complex tasks often require counterfactual reasoning—simulating "what if" scenarios without committing to irreversible actions in the real environment. To achieve this, we implement a System-2 planner using Monte Carlo Tree Search (MCTS), grounded by a pre-trained World Model~\citep{li2025word}.

\textbf{Generative Action Proposal (Top-$K$).} 
Unlike traditional RL environments where the action space is discrete and known, text-based environments feature an unbounded action space. During the expansion phase of MCTS, the World Model cannot list valid moves. Therefore, we utilize the Agent LLM as a \textit{Generator}. It analyzes the current observation and proposes the Top-$K$ (typically $K=3$) most logical actions. As shown in the prompt below, we enforce a ``Mental Scratchpad`` step, requiring the agent to explicitly verify inventory status and object visibility before generating commands, thereby reducing hallucinated actions.

\textbf{Simulation via World Model.} 
Selected actions are executed in a virtual environment simulated by a LLaMA-based World Model~\citep{li2025word}. This model predicts the next textual observation $o_{t+1}$ given the history $H_t$ and action $a_t$. 

\textbf{Truncated Rollout \& Evaluation.} 
A major challenge in text simulation is "hallucination drift," where errors accumulate over long horizons. To mitigate this, we adopt a conservative \textit{Single-Step Rollout} strategy ($d=1$). We prioritize accurate, immediate state evaluation over uncertain long-term planning. The simulated outcome is immediately scored by the \textsc{StateFactory} engine, which calculates the reward based on the semantic state change. This score backpropagates to guide the root action selection.

\textbf{StateFactory as a Semantic Value Function.}
The efficacy of our truncated planning ($d=1$) heavily relies on the density of the evaluation signal. In traditional sparse-reward settings, a single-step lookahead is typically insufficient to capture value, as the terminal goal remains distant. However, \textsc{StateFactory} bridges this gap by converting semantic state transitions into immediate scalar feedback, serving as a robust heuristic for MCTS.


\begin{tcolorbox}[
    breakable,
    colback=white,
    colframe=violet!70!black, 
    title=\textbf{\textsf{System-2 Action Generation: AlfWorld}},
    fonttitle=\bfseries\small,
    boxrule=0.9pt, arc=4pt,
    left=8pt, right=8pt, top=8pt, bottom=8pt,
    shadow={1.5pt}{-1.5pt}{0pt}{black!10}
]
\footnotesize\sffamily\raggedright

You are an expert agent playing AlfWorld, a text-based simulation game. \par
Your goal is to analyze the current state and generate \var{\{k\}} valid, logical next actions to complete the task.

\vspace{0.6em}
\textbf{Contextual Information:}\par
\textbf{Task:} \var{\{task\}} \par
\textbf{Recent History:} \var{\{history\}} \par
\textbf{Current Observation:} \var{\{obs\}}

\vspace{0.8em}
\textbf{Mental Scratchpad (MANDATORY STEP):}\par
You must analyze the situation step-by-step before generating actions:
\begin{enumerate}[label=\arabic*., leftmargin=1.8em, nosep]
    \item \textbf{Inventory Status}: What are you holding right now?
    \item \textbf{Current Location}: Where are you?
    \item \textbf{Visible Objects}: List VALID interactable objects/receptacles.
    \item \textbf{Immediate Sub-goal}: Decide next logical step (Explore/Interact/Process).
\end{enumerate}

\vspace{0.8em}
\textbf{Strict Action Rules:}\par
\begin{enumerate}[label=\arabic*., leftmargin=1.8em, nosep]
    \item \textbf{Object IDs}: ALWAYS use specific numbers (e.g., \textit{mug 1}).
    \item \textbf{Interaction}: Actions ONLY work on visible or inventory objects.
    \item \textbf{Navigation}: You can \textit{go to} any standard furniture in the room.
\end{enumerate}

\vspace{0.8em}
\textbf{Allowed Templates:}\par
\begin{tcolorbox}[colback=black!3, colframe=black!10, left=4pt, right=4pt, top=4pt, bottom=4pt, boxrule=0.4pt]
\scriptsize
\begin{minipage}[t]{0.48\textwidth}
$\bullet$ go to [receptacle X] \\
$\bullet$ take [object X] from [receptacle X] \\
$\bullet$ put [object X] in/on [receptacle X] \\
$\bullet$ open [receptacle X] / close [receptacle X] \\
$\bullet$ toggle [object X] [receptacle X]
\end{minipage}
\begin{minipage}[t]{0.48\textwidth}
$\bullet$ clean [object X] with [receptacle X] \\
$\bullet$ heat [object X] with [receptacle X] \\
$\bullet$ cool [object X] with [receptacle X] \\
$\bullet$ use [object X] / look
\end{minipage}
\end{tcolorbox}

\vspace{0.8em}
\textbf{Output Format:}\par
Return a JSON object containing your analysis and top \var{\{k\}} actions.

\vspace{0.4em}
\begin{tcolorbox}[colback=gray!5, colframe=gray!25, boxrule=0.5pt, arc=2pt, left=5pt, right=5pt, top=5pt, bottom=5pt]
\scriptsize
\{ \\
\hspace*{1em} "analysis": \{ \\
\hspace*{2em} "inventory": "Nothing", \\
\hspace*{2em} "current\_location": "countertop 1", \\
\hspace*{2em} "visible\_objects": ["apple 1", "knife 1", "cabinet 1"], \\
\hspace*{2em} "sub\_goal": "Need to wash the apple, but first I need to take it." \\
\hspace*{1em} \}, \\
\hspace*{1em} "generated\_actions": ["take apple 1 from countertop 1", "go to sinkbasin 1"] \\
\}
Now, generate the response for the current situation.
\end{tcolorbox}
\end{tcolorbox}

\vspace{1em}

\begin{tcolorbox}[
    breakable,
    colback=white,
    colframe=violet!60!black,
    title=\textbf{World Model: State-Aware Simulator} \hfill \textit{(Internal Representation)},
    fonttitle=\bfseries\small,
    boxrule=0.8pt, arc=3pt,
    left=8pt, right=8pt, top=8pt, bottom=8pt
]
\footnotesize\sffamily
\textbf{\color{violet!80!black}[System: Ground Truth State]} \\
You act as a deterministic physics engine for AlfWorld. Your goal is to update the environment state based on user actions. \\
\textit{Hidden Environment Details:}
\begin{itemize}[leftmargin=1.5em, nosep]
    \item \textbf{Receptacles:} On bed 1: book 1, laptop 1; On desk 2: alarmclock 1, mug 1; Drawer 1 (closed): creditcard 1.
    \item \textbf{Toggles:} desklamp 1 is currently OFF.
    \item \textbf{Task Goal:} examine the alarmclock with the desklamp.
\end{itemize}

\vspace{4pt}
\centering \rule{0.9\textwidth}{0.4pt} \par
\vspace{4pt}
\raggedright

\textbf{\color{blue!70!black}[Interaction History]} \\
\textbf{Initial Observation:} You are in a room. You see a bed 1, desk 1, desk 2... \\
\textbf{User Action:} \texttt{go to desk 2} \\
\textbf{Observation:} On the desk 2, you see a alarmclock 1 and a mug 1. \\
\textbf{User Action:} \texttt{use desklamp 1} \\
\textbf{Next Observation (Prediction):} 
\end{tcolorbox}

\subsubsection{Case Study of System-2 Planning} 

\begin{table}[H]
\small
\centering
\caption{Trajectory Trace of Case Study 1: \textbf{``Task: Look at CD under the desklamp.''} This case demonstrates the internal logic of the \textsc{StateFactory} engine, specifically how it facilitates \textbf{dynamic goal interpretation} by anchoring abstract instructions to crystallized physical entities ($g_t$). By extracting structured states ($s_t$) from observations, the engine generates \textbf{continuous reward gradients} (Relative Gain) that guide the MCTS search. Note the significant reward spikes upon environmental constraint satisfaction (Step 7) and the identification of the target entity (Step 8), which effectively steers the agent away from redundant exploration towards a successful execution path.}
\label{tab:system2planning-case1}
\begin{tabularx}{\textwidth}{l p{3.5cm} X p{3.8cm} r}
\toprule
\textbf{Step} & \textbf{Action} & \textbf{Simplified Obs \& State Extraction ($s_t$)} & \textbf{Goal Interpretation ($g_t$)} & \textbf{Reward / Gain} \\
\midrule
1 & \texttt{go to desk 2} & \textbf{Obs:} See \textbf{cd 1} on desk 2. \newline \textbf{State:} \{cd 1: [on desk 2]\} & \textbf{Entity Discovery:} \newline CD $\to$ \textbf{cd 1} & +0.29 $\uparrow$ \\
\midrule
\dots & \multicolumn{4}{c}{\textit{Steps 2--5: Initial Exploration of the Environment}} \\
\midrule
6 & \texttt{go to desk 1} & \textbf{Obs:} See \textbf{desklamp 1} on desk 1. \newline \textbf{State:} \{desklamp 1: [on desk 1]\} & \textbf{Constraint Identification:} \newline Lamp $\to$ \textbf{lamp 1} & +0.02 $\uparrow$ \\
\midrule
7 & \texttt{use desklamp 1} & \textbf{Obs:} Lamp 1 is turned ON. \newline \textbf{State:} \{lamp 1: [\textbf{is turned on}]\} & \textbf{Condition Met:} \newline cd 1 under desklamp 1 & +0.22 $\uparrow$ \\
\midrule
8 & \texttt{go to desk 2} & \textbf{Obs:} Navigate back to retrieve cd 1 with satisfied environment constraints. & \textbf{(Unchanged)} & +0.31 $\uparrow$ \\
\midrule
\dots & \multicolumn{4}{c}{\textit{Steps 9--11: Redundant Exploration \& State Verification}} \\
\midrule
12 & \texttt{take cd 1 from desk 2} & \textbf{Obs:} Pick up cd 1 from desk 2. \newline \textbf{State:} \{\textbf{cd 1: [in hand]}\} & \textbf{(Unchanged)} & +0.12 $\uparrow$ \\
\midrule
13 & \texttt{go to desk 1} & \textbf{Obs:} Arrive at desk 1 with cd 1. \newline \textbf{State:} \{cd 1: [under lamp 1]\} & \textbf{Goal Satisfied.} & 1.00 $\uparrow$ \\
\bottomrule
\end{tabularx}
\end{table}

\section{\textbf{RewardPrediction} Examples}
\label{app:reward_prediction_examples}
\begin{table}[H]
\centering
\small 
\caption{\textbf{Visualized Positive Sample (ScienceWorld).} Task: \textit{"find a(n) living thing... move it to the yellow box."} The agent successfully navigates to the outside, identifies the target (baby wolf), and transports it to the goal location.}
\label{tab:sample_pos_sw}
\begin{tabularx}{\textwidth}{@{}c l X c@{}} 
\toprule
\textbf{Step} & \textbf{Action ($a_t$)} & \textbf{Observation ($o_t$)} & \textbf{GT ($R_t$)} \\ \midrule
0 & look around & This room is called the foundry. In it, you see: the agent... a blast furnace... & 0.00 \\
1 & open door to outside & The door is now open. & 0.08 \\
2 & go to outside & You move to the outside. & 0.17 \\
3 & look around & This outside location is called the outside. Here you see: ... a baby wolf... & 0.17 \\
4 & focus on baby baby wolf & You focus on the baby wolf. & 0.67 \\
5 & pick up baby baby wolf & You move the wolf to the inventory. & 0.75 \\
6 & open door to kitchen & The door is now open. & 0.75 \\
7 & go to kitchen & You move to the kitchen. & 0.75 \\
8 & open door to hallway & The door is now open. & 0.75 \\
9 & go to hallway & You move to the hallway. & 0.75 \\
10 & open door to living room & The door is now open. & 0.75 \\
11 & go to living room & You move to the living room. & 0.83 \\ \midrule
\textbf{12} & \textbf{move baby... to yellow box} & \textbf{You move the wolf to the yellow box.} & \textbf{1.00} \\ \bottomrule
\end{tabularx}
\end{table}

\begin{table}[H]
\centering
\small 
\caption{\textbf{Visualized Negative Sample (ScienceWorld).} Task: \textit{"find a(n) living thing... move it to the yellow box."} The agent performs irrelevant interactions (connecting air, mixing door) that fail to advance the specific task objective.}
\label{tab:sample_neg_sw}
\begin{tabularx}{\textwidth}{@{}c l X c@{}}
\toprule
\textbf{Step} & \textbf{Action ($a_t$)} & \textbf{Observation ($o_t$)} & \textbf{GT ($R_t$)} \\ \midrule
0 & look at door & ERROR: SHOULD USE OVERRIDE FOR PORTAL. & 0.00 \\
1 & connect orange to outside & terminal 1 on orange is now connected to terminal 1 on outside & 0.00 \\
2 & connect air to foundry & terminal 1 on air is now connected to terminal 1 on foundry & 0.00 \\
3 & connect air to steel table & terminal 2 on air is now connected to terminal 1 on table & 0.00 \\
4 & open door & The door is now open. & 0.08 \\
5 & disconnect orange & orange has been disconnected & 0.08 \\
6 & connect drain to foundry & terminal 1 on drain is now connected to terminal 2 on foundry & 0.08 \\
7 & connect agent to inventory & terminal 1 on agent is now connected to terminal 1 on inventory & 0.08 \\
8 & connect sink... & terminal 1 on sink is now connected to terminal 1 on blast furnace & 0.08 \\
9 & look at outside & This outside location is called the outside. Here you see: ... & 0.08 \\
10 & mix door & That container is empty, so there are no items to mix. & 0.08 \\
11 & look in steel table & Inside the table is: nothing & 0.08 \\
12 & mix agent & There is only one thing (inventory) & 0.08 \\ \bottomrule
\end{tabularx}
\end{table}

\vspace{-10pt}

\begin{table}[H]
\centering
\small 
\caption{\textbf{Visualized Positive Sample (\texttt{AlfWorld}).} Task: \textit{"put some cd on safe."} The agent identifies the target object (CD 2) on the dresser, navigates to the safe, opens it, and deposits the item.}
\label{tab:sample_pos_alf_cd}
\begin{tabularx}{\textwidth}{@{}c l X c@{}} 
\toprule
\textbf{Step} & \textbf{Action ($a_t$)} & \textbf{Observation ($o_t$)} & \textbf{GT ($R_t$)} \\ \midrule
0 & go to dresser 1 & You arrive at dresser 1. On the dresser 1, you see a bowl 1... a cd 2... & 0.20 \\
1 & take cd 2 from dresser 1 & You pick up the cd 2 from the dresser 1. & 0.40 \\
2 & go to safe 1 & You arrive at safe 1. The safe 1 is closed. & 0.60 \\
3 & open safe 1 & You open the safe 1. The safe 1 is open. In it, you see nothing. & 0.80 \\ \midrule
\textbf{4} & \textbf{move cd 2 to safe 1} & \textbf{You move the cd 2 to the safe 1.} & \textbf{1.00} \\ \bottomrule
\end{tabularx}
\end{table}

\vspace{-10pt}

\begin{table}[H]
\centering
\small 
\caption{\textbf{Visualized Negative Sample (\texttt{AlfWorld}).} Task: \textit{"put some cd on safe."} The agent wanders to irrelevant locations (laundry hamper, bed, drawers) and fails to interact with the target objects.}
\label{tab:sample_neg_alf_cd}
\begin{tabularx}{\textwidth}{@{}c l X c@{}}
\toprule
\textbf{Step} & \textbf{Action ($a_t$)} & \textbf{Observation ($o_t$)} & \textbf{GT ($R_t$)} \\ \midrule
0 & go to laundryhamper 1 & You arrive at laundryhamper 1. On the laundryhamper 1, you see nothing. & 0.00 \\
1 & go to bed 1 & You arrive at bed 1. On the bed 1, you see a cellphone 2... & 0.00 \\
2 & go to drawer 11 & You arrive at drawer 11. The drawer 11 is closed. & 0.00 \\
3 & look & You are facing the drawer 11. Next to it, you see nothing. & 0.00 \\
4 & go to drawer 6 & You arrive at drawer 6. The drawer 6 is closed. & 0.00 \\ \bottomrule
\end{tabularx}
\end{table}

\vspace{-10pt}

\begin{table}[H]
\centering
\small 
\caption{\textbf{Visualized Positive Sample (\texttt{BlocksWorld}).} Task: \textit{"the red block is on top of the orange block."} The agent unstack the obstructing orange block to clear the red block, then stacks red onto orange. Rewards are linearly interpolated.}
\label{tab:sample_pos_bw}
\begin{tabularx}{\textwidth}{@{}c l X c@{}} 
\toprule
\textbf{Step} & \textbf{Action ($a_t$)} & \textbf{Observation ($o_t$)} & \textbf{GT ($R_t$)} \\ \midrule
0 & unstack the orange block... & ...hand is currently holding orange block, the red block is on the table... & 0.25 \\
1 & put down the orange block & ...hand is empty, the red block is on the table... orange block is on the table... & 0.50 \\
2 & pick up the red block & ...hand is currently holding red block, the blue block is on the table... & 0.75 \\ \midrule
\textbf{3} & \textbf{stack the red block...} & \textbf{...hand is empty, the red block is on top of the orange block...} & \textbf{1.00} \\ \bottomrule
\end{tabularx}
\end{table}

\vspace{-10pt}

\begin{table}[H]
\centering
\small 
\caption{\textbf{Visualized Negative Sample (\texttt{BlocksWorld}).} Task: \textit{"the red block is on top of the orange block."} The agent engages in a repetitive loop with an irrelevant blue block, failing to change the state of the target blocks.}
\label{tab:sample_neg_bw}
\begin{tabularx}{\textwidth}{@{}c l X c@{}}
\toprule
\textbf{Step} & \textbf{Action ($a_t$)} & \textbf{Observation ($o_t$)} & \textbf{GT ($R_t$)} \\ \midrule
0 & pick up the blue block & ...hand is currently holding blue block, the orange block is on top of the red block... & 0.00 \\
1 & put down the blue block & ...hand is empty, the orange block is on top of the red block... & 0.00 \\
2 & pick up the blue block & ...hand is currently holding blue block, the orange block is on top of the red block... & 0.00 \\
3 & put down the blue block & ...hand is empty, the orange block is on top of the red block... & 0.00 \\ \bottomrule
\end{tabularx}
\end{table}

\begin{table}[H]
\centering
\small 
\caption{\textbf{Visualized Positive Sample (\texttt{BlocksWorld}).} Task: \textit{"red on blue, blue on orange."} The agent dismantles the initial wrong pile (Red on Orange on Blue) and reconstructs it in the correct order.}
\label{tab:sample_pos_bw_stack}
\begin{tabularx}{\textwidth}{@{}c l X c@{}} 
\toprule
\textbf{Step} & \textbf{Action ($a_t$)} & \textbf{Observation ($o_t$)} & \textbf{GT ($R_t$)} \\ \midrule
0 & unstack red from orange & ...hand is holding red block, the orange block is on top of the blue block... & 0.13 \\
1 & put down red & ...hand is empty, orange block is on top of blue block, red is on table... & 0.25 \\
2 & unstack orange from blue & ...hand is holding orange block, the blue block is on the table... & 0.38 \\
3 & put down orange & ...hand is empty... blue block is on table... orange block is on table... & 0.50 \\
4 & pick up blue & ...hand is holding blue block, the orange block is on the table... & 0.63 \\
5 & stack blue on orange & ...hand is empty, the blue block is on top of the orange block... & 0.75 \\
6 & pick up red & ...hand is holding red block, blue block is on top of orange block... & 0.88 \\ \midrule
\textbf{7} & \textbf{stack red on blue} & \textbf{...red block is on top of the blue block... blue on top of orange...} & \textbf{1.00} \\ \bottomrule
\end{tabularx}
\end{table}

\vspace{-10pt}

\begin{table}[H]
\centering
\small 
\caption{\textbf{Visualized Negative Sample (\texttt{BlocksWorld}).} Task: \textit{"red on blue, blue on orange."} The agent falls into a reversible loop, repeatedly unstacking and restacking the top block without changing the base structure.}
\label{tab:sample_neg_bw_stack}
\begin{tabularx}{\textwidth}{@{}c l X c@{}}
\toprule
\textbf{Step} & \textbf{Action ($a_t$)} & \textbf{Observation ($o_t$)} & \textbf{GT ($R_t$)} \\ \midrule
0 & unstack red from orange & ...hand is holding red block, the orange block is on top of the blue block... & 0.00 \\
1 & stack red on orange & ...hand is empty, the red block is on top of the orange block... & 0.00 \\
2 & unstack red from orange & ...hand is holding red block, the orange block is on top of the blue block... & 0.00 \\
3 & stack red on orange & ...hand is empty, the red block is on top of the orange block... & 0.00 \\
4 & unstack red from orange & ...hand is holding red block, the orange block is on top of the blue block... & 0.00 \\ \bottomrule
\end{tabularx}
\end{table}

\vspace{-10pt}

\begin{table}[H]
\centering
\small 
\caption{\textbf{Visualized Positive Sample (\texttt{WebShop}).} Task: \textit{"i want a gluten free packaged meal, and price lower than 50.00 dollars."} The agent correctly filters search results, verifies the item description, and purchases a product (\$24.99) meeting the price constraint.}
\label{tab:sample_pos_ws}
\begin{tabularx}{\textwidth}{@{}c l X c@{}} 
\toprule
\textbf{Step} & \textbf{Action ($a_t$)} & \textbf{Observation ($o_t$)} & \textbf{GT ($R_t$)} \\ \midrule
0 & search[gluten free...] & ...Banza Chickpea Mac and Cheese... \textbf{\$35.3} ... Banza Chickpea Pasta Shells... \textbf{\$24.99} ... & 0.20 \\
1 & click[item - banza...] & ...Banza Chickpea Pasta Shells... Price: \textbf{\$24.99} ... [button] Description ... [button] Buy Now & 0.40 \\
2 & click[description] & ...high protein, lower carb, gluten free alternative to traditional mac... & 0.60 \\
3 & click[< prev] & ...Banza Chickpea Pasta Shells... Price: \textbf{\$24.99} ... [button] Buy Now & 0.80 \\ \midrule
\textbf{4} & \textbf{click[buy now]} & \textbf{You have successfully purchased the product.} & \textbf{1.00} \\ \bottomrule
\end{tabularx}
\end{table}

\vspace{-10pt}

\begin{table}[H]
\centering
\small 
\caption{\textbf{Visualized Negative Sample (\texttt{WebShop}).} Task: \textit{"i want a gluten free packaged meal, and price lower than 50.00 dollars."} The agent selects an item priced at \$100.0 (violating the <\$50 constraint) and fails to complete a purchase.}
\label{tab:sample_neg_ws}
\begin{tabularx}{\textwidth}{@{}c l X c@{}}
\toprule
\textbf{Step} & \textbf{Action ($a_t$)} & \textbf{Observation ($o_t$)} & \textbf{GT ($R_t$)} \\ \midrule
0 & search[i want a gluten...] & ...Amy's Vegan... Burrito... \textbf{\$100.0} ... Annie's Shell Pasta... \$39.48 ... & 0.00 \\
1 & click[item - B01FW...] & ...Amy's Vegan, Gluten Free Beans \& Rice Burrito... Price: \textbf{\$100.0} ... & 0.00 \\
2 & click[Features] & One 5.5 oz. Amy's Bean \& Rice... Gluten Free... & 0.00 \\
3 & click[Back to Search] & [button] Search [button\_] & 0.00 \\
4 & click[Search] & [button] Search [button\_] & 0.00 \\ \bottomrule
\end{tabularx}
\end{table}

\vspace{-10pt}

\begin{table}[H]
\centering
\small 
\caption{\textbf{Visualized Positive Sample (\texttt{WebShop}).} Task: \textit{"women ankle strap sandal, black, < 80.00 dollars."} The agent navigates through multiple search pages, identifies a compliant item (Gibobby Sandals, \$1.87-\$3.37), selects the required 'black' attribute, and completes the purchase.}
\label{tab:sample_pos_ws_sandal}
\begin{tabularx}{\textwidth}{@{}c l X c@{}} 
\toprule
\textbf{Step} & \textbf{Action ($a_t$)} & \textbf{Observation ($o_t$)} & \textbf{GT ($R_t$)} \\ \midrule
0 & search[sandal ankle...] & ...Wedge Sandals... \$8.13 ... Gibobby Summer Sandals... \$14.71 ... & 0.14 \\
1 & click[next >] & [button] Back to Search ... Page 2 (Total results: 50) ... & 0.29 \\
2 & click[next >] & [button] Back to Search ... Page 3 (Total results: 50) ... & 0.43 \\
3 & click[next >] & [button] Back to Search ... Page 4 ... Gibobby Summer Sandals... \$1.87 ... & 0.57 \\
4 & click[item - gibobby...] & ...Gibobby Summer Sandals... Price: \textbf{\$1.87 to \$3.37} ... [button] Buy Now & 0.71 \\
5 & click[black] & ...color [clicked button] black... size... & 0.86 \\ \midrule
\textbf{6} & \textbf{click[buy now]} & \textbf{You have successfully purchased the product.} & \textbf{1.00} \\ \bottomrule
\end{tabularx}
\end{table}

\vspace{-10pt}

\begin{table}[H]
\centering
\small 
\caption{\textbf{Visualized Negative Sample (\texttt{WebShop}).} Task: \textit{"women ankle strap sandal, black, < 80.00 dollars."} The agent performs a completely irrelevant search ("plant based") and interacts with food items (Jerky) instead of footwear, failing the task constraints entirely.}
\label{tab:sample_neg_ws_sandal}
\begin{tabularx}{\textwidth}{@{}c l X c@{}}
\toprule
\textbf{Step} & \textbf{Action ($a_t$)} & \textbf{Observation ($o_t$)} & \textbf{GT ($R_t$)} \\ \midrule
0 & search[plant based] & ...Plant-Based Bolognese... \textbf{\$62.99} ... Plant-Based Jerky... \textbf{\$4.98} ... & 0.00 \\
1 & click[item - B085...] & ...Gardein Ultimate Plant-Based Jerky... Price: \$6.29 ... & 0.00 \\
2 & click[teriyaki] & ...flavor name... [clicked button] teriyaki... & 0.00 \\
3 & click[original] & ...flavor name... [clicked button] original... & 0.00 \\
4 & click[Features] & ...Contains one 2.25-ounce package of Gardein... & 0.00 \\
5 & click[< Prev] & ...[clicked button] original... [clicked button] teriyaki... & 0.00 \\
6 & click[hot \& spicy] & ...[clicked button] hot \& spicy... & 0.00 \\ \bottomrule
\end{tabularx}
\end{table}

\vspace{-10pt}

\begin{table}[H]
\centering
\small 
\caption{\textbf{Visualized Positive Sample (\texttt{TextWorld}).} Task: \textit{"lift the cuboid keycard... take the knife... place the knife on the board."} After an initial inspection, the agent executes a perfect sequence: getting the key, unlocking the chest, retrieving the knife, and placing it on the target.}
\label{tab:sample_pos_tw}
\begin{tabularx}{\textwidth}{@{}c l X c@{}} 
\toprule
\textbf{Step} & \textbf{Action ($a_t$)} & \textbf{Observation ($o_t$)} & \textbf{GT ($R_t$)} \\ \midrule
0 & examine type T box & The type T box looks strong, and impossible to destroy... & 0.00 \\
1 & take cuboid keycard & You pick up the cuboid keycard from the ground. & 0.20 \\
2 & unlock cuboid chest... & You unlock the cuboid chest. & 0.40 \\
3 & open cuboid chest & You open the cuboid chest, revealing a potato and a knife. & 0.60 \\
4 & take knife... & You take the knife from the cuboid chest. & 0.80 \\ \midrule
\textbf{5} & \textbf{put knife on board} & \textbf{You put the knife on the board...  The End} & \textbf{1.00} \\ \bottomrule
\end{tabularx}
\end{table}

\vspace{-10pt}

\begin{table}[H]
\centering
\small 
\caption{\textbf{Visualized Negative Sample (\texttt{TextWorld}).} Task: \textit{"lift the cuboid keycard... take the knife... place the knife on the board."} The agent interacts with irrelevant objects (mop, fridge) and picks up the wrong keycard (Type T instead of Cuboid), failing to progress.}
\label{tab:sample_neg_tw}
\begin{tabularx}{\textwidth}{@{}c l X c@{}}
\toprule
\textbf{Step} & \textbf{Action ($a_t$)} & \textbf{Observation ($o_t$)} & \textbf{GT ($R_t$)} \\ \midrule
0 & inventory & You are carrying: a mop. & 0.00 \\
1 & put mop on board & You put the mop on the board. & 0.00 \\
2 & close fridge & You close the fridge. & 0.00 \\
3 & examine mop & The mop would seem to be out of place here & 0.00 \\
4 & take type T keycard & You pick up the type T keycard from the ground. & 0.00 \\
5 & examine board & The board is stable. On the board is a mop. & 0.00 \\ \bottomrule
\end{tabularx}
\end{table}

\vspace{-10pt}

\begin{table}[H]
\centering
\small 
\caption{\textbf{Visualized Positive Sample (\texttt{TextWorld}).} Task: \textit{"retrieve key... unlock safe... insert spoon."} The agent successfully identifies the key, unlocks the container, and retrieves the target tool to complete the puzzle.}
\label{tab:sample_pos_tw_cook}
\begin{tabularx}{\textwidth}{@{}c l X c@{}} 
\toprule
\textbf{Step} & \textbf{Action ($a_t$)} & \textbf{Observation ($o_t$)} & \textbf{GT ($R_t$)} \\ \midrule
0 & take key & You pick up the key from the ground. & 0.20 \\
1 & unlock safe with key & (the safe with the key) You unlock the safe. & 0.40 \\
2 & open safe & (the safe) You open the safe, revealing a spoon. & 0.60 \\
3 & take spoon from safe & (the safe) You take the spoon from the safe. & 0.80 \\ \midrule
\textbf{4} & \textbf{insert spoon...} & \textbf{You put the spoon into the spherical safe...  The End } & \textbf{1.00} \\ \bottomrule
\end{tabularx}
\end{table}

\vspace{-10pt}

\begin{table}[H]
\centering
\small 
\caption{\textbf{Visualized Negative Sample (\texttt{TextWorld}).} Task: \textit{"retrieve key... unlock safe... insert spoon."} The agent gets distracted by irrelevant objects (saucepan, coffee cup) and consumes food instead of solving the lock puzzle.}
\label{tab:sample_neg_tw_cook}
\begin{tabularx}{\textwidth}{@{}c l X c@{}}
\toprule
\textbf{Step} & \textbf{Action ($a_t$)} & \textbf{Observation ($o_t$)} & \textbf{GT ($R_t$)} \\ \midrule
0 & examine saucepan & The saucepan is stable. & 0.00 \\
1 & examine coffee cup & The coffee cup is cheap looking. & 0.00 \\
2 & examine spherical latchkey & The spherical latchkey is cold to the touch & 0.00 \\
3 & eat legume & You eat the legume. Not bad. & 0.00 \\
4 & examine saucepan & The saucepan is stable. & 0.00 \\ \bottomrule
\end{tabularx}
\end{table}

\end{document}